\documentclass[letterpaper]{article} 
\usepackage{aaai2026}  
\usepackage{times}  
\usepackage{helvet}  
\usepackage{courier}  
\usepackage[hyphens]{url}  
\usepackage{graphicx} 
\usepackage{natbib}  
\usepackage{caption} 
\usepackage{algorithm}
\usepackage{algorithmic}
\usepackage{subfigure}
\usepackage{multirow}
\usepackage{booktabs}
\usepackage{amsmath}
\usepackage{amssymb}
\usepackage{xcolor}

\usepackage{newfloat}
\usepackage{listings}
\DeclareCaptionStyle{ruled}{labelfont=normalfont,labelsep=colon,strut=off} 
\floatstyle{ruled}
\newfloat{listing}{tb}{lst}{}
\floatname{listing}{Listing}
\title{D\textsuperscript{3}ToM: Decider-Guided Dynamic Token Merging for Accelerating Diffusion MLLMs}
\author{
    Shuochen Chang,~%
    Xiaofeng Zhang%
    \thanks{Project leader.},~%
    Qingyang Liu,~%
    Li Niu%
    \thanks{Corresponding author.}~%
}
\affiliations{
    MoE Key Lab of Artificial Intelligence, Shanghai Jiao Tong University\\
    
    \{csc1332741686, framebreak, narumimaria, ustcnewly\}@sjtu.edu.cn
}

\usepackage{bibentry}

\begin{document}

\maketitle

\begin{abstract}
Diffusion-based multimodal large language models (Diffusion MLLMs) have recently demonstrated impressive non-autoregressive generative capabilities across vision-and-language tasks. However, Diffusion MLLMs exhibit substantially slower inference than autoregressive models: Each denoising step employs full bidirectional self-attention over the entire sequence, resulting in cubic decoding complexity that becomes computationally impractical with thousands of visual tokens. To address this challenge, we propose D\textsuperscript{3}ToM, a Decider-guided dynamic token merging method that dynamically merges redundant visual tokens at different denoising steps to accelerate inference in Diffusion MLLMs. At each denoising step, D\textsuperscript{3}ToM uses decider tokens—the tokens generated in the previous denoising step—to build an importance map over all visual tokens. Then it maintains a proportion of the most salient tokens and merges the remainder through similarity-based aggregation. This plug-and-play module integrates into a single transformer layer, physically shortening the visual token sequence for all subsequent layers without altering model parameters. Moreover, D\textsuperscript{3}ToM employs a merge ratio that dynamically varies with each denoising step, aligns with the native decoding process of Diffusion MLLMs, achieving superior performance under equivalent computational budgets. Extensive experiments show that D\textsuperscript{3}ToM accelerates inference while preserving competitive performance. The code is released at https://github.com/bcmi/D3ToM-Diffusion-MLLM.
\end{abstract}


\section{Introduction}

Diffusion-based large language models (Diffusion LLMs) have recently emerged as a promising alternative generative paradigm to autoregressive LLMs~\cite{yang2025qwen3}. They implement a discrete diffusion process at inference by starting from a sequence of mask tokens and iteratively decoding discrete text tokens via a trained reverse diffusion process. Pioneering Diffusion-LLMs such as LLaDA~\cite{nie2025large,zhu2025llada} and Dream~\cite{dream2025}, have achieved comparable results to autoregressive LLMs across diverse language tasks~\cite{ye2024beyond,gong2024scaling,gong2025diffucoder,wen2025devil,huang2025reinforcingdiffusionchainlateral}. Subsequently, LLaDA-V~\cite{you2025llada} , MMaDA~\cite{yang2025mmadamultimodallargediffusion} and LaViDa~\cite{li2025lavida} extend this paradigm to vision–and-language tasks via a visual instruction tuning framework that integrates a vision encoder~\cite{tschannen2025siglip} and connector to map visual features into the Diffusion LLM's text embedding space. Diffusion MLLMs achieve impressive non-autoregressive generative capabilities across multimodal benchmarks.

\begin{figure}[t]
  \centering
  \includegraphics[width=0.45\textwidth]{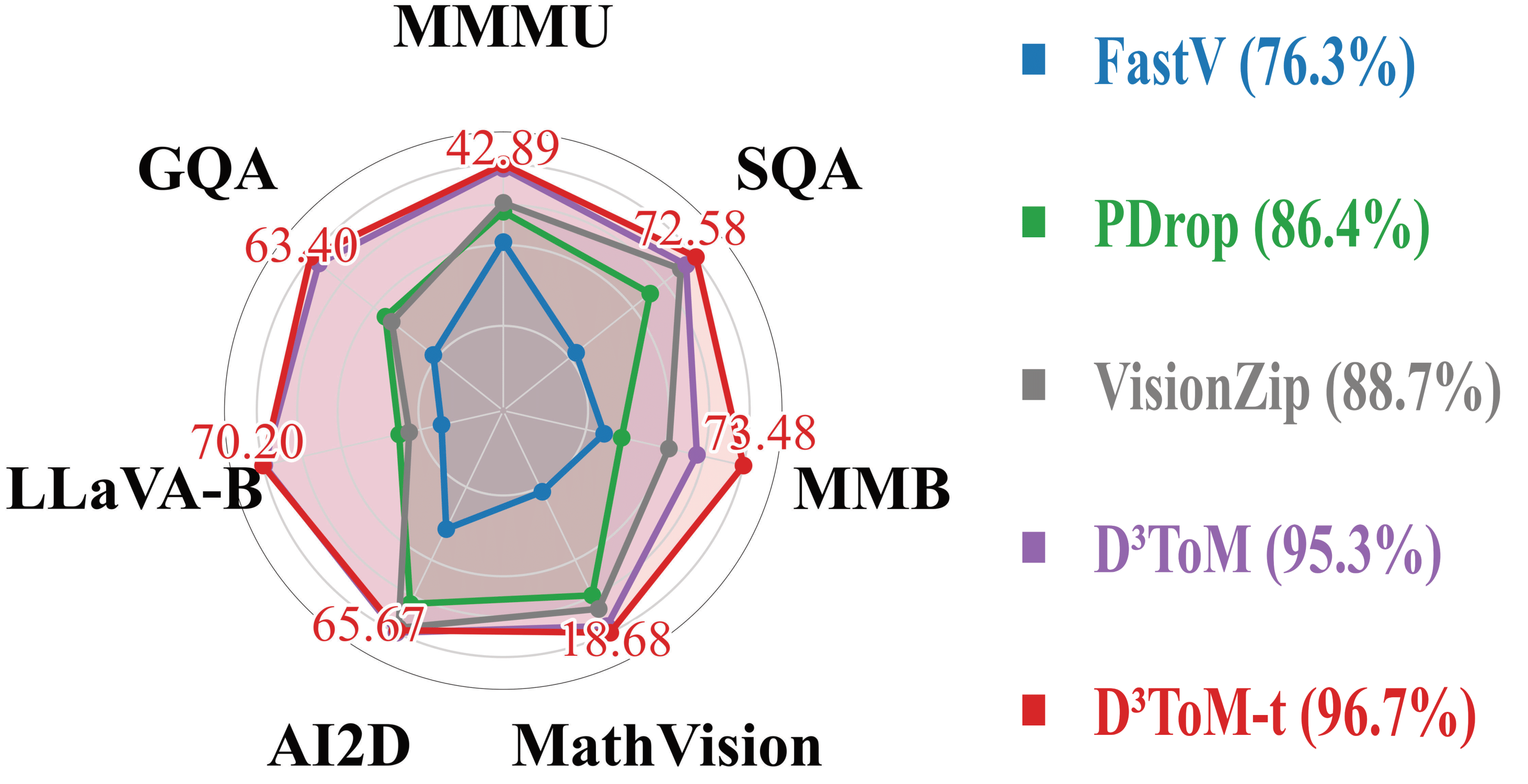}

  \caption{\textbf{D\textsuperscript{3}ToM Performance.} Our D\textsuperscript{3}ToM outperforms the current SOTA methods such as FastV, PDrop and VisionZip, achieving over 96\% of the performance with only 10\% of the visual tokens on LaViDa.}
  \label{fig:perform}
\end{figure}

\begin{figure}[t]
  \centering

  \includegraphics[width=0.45\textwidth]{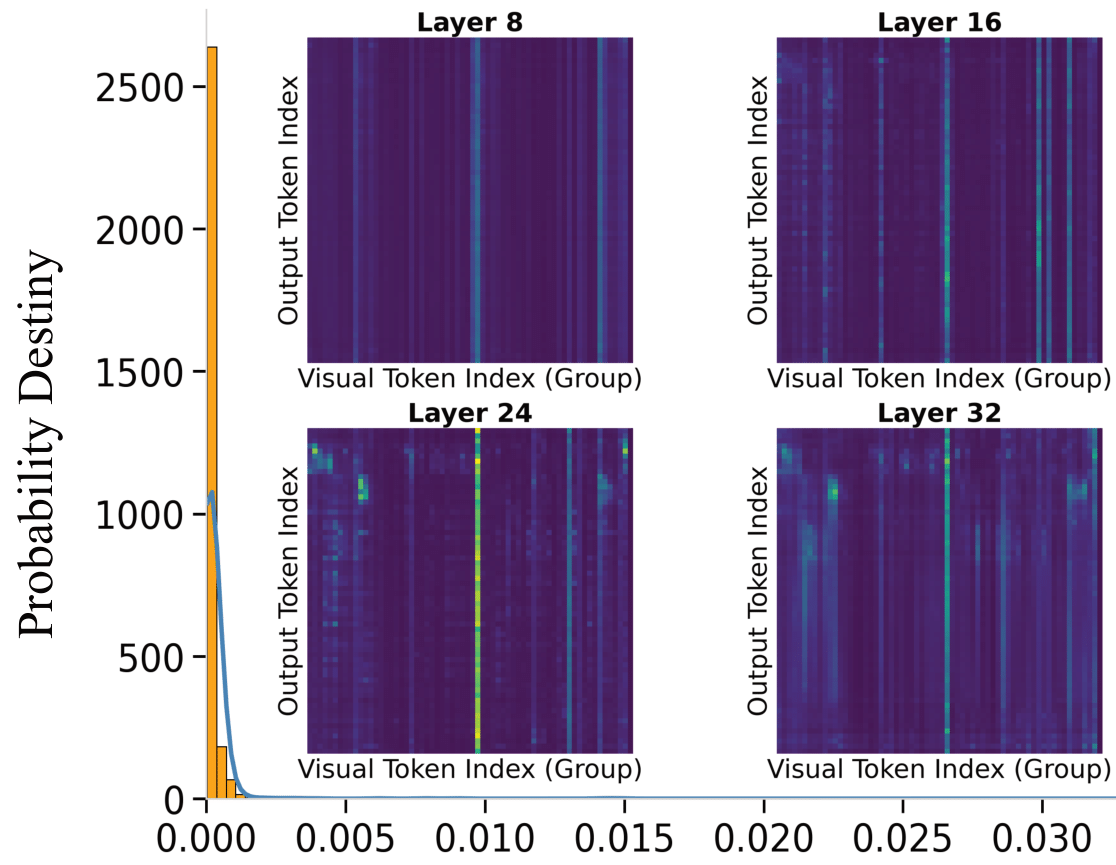}
  \caption{\textbf{Illustration of visual token redundancy.} Probability density distribution of attention weights from output tokens to grouped visual tokens. Because an image contains over one thousand visual tokens while the decoder generates only 64 output tokens, every 16 consecutive visual tokens are averaged to form a single group, and attention from each output token is aggregated over these groups. The resulting distribution shows that most visual-token groups receive near-zero attention, indicating substantial redundancy.}

  \label{fig:reduction}
\end{figure}

Diffusion MLLMs theoretically allow faster decoding than autoregressive models due to their capability of parallel decoding multiple tokens per denoising step, unlike the autoregressive models' sequential one-token-per-step decoding. However, in practice Diffusion MLLMs suffer from cubic time complexity during full-sequence decoding~\cite{ma2025dkv}. Formally, given a total sequence length $N$ (including input and output tokens) and $T=\mathcal{O}(N)$ denoising steps, each step involves full bidirectional self-attention over all $N$ tokens with a complexity of $\mathcal{O}(N^2)$, thus resulting in an overall cubic complexity of $\mathcal{O}(N^3)$. Moreover, Diffusion MLLMs process thousands of visual tokens, making $N$ far larger than in text-only settings and amplifying the cubic cost of full-sequence denoising. Recent advances in Diffusion LLMs, such as Fast-dLLM~\cite{wu2025fast}, dLLM-Cache~\cite{liu2025dllm}, dKV-Cache~\cite{ma2025dkv}, and Prefix-DLM~\cite{li2025lavida}, accelerate inference by integrating existing KV-Cache mechanisms into bidirectional attention. Parallel decoding schemes such as Confidence-Aware Parallel Decoding~\cite{wu2025fast} and SlowFast~\cite{wei2025accelerating} enable simultaneous generation of multiple tokens per denoising step. However, these strategies do not reduce the number of visual tokens, so $N$ remains large and the decoding cost persists in diffusion MLLMs.

\begin{figure}[t]
    \centering
    \includegraphics[width=0.45\textwidth]{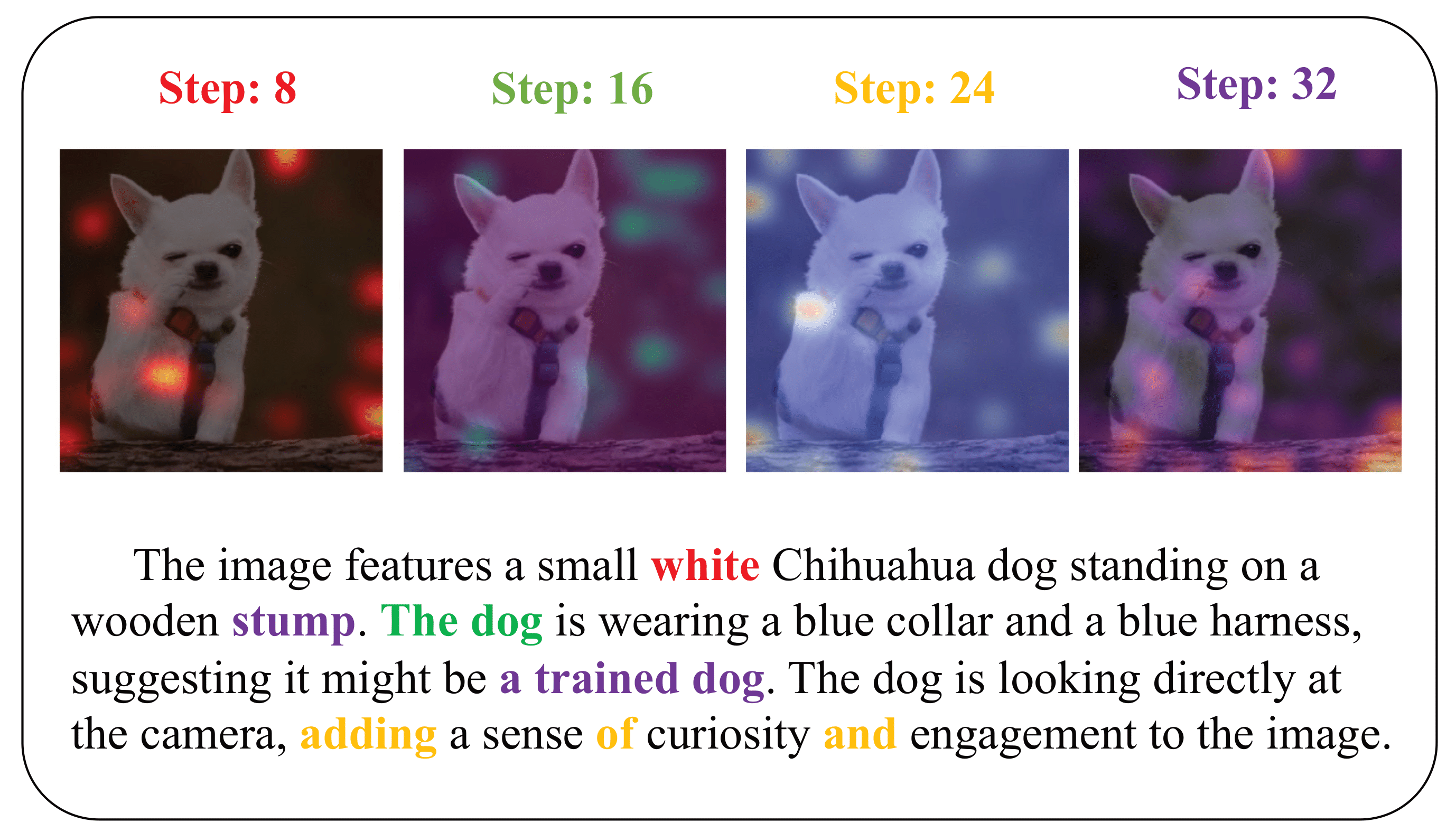}
    \caption{\textbf{Visualization of attention weights.} The figure displays four snapshots of attention weights at different denoising steps (8, 16, 24, and 32) assigned to visual tokens. Each snapshot shows attention distribution of the output tokens generated at the current denoising step on the input image of a small white Chihuahua dog on a wooden stump. The highlighted regions in each image represent the areas of the input image that receive the most attention from the output tokens generated at that specific decoding step. Different colors indicate the attention focus of output tokens generated at different steps.}
    \label{fig:diffusion_attention}
\end{figure}

In our study, we find that Diffusion MLLMs exhibit considerable redundancy among visual tokens. As shown in Figure~\ref{fig:reduction}, the self-attention weight distribution is heavily skewed toward zero and attention values concentrate on a small subset of visual tokens, indicating that only a few tokens capture the majority of visual information while the rest contribute minimally. Although autoregressive MLLM research has leveraged this redundancy through token pruning and merging strategies~\cite{DBLP:conf/iclr/BolyaFDZFH23,DBLP:journals/tomccap/FengXMZ24,DBLP:conf/cvpr/ChoiLCCK24,hyun2025multi,DBLP:conf/eccv/ChenZLBLZC24,zhang2024sparsevlm,jeddi2025similarity,DBLP:conf/cvpr/YangCTWL0J25,DBLP:conf/emnlp/ZhangHXYXJZ024}, similar token compression techniques for Diffusion MLLMs remain unexplored. Furthermore, we observe that the subset of visual tokens receiving high attention shifts significantly across successive denoising steps. As illustrated in Figure~\ref{fig:diffusion_attention}, early iterations concentrate on the Chihuahua’s body, corresponding to the decoding of “white”, whereas later steps focus attention on the background stump as the word “stump” is generated. This evolving attention pattern reveals a step-wise sparsity in visual saliency that static pruning cannot effectively capture.

Motivated by these dynamic attention patterns, we introduce D\textsuperscript{3}ToM, a decider-guided dynamic token merging method for diffusion MLLMs. We define decider tokens as the output token decoded at the previous denoising iteration. At each denoising step $t$, D³ToM treats the attention weights from these decider tokens to all visual tokens as importance scores. Given a merge ratio $\alpha$, it selects the top $1-\alpha$ fraction of visual tokens to keep. For each of the remaining tokens, the method computes its similarity to all kept tokens and merges it into the most similar kept token by aggregating their embedding vectors. This merging is executed within one transformer layer, physically reducing the visual sequence for all downstream layers without modifying any model parameters, and thereby accelerating the denoising inference process.

Moreover, D\textsuperscript{3}ToM modulates its merge ratio $\alpha$ across denoising timesteps to match the shift from global semantic construction to fine-grained refinement. Initial iterations demand broad visual coverage and thus retain more tokens, while later iterations operate on increasingly localized details and can sustain stronger merging. This timestep-dependent adjustment ensures that token merging intensity adapts to dynamic shifts in visual saliency during the diffusion process.

Our extensive experiments validate the proposed design from three key perspectives. For performance, our method retains over 96\% of the baseline model's accuracy across seven standard vision-language benchmarks, even when compressing the visual sequence to just 10\% of its original size (Sec.\ref{sec:main_results}). In terms of efficiency, this same configuration reduces computational cost to only 30\% of the baseline FLOPs and 43\% of the wall-clock time, matching the performance of state-of-the-art inner-LLM pruning techniques (Sec.\ref{sec:efficiency}). Finally, our method demonstrates compatibility with other optimizations (Sec.\ref{sec:ablation_cache}). When combined with KV-Cache, D\textsuperscript{3}ToM yields complementary savings, confirming that our token merging approach addresses an orthogonal computational bottleneck.

Our contributions are summarized as follows:
\begin{itemize}
    \item We propose D\textsuperscript{3}ToM, a novel token merging strategy guided by decider tokens. Our method significantly reduces sequence length by exploiting visual redundancy and adaptively adjusts its merge ratio throughout the denoising process to maintain high model performance.
    \item Our method is designed as a training-free, plug-and-play module requiring no modification of existing model parameters. This ensures broad applicability and seamless integration into diverse Diffusion MLLM architectures.
    \item We demonstrate that our approach is orthogonal to existing KV caching optimizations. When used in conjunction, these methods yield additive efficiency gains by addressing different computational bottlenecks.
\end{itemize}

\begin{figure*}[t]
    \centering
    \includegraphics[width=\textwidth]{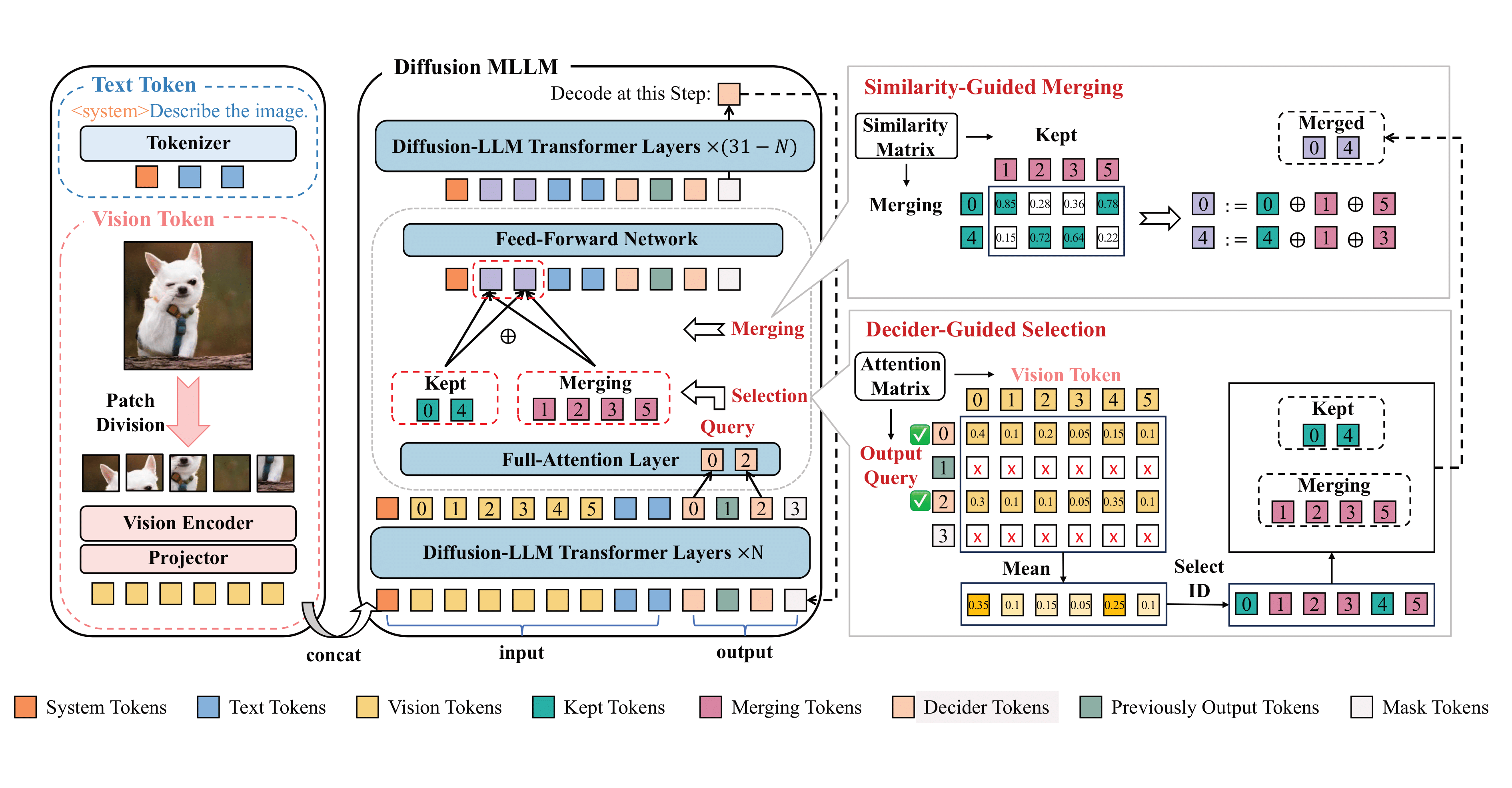}
    \caption{\textbf{The architecture of the D\textsuperscript{3}ToM framework.} The main architecture (left) shows the merging operation occurring at layer $l^*$. The detailed process (right) illustrates the two key stages: (1) Decider-Guided Selection, where decider tokens guide the selection of visual tokens to be kept, and (2) Similarity-Guided Merging, where merging tokens are aggregated into their most similar kept tokens.}
    \label{fig:d3tom_architecture}
\end{figure*}

\section{Related Work}

\subsection{Visual Token Reduction in Autoregressive MLLMs}
To mitigate the computational overhead of lengthy visual token sequences in autoregressive MLLMs, several methods~\cite{xu2025rethinking,huang2024dynamic,xu2025rethinkingvisualtokenreduction,liu2025meteormultiencodercollaborativetoken,DBLP:conf/aaai/Jiang0L0Z25,DBLP:conf/aaai/Ye0LZ25,DBLP:conf/cvpr/AlvarSAZ25,DBLP:conf/cvpr/YangSXHGLYBSH025,DBLP:conf/iclr/KalliniMMPC25} have been proposed. For instance, FastV~\cite{DBLP:conf/eccv/ChenZLBLZC24} prunes tokens with minimal attention scores at intermediate layers. LLaVA-PruMerge~\cite{shang2024llava} adaptively prunes less informative tokens via attention sparsity and subsequently merges them using a k-nearest neighbor strategy. PyramidDrop~\cite{xing2024pyramiddrop} progressively reduces token counts in stages with a lightweight attention-based strategy. Other approaches focus on different criteria. SparseVLM~\cite{zhang2024sparsevlm} prunes visual tokens with low contribution to text-related signals and reconstructs compact representations through clustering. VisionZip~\cite{DBLP:conf/cvpr/YangCTWL0J25} merges dominant visual tokens based on attention and similarity scores. HoliTom~\cite{shao2025holitom} employs pruning by spatial-temporal merging for video LLMs. However, these effective approaches are fundamentally designed for autoregressive specific architectures, rendering them incompatible with the non-causal denoising process in diffusion based MLLMs.

\subsection{Efficiency Enhancements for Diffusion LLMs}
Diffusion LLMs have primarily pursued two directions. The first adapts caching mechanisms to the non-causal attention of diffusion models. Methods like Fast-dLLM~\cite{wu2025fast}, dKV-Cache~\cite{ma2025dkv}, dLLM-Cache~\cite{liu2025dllm}, and Prefix-DLM~\cite{li2025lavida} introduce tailored strategies to cache key-value pairs from stable prefixes or employ adaptive schedules, reducing redundant computations across denoising steps. The second direction focuses on accelerated sampling. Techniques such as Confidence-Aware Parallel Decoding~\cite{wu2025fast}, Confident Decoding in Dimple~\cite{yu2025dimple}, and SlowFast~\cite{wei2025accelerating} Sampling reduce the required denoising iterations by dynamically reordering or parallelizing token decoding. Although these advances substantially accelerate Diffusion LLMs, they do not specifically address the redundancy and computational burden associated with extensive visual token sequences in Diffusion MLLMs. Consequently, the central challenge posed by long visual sequences remains unresolved.

\section{Method}

\subsection{Preliminary: Diffusion MLLMs}
\label{sec:preliminary}

Diffusion MLLMs are a class of generative models that produce text conditioned on multimodal inputs through a non-autoregressive denoising process. The core mechanism involves progressively refining a fully masked sequence into a coherent output that is contextually grounded in both visual and textual information.

Let $\texttt{[MASK]}$ be a dedicated mask token in the token vocabulary set. The model receives a visual input, such as an image $I$, and a text prompt $P$. A vision encoder first processes the image to produce a sequence of visual embeddings $V = E_{\text{vision}}(I)$. The total sequence length for the model's forward pass is denoted by $N$, which includes visual tokens, prompt tokens, and the text sequence to be generated.

The model generates a response sequence $X = (x_1, \dots, x_{O})$ of a fixed length $|O|$ over $T$ discrete denoising steps, indexed by $t = T, \dots, 1$. Let $X^{(t)}$ denote the state of the generated text sequence at step $t$. The process begins with a sequence composed entirely of mask tokens:
\begin{equation}
    X^{(T)} = (\texttt{[MASK]}, \dots, \texttt{[MASK]}).
\end{equation}

At each step $t$, a Transformer-based model $p_\theta$ with $L_{\text{model}}$ layers, parameterized by $\theta$, predicts the probability distribution over the clean sequence $X$ given the current noisy state $X^{(t)}$ and the multimodal context $(V, P)$:
\begin{equation}
    p_\theta(X | X^{(t)}, V, P).
\end{equation}
From this distribution, the most probable clean sequence, $\hat{X}$, is typically identified via greedy decoding. Subsequently, a scheduling function $\mathcal{S}$ determines the state for the next step, $X^{(t-1)}$, by selectively replacing a subset of mask tokens in $X^{(t)}$ with the corresponding predicted tokens from $\hat{X}$:
\begin{equation}
    X^{(t-1)} = \mathcal{S}(\hat{X}, X^{(t)}, t).
\end{equation}
This iterative process is repeated until $t=1$, yielding the final generated sequence $X^{(0)}$. The set of newly predicted tokens in the transition from $X^{(t)}$ to $X^{(t-1)}$ will serve as crucial guides in the subsequent denoising step, as we will detail next.

\subsection{Decider-Guided Dynamic Token Merging}
The core of our methodology is D\textsuperscript{3}ToM, a decider-guided dynamic token merging strategy. We design it as a plug-and-play module to address the computational demands of the iterative denoising process. For clarity, we denote the matrix of hidden states for a sequence of length $N$ as $\mathcal{H} \in \mathbb{R}^{N \times d_{\text{model}}}$, where $d_{\text{model}}$ is the hidden dimension.

\subsubsection{The Decider-Guided Mechanism}
At each diffusion step $t$ ($1\!\le\!t\!\le\!T-1$), we designate the tokens revealed in the preceding iteration as decider tokens. Formally, we define this set as follows:
\begin{equation}
\small
\mathcal{D}^{(t)} =
\Bigl\{\,x_i^{(t)}
      \;\Bigl|\;
      x_i^{(t)}\neq\texttt{[MASK]}
      \;\land\;
      x_i^{(t+1)}=\texttt{[MASK]}
\Bigr\},
\label{eq:decider_def_new}
\end{equation}
where $X^{(t)} = (x_1^{(t)}, \dots, x_{O}^{(t)})$ denotes the partially decoded sequence at step $t$ and $x_i^{(t)}$ is its $i$-th token. At the designated merge layer $l^{*}$, let $\mathcal{H}^{(l^{*},t)} \in \mathbb{R}^{N \times d_{\text{model}}}$ denote the token representations at step $t$, and let $W_Q, W_K \in \mathbb{R}^{d_{\text{model}} \times d_{\text{model}}}$ be the query and key projection matrices.We then compute the self-attention matrix:
\begin{equation}
    A = \operatorname{softmax}\!\Bigl(
        \frac{(\mathcal{H}^{(l^{*},t)}W_Q)\,(\mathcal{H}^{(l^{*},t)}W_K)^{\!\top}}
             {\sqrt{d_{\text{model}}}}
    \Bigr),
\end{equation}
where $A_{i,j}$ denotes the attention weight from output token $x_i^{(t)}$ (query position) to visual token $v_j^{(t)}$ (key position).

We quantify the importance of each visual token $v_j^{(t)}$ by aggregating the attention it receives from all decider tokens in $\mathcal{D}^{(t)}$:
\begin{equation}
    S_j^{(t)} = \sum_{x_i^{(t)} \in \mathcal{D}^{(t)}} A_{i,j}.
    \label{eq:decider_score}
\end{equation}

Based on these scores, we partition the set of all visual tokens $\mathcal{V}$ into two disjoint subsets: the tokens to be kept, $\mathcal{V}_{\text{kept}}(t)$, and those to be merged, $\mathcal{V}_{\text{merge}}(t)$. Let $\text{Rank}(S_j^{(t)})$ be the rank of token $v_j$ according to its score. We define these sets as:
\begin{equation}
\begin{split}
    & \mathcal{V}_{\text{kept}}(t) := \{v_j \in \mathcal{V} \mid \text{Rank}(S_j^{(t)}) \le (1-\alpha_t)|V|\}, \\
    & \mathcal{V}_{\text{merge}}(t) := \mathcal{V} \setminus \mathcal{V}_{\text{kept}}(t).
\end{split}
\end{equation}
To prepare for the next iteration, we update the decider set to exclusively include the most recently revealed tokens. This ensures that only newly decoded tokens guide the next importance computation.

\subsubsection{Similarity-Based Merging}
Given a merge ratio $\alpha_t$, our method merges the tokens in $\mathcal{V}_{\text{merge}}(t)$ into their nearest neighbors within $\mathcal{V}_{\text{kept}}(t)$, based on cosine similarity. This process reduces the sequence length from its pre-merge value, $N$, to a post-merge length, $N_{\mathrm{m}}(t)$, given by:
\begin{equation}
N_{\mathrm{m}}(t)=N-\alpha_t|V|.
\label{eq:Nm}
\end{equation}
The merging operates on the post-attention hidden states 
$\widetilde{\mathcal{H}}^{(l^{*},t)} \!\in\! \mathbb{R}^{N \times d_{\text{model}}}$ 
from layer~$l^{*}$, where 
$\widetilde{\mathcal{H}}^{(l^{*},t)} = (\tilde{h}_1^{(t)}, \dots, \tilde{h}_N^{(t)})$ 
and each $\tilde{h}_j^{(t)} \in \mathbb{R}^{d_{\text{model}}}$ denotes the 
post-attention representation of token.

For each token $v_m^{(t)} \in \mathcal{V}_{\text{merge}}(t)$, we identify its most 
similar counterpart in the kept set using cosine similarity:
\begin{equation}
    k^{\ast} = 
    \arg\max_{v_k^{(t)} \in \mathcal{V}_{\text{kept}}(t)}
    \frac{\tilde{h}_m^{(t)} \cdot \tilde{h}_k^{(t)}}
         {\lVert \tilde{h}_m^{(t)} \rVert_2 \, \lVert \tilde{h}_k^{(t)} \rVert_2}.
\end{equation}

The merging step then aggregates the representations:
\begin{equation}
    \tilde{h}_{k^{\ast}}^{(t)} \leftarrow 
    \tilde{h}_{k^{\ast}}^{(t)} + \tilde{h}_m^{(t)}.
\end{equation}

Finally, we physically remove the merged token positions from the hidden state matrix, yielding a shortened tensor $\widetilde{\mathcal{H}}^{(t)}_{merge}\!\in\!\mathbb{R}^{N_{\mathrm{m}}(t)\times d_{\text{model}}}$. This tensor then propagates through the subsequent layers, reducing the computational complexity for these layers.

\subsubsection{Timestep-Dependent Merge Schedule}
\label{sec:timestep_schedule}

To further align token merging with the evolving nature of the diffusion process, we introduce a timestep-aware variant, D$^{3}$ToM-t, in which the merge ratio $\alpha_t$ varies with the denoising timestep. Intuitively, early iterations construct a coarse semantic draft and therefore benefit from high token retention, whereas later iterations focus on detail refinement and can tolerate more aggressive merging.

Let $\alpha_{\min}$ and $\alpha_{\max}$ denote the minimum and maximum merge ratios allowed during inference. We define a linear schedule:
\begin{equation}
    \alpha_t = 
    \alpha_{\min}
    + \left(\alpha_{\max} - \alpha_{\min}\right)
      \frac{t-1}{T-1},
    \qquad 1 \le t \le T.
    \label{eq:linear_schedule}
\end{equation}
Under this schedule, $\alpha_t$ is smallest at the beginning of decoding and largest near the end, merging more redundant tokens.

\begin{algorithm}[t]
\caption{D\textsuperscript{3}ToM Denoising with Decider-Guided Dynamic Token Merging}
\label{alg:main_process}
\begin{algorithmic}[1]
\STATE \textbf{Input:} visual tokens $V$, prompt tokens $P$, total denoising steps $T$, total layers $L_{\text{model}}$, merge layer $l^{*}$, merge schedule $\{\alpha_t\}_{t=1}^{T}$
\STATE \textbf{Output:} final decoded sequence $X^{(0)}$
\STATE \textbf{Init:} $X^{(T)} \leftarrow (\texttt{[MASK]},\dots,\texttt{[MASK]})$, \quad $\mathcal{D}^{(T)} \leftarrow \emptyset$
\STATE Let $\mathcal{V}$ be the index set of visual tokens in the concatenated sequence
\vspace{0.5ex}
\FOR{$t = T$ \textbf{down to} $1$}
    \STATE $\mathcal{H}^{(0,t)} \leftarrow \operatorname{Embed}(\text{Concat}(V,P,X^{(t)}))$
    \FOR{$l = 1$ \TO $L_{\text{model}}$}
        \STATE $(\widetilde{\mathcal{H}}^{(l,t)}, A^{(l,t)}) \leftarrow \operatorname{SelfAttn}_l\big(\mathcal{H}^{(l-1,t)}\big)$
        \IF{$l = l^{*}$ \AND $\mathcal{D}^{(t)} \neq \emptyset$}
            \STATE $\alpha \leftarrow \alpha_t$ 
            \STATE $S_j^{(t)} \leftarrow \sum_{x_i^{(t)} \in \mathcal{D}^{(t)}} A^{(l,t)}_{i,j}, \quad \forall j \in \mathcal{V}$ 
            \STATE $\mathcal{V}_{\text{kept}}(t) \leftarrow \text{TopK}_{(1-\alpha)|V|}\big(S^{(t)}\big)$
            \STATE $\mathcal{V}_{\text{merge}}(t) \leftarrow \mathcal{V} \setminus \mathcal{V}_{\text{kept}}(t)$
            \FORALL{$v_m \in \mathcal{V}_{\text{merge}}(t)$}
                \STATE $k^{\ast} \leftarrow \displaystyle\arg\max_{v_k \in \mathcal{V}_{\text{kept}}(t)}
                \cos\!\big(\tilde{h}^{(l,t)}_m,\tilde{h}^{(l,t)}_k\big)$
                \STATE $\tilde{h}^{(l,t)}_{k^{\ast}} \leftarrow \tilde{h}^{(l,t)}_{k^{\ast}} + \tilde{h}^{(l,t)}_m$
            \ENDFOR
            \STATE Remove positions in $\mathcal{V}_{\text{merge}}(t)$ from $\widetilde{\mathcal{H}}^{(l,t)}$ to obtain $\widehat{\mathcal{H}}^{(l,t)}$
        \ELSE
            \STATE $\widehat{\mathcal{H}}^{(l,t)} \leftarrow \widetilde{\mathcal{H}}^{(l,t)}$
        \ENDIF
        \STATE $\mathcal{H}^{(l,t)} \leftarrow \operatorname{FFN}_l\big(\widehat{\mathcal{H}}^{(l,t)}\big)$
    \ENDFOR
    \STATE $\hat{X} \leftarrow \operatorname{Decode}\big(\mathcal{H}^{(L_{\text{model}},t)}\big)$
    \STATE $X^{(t-1)} \leftarrow \mathcal{S}\big(\hat{X}, X^{(t)}, t\big)$
    \STATE $\mathcal{D}^{(t-1)} \leftarrow 
           \big\{x_i^{(t-1)} \,\big|\,
           x_i^{(t-1)}\neq\texttt{[MASK]},\;
           x_i^{(t)}=\texttt{[MASK]}\big\}$
\ENDFOR
\RETURN $X^{(0)}$
\end{algorithmic}
\end{algorithm}

\begin{table*}[t]
\centering
\small
\begin{tabular}{l|ccccccc|c}
\toprule
\textbf{Method} & \textbf{MMMU} & \textbf{SQA} & \textbf{MMB} & \textbf{MathVision} & \textbf{AI2D} & \textbf{LLaVA-B}  & \textbf{GQA} & \textbf{Avg.$\uparrow$} \\
\midrule
\multicolumn{9}{c}{\textbf{Upper Bound, Retain 100\% Tokens}} \\
\cmidrule(lr){1-9}
LaViDa & 43.78 & 72.34 & 74.24 & 20.39 & 69.46 & 71.60 & 66.20 & 100.0\% \\
\midrule

\multicolumn{9}{c}{\textbf{Retain 50\% Tokens}} \\
\cmidrule(lr){1-9}
FastV & 40.68 & 69.16 & 64.37 & 16.85 & 62.85 & 58.70 & 56.20 & 87.89\% \\
PDrop & 41.45 & 70.64 & 66.45 & 17.12 & 64.24 & 62.90 & 60.30 & 91.03\% \\
VisionZip & 42.67 & 70.30 & 68.18 & 17.96 & 66.06 & 63.30 & 59.40 & 92.54\% \\
\textbf{D³ToM } & \textbf{43.00} & 72.38 & 71.97 & 18.75 & \textbf{67.76} & 70.20 & 63.00 & 96.85\% \\
\textbf{D³ToM-t } & 42.78 & \textbf{72.53} & \textbf{72.73} & \textbf{20.39} & 66.71 & \textbf{70.80} & \textbf{63.20} & \textbf{98.05\%} \\
\midrule

\multicolumn{9}{c}{\textbf{Retain 33.3\% Tokens}} \\
\cmidrule(lr){1-9}
FastV & 40.87 & 66.47 & 62.75 & 15.46 & 60.38 & 52.60 & 52.80 & 83.68\% \\
PDrop & 41.74 & 70.16 & 64.08 & 17.67 & 64.51 & 62.40 & 57.80 & 90.38\% \\
VisionZip & 42.11 & 70.10 & 68.18 & 18.45 & 65.58 & 62.10 & 57.60 & 91.94\% \\
\textbf{D³ToM } & \textbf{43.11} & 71.92 & 70.70 & 19.28 & \textbf{66.71} & 71.00 & 62.40 & 96.73\% \\
\textbf{D³ToM-t } & 42.67 & \textbf{72.43} & \textbf{73.48} & \textbf{19.74} & 65.67 & \textbf{71.40} & \textbf{63.20} & \textbf{97.59\%} \\

\midrule
\multicolumn{9}{c}{\textbf{Retain 25\% Tokens}} \\
\cmidrule(lr){1-9}
FastV & 40.12 & 61.87 & 60.38 & 15.13 & 57.74 & 49.70 & 50.40 & 80.20\% \\
PDrop & 40.59 & 69.47 & 62.20 & 17.54 & 64.19 & 59.80 & 56.40 & 88.53\% \\
VisionZip & 40.78 & 70.90 & 64.40 & 18.39 & 65.38 & 58.60 & 58.20 & 90.28\% \\
\textbf{D³ToM } & 42.33 & 71.64 & 69.70 & 18.68 & \textbf{66.19} & 70.80 & 60.40 & 95.23\% \\
\textbf{D³ToM-t } & \textbf{43.00} & \textbf{72.48} & \textbf{73.48} & \textbf{20.07} & 65.45 & \textbf{72.10} & \textbf{63.60} & \textbf{98.12\%} \\

\midrule
\multicolumn{9}{c}{\textbf{Retain 16.7\% Tokens}} \\
\cmidrule(lr){1-9}
FastV & 38.13 & 57.45 & 58.46 & 14.26 & 56.68 & 48.90 & 50.10 & 77.25\% \\
PDrop & 40.85 & 68.74 & 60.83 & 17.28 & 63.14 & 60.60 & 56.20 & 87.92\% \\
VisionZip & 41.56 & 70.75 & 67.42 & 17.73 & 65.25 & 58.90 & 57.20 & 90.44\% \\
\textbf{D³ToM } & 41.78 & 72.19 & 71.21 & 18.85 & \textbf{65.67} & 71.30& 62.60 & 96.04\% \\
\textbf{D³ToM-t } & \textbf{43.11} & \textbf{72.73} & \textbf{73.48} & \textbf{20.07} & \textbf{65.67} & \textbf{71.90} & \textbf{63.40} & \textbf{98.16\%} \\

\midrule
\multicolumn{9}{c}{\textbf{Retain 10\% Tokens}} \\
\cmidrule(lr){1-9}
FastV & 38.42 & 55.38 & 57.74 & 14.42 & 55.40 & 49.20 & 48.40 & 76.34\% \\
PDrop & 40.27 & 66.95 & 60.17 & 17.74 & 63.27 & 55.70 & 55.20 & 86.41\% \\
VisionZip & 40.78 & 70.80 & 65.91 & 18.09 & 65.45 & 54.30 & 54.40 & 88.68\% \\
\textbf{D³ToM } & 42.67 & 71.44 & 68.93 & 18.52 & \textbf{65.87} & 70.10 & 62.60 & 95.31\% \\
\textbf{D³ToM-t } & \textbf{42.89} & \textbf{72.58} & \textbf{73.48} & \textbf{18.68} & 65.67 & \textbf{70.20} & \textbf{63.40} & \textbf{96.75\%} \\

\bottomrule
\end{tabular}
\caption{Performance evaluation on standard multimodal benchmarks. Scores are reported for each benchmark, along with the average performance retention relative to the vanilla model (100\%).}
\label{tab:main_results}
\end{table*}

\begin{table}[t]
    \centering
    \small
    \setlength{\tabcolsep}{6pt}
    \begin{tabular}{l|cc|cc}
        \toprule
        & \multicolumn{2}{c|}{\textbf{TFLOPs$\downarrow$}} & \multicolumn{2}{c}{\textbf{Time (s)$\downarrow$}}\\
        \textbf{Method} & Abs. & Rel. & Abs. & Rel. \\
        \midrule
        \multicolumn{5}{c}{\textbf{Upper Bound, Retain 100\% tokens}} \\
        \cmidrule(lr){1-5}
        LaViDa          & 262.60 & 100\% & 1156.65s & 100\% \\
        \midrule
        \multicolumn{5}{c}{\textbf{Retain 50\% tokens}} \\
        \cmidrule(lr){1-5}
        FastV          & 158.10 & 60.2\% & 786.42s & 68.0\% \\
        PDrop          & 187.18 & 71.3\% & 845.61s & 73.1\% \\
        VisionZip      & 143.57 & 54.7\% & 764.14s & 66.1\% \\
        \textbf{D$^{3}$ToM}     & 159.42 & 60.7\% & 766.40s & 66.3\% \\
        \textbf{D$^{3}$ToM-t}   & 160.03 & 60.9\% & 764.89s & 66.1\% \\
        \midrule
        \multicolumn{5}{c}{\textbf{Retain 33.3\% tokens}} \\
        \cmidrule(lr){1-5}
        FastV          & 124.01 & 47.2\% & 667.24s & 57.7\% \\
        PDrop          & 137.13 & 52.2\% & 718.06s & 62.1\% \\
        VisionZip      & 104.74 & 39.9\% & 636.99s & 55.1\% \\
        \textbf{D$^{3}$ToM}     & 125.73 & 47.9\% & 664.89s & 57.5\% \\
        \textbf{D$^{3}$ToM-t}   & 125.99 & 48.0\% & 667.30s & 57.7\% \\
        \midrule
        \multicolumn{5}{c}{\textbf{Retain 25\% tokens}} \\
        \cmidrule(lr){1-5}
        FastV          & 107.22 & 40.8\% & 612.72s & 53.0\% \\
        PDrop          & 119.57 & 45.6\% & 661.43s & 57.2\% \\
        VisionZip      &  85.62 & 32.6\% & 557.85s & 48.2\% \\
        \textbf{D$^{3}$ToM}     & 109.12 & 41.6\% & 604.87s & 52.3\% \\
        \textbf{D$^{3}$ToM-t}   & 109.27 & 41.6\% & 604.68s & 52.3\% \\
        \midrule
        \multicolumn{5}{c}{\textbf{Retain 16.7\% tokens}} \\
        \cmidrule(lr){1-5}
        FastV          &  90.54 & 34.5\% & 554.13s & 47.9\% \\
        PDrop          & 105.87 & 40.3\% & 609.27s & 52.7\% \\
        VisionZip      &  66.62 & 25.4\% & 504.37s & 43.6\% \\
        \textbf{D$^{3}$ToM}    &  92.61 & 35.3\% & 550.37s & 47.6\% \\
        \textbf{D$^{3}$ToM-t}   &  92.68 & 35.3\% & 552.14s & 47.8\% \\
        \midrule
        \multicolumn{5}{c}{\textbf{Retain 10\% tokens}} \\
        \cmidrule(lr){1-5}
        FastV          &  77.14 & 29.4\% & 487.19s & 42.1\% \\
        PDrop          &  97.16 & 37.0\% & 578.31s & 50.0\% \\
        VisionZip      &  51.36 & 19.6\% & 420.01s & 36.3\% \\
        \textbf{D$^{3}$ToM}     &  79.35 & 30.2\% & 492.70s & 42.6\% \\
        \textbf{D$^{3}$ToM-t}   &  79.37 & 30.2\% & 493.88s & 42.7\% \\
        \bottomrule
    \end{tabular}
    \caption{Inference cost under Sec.\ref{sec:flops}. Values are averaged over LLaVA-Bench with $T{=}32$ and $O{=}64$.}
    \label{tab:efficiency_results}
\end{table}

\begin{table}[h]
\centering
\small
\begin{tabular}{l|cccc|c}
\toprule
\textbf{Method*} & \textbf{MMMU} & \textbf{SQA} & \textbf{MMB} & \textbf{AI2D} & \textbf{Avg.$\uparrow$} \\
\midrule
\multicolumn{6}{c}{\textbf{Upper Bound, Retain 100\% Tokens}} \\
\cmidrule(lr){1-6}
LaViDa* & 43.78 & 71.34 & 70.44 & 69.82 & 100\%\\
\midrule
\multicolumn{6}{c}{\textbf{Retain 50\% Tokens}} \\
\cmidrule(lr){1-6}
D³ToM*  & 42.89 & 71.39 & 70.45 & 69.92 & 99.5\% \\
D³ToM-t*  & 43.00 & 71.19 & 70.36 & 69.98 & 99.5\%\\

\midrule
\multicolumn{6}{c}{\textbf{Retain 33.3\% Tokens}} \\
\cmidrule(lr){1-6}
D³ToM*  & 43.00 & 71.29 & 70.10 & 69.88 & 99.4\% \\
D³ToM-t*  & 42.56 & 71.19 & 70.27 & 69.79 & 99.2\% \\

\midrule
\multicolumn{6}{c}{\textbf{Retain 25\% Tokens}} \\
\cmidrule(lr){1-6}
D³ToM* & 43.22 & 71.54 & 70.28& 69.62 & 99.6\% \\
D³ToM-t* & 42.67 & 70.47 & 70.36& 69.92 & 99.1\% \\

\midrule
\multicolumn{6}{c}{\textbf{Retain 16.7\% Tokens}} \\
\cmidrule(lr){1-6}
D³ToM* & 43.22 & 71.59 & 70.10 & 69.72 & 99.6\% \\
D³ToM-t* & 42.44 & 71.34 & 70.45 & 69.75 & 99.2\% \\

\midrule
\multicolumn{6}{c}{\textbf{Retain 10\% Tokens}} \\
\cmidrule(lr){1-6}
D³ToM* & 43.67 & 71.69 & 69.50 & 69.66 & 99.7\% \\
D³ToM-t* & 42.78 & 71.24 & 70.36 & 69.92 & 99.4\% \\
\bottomrule
\end{tabular}
\caption{Compatibility with KV-Cache: \;*~indicates that the method is augmented with the Prefix-DLM Cache.}
\label{tab:ablation_cache}
\end{table}

\section{Experiments}
\label{sec:experiments}

\subsection{Experimental Setup}
\label{sec:exp_setup}

\paragraph{Models and Baselines} 
We implement D³ToM on LaViDa~\cite{li2025lavida} and compare it with representative autoregressive token-reduction methods including FastV~\cite{DBLP:conf/eccv/ChenZLBLZC24}, PyramidDrop~\cite{xing2024pyramiddrop}, and VisionZip~\cite{DBLP:conf/cvpr/YangCTWL0J25}, adapted to the diffusion setting.



\paragraph{Benchmarks.}
We evaluate our method on 7 multimodal benchmarks including MMMU~\cite{DBLP:conf/cvpr/YueNZ0LZSJRSWYY24}, MMBench~\cite{DBLP:conf/eccv/LiuDZLZZYWHLCL24}, LLaVA-Bench~\cite{DBLP:conf/cvpr/LiuLLL24}, GQA~\cite{DBLP:conf/cvpr/HudsonM19}, ScienceQA~\cite{DBLP:journals/jodl/SaikhGMEB22}, AI2D~\cite{DBLP:journals/lre/HiippalaAHKLOTS21} and MathVision~\cite{DBLP:journals/access/AhmadAARABS25}.

\subsection{Main Results}
\label{sec:main_results}
Table~\ref{tab:main_results} demonstrates that D\textsuperscript{3}ToM and its timestep-aware variant D\textsuperscript{3}ToM-t consistently preserve more task accuracy than all competing compression strategies under every token-retention regime. When we retain only a quarter of the visual tokens, D\textsuperscript{3}ToM-t maintains over 98\% of the original LaViDa performance. At the extreme 10\% retained setting, it retains 96.75\%.

The dynamic schedule in D\textsuperscript{3}ToM-t provides consistent gains over the fixed-ratio version across all budgets, indicating that our decider-guided merging preserves critical visual information even under aggressive compression. Collectively, these results confirm that D\textsuperscript{3}ToM delivers state-of-the-art performance retention for diffusion-based MLLMs across a broad spectrum of multimodal tasks.

\subsection{Efficiency Analysis}
\label{sec:efficiency}

\paragraph{Efficientness Evaluation}
\label{sec:flops}
We begin by defining the core parameters for our analysis: $d$ is the hidden size, $m$ is the intermediate feed-forward width, $L$ is the number of layers, and $T$ is the number of denoising steps. The total token count $N$ per step is given by:
\begin{equation}
N = |V| + |P| + |O|.
\end{equation}
The computational cost for a single Transformer layer processing $n$ tokens, denoted $\mathrm{FLOPs}_{\mathrm{layer}}(n)$, consists of self-attention and feed-forward network (FFN) components, as follows:
\begin{equation}
\label{eq:layer_cost}
\mathrm{FLOPs}_{\mathrm{layer}}(n) = \underbrace{4 n d^{2} + 2 n^{2} d}_{\text{self-attention}} + \underbrace{3 n d m}_{\text{FFN}}.
\end{equation}
Consequently, the total cost for a baseline forward pass without merging is:
\begin{equation}
\mathrm{FLOPs}_{\text{baseline}} = T \times L \times \mathrm{FLOPs}_{\mathrm{layer}}(N).
\label{eq:baseline_cost}
\end{equation}
For our method, D$^{3}$ToM, we introduce a merge ratio $\alpha_t\in[0,1)$ at each step $t$, yielding a reduced sequence length $N_{\text{m}}(t) = N - \alpha_t |V|$. This operation introduces a small computational overhead:
\begin{equation}
\mathrm{FLOPs}_{\text{merge}}(t) = 2\alpha_t(1-\alpha_t)\,|V|^{2}d \;+\;\alpha_t |V| d.
\label{eq:merge_overhead}
\end{equation}
The total per-step cost, with merging applied at layer $l^*$, is the sum of costs from the pre-merge layers ($<l^*$), the split-computation layer $l^*$, the post-merge layers ($>l^*$), and the merge overhead itself:
\begin{equation}\small
\label{eq:step_cost_new}
\begin{split}
\mathrm{FLOPs}_{\text{step}}(t) ={}& l^{*}\times\mathrm{FLOPs}_{\mathrm{layer}}(N) \\
& + \bigl[\mathrm{FLOPs}_{\mathrm{attn}}(N) +\mathrm{FLOPs}_{\mathrm{ffn}}\bigl(N_{\text{m}}(t)\bigr)\bigr] \\
& + (L-l^{*}-1)\times\mathrm{FLOPs}_{\mathrm{layer}}\bigl(N_{\text{m}}(t)\bigr) \\
& + \mathrm{FLOPs}_{\text{merge}}(t).
\end{split}
\end{equation}
The total computational cost for D$^{3}$ToM is the sum of these per-step costs over the entire trajectory:
\begin{equation}
\mathrm{FLOPs}_{\text{D${}^{3}$ToM}} = \sum_{t=1}^{T} \mathrm{FLOPs}_{\text{step}}(t).
\label{eq:d3tom_cost}
\end{equation}
When comparing a constant merge ratio $\bar\alpha$ with a time-varying linear schedule $\{\alpha_t\}$ that has the same mean, the primary difference in FLOPs, $\Delta$, arises only from the quadratic self-attention term:
\begin{equation}
\Delta = 4 d |V|^{2}\bigl[(T-1)\operatorname{Var}(\alpha_t)\bigr].
\label{eq:d3tom_t_cost}
\end{equation}
This difference is negligible when $|V|\ll N$, satisfying $\Delta / (4 d N^{2}) < 1\%$. This proves that the cost of the time-varying schedule is approximately equal to that of the constant-ratio configuration:
\begin{equation}
\mathrm{FLOPs}_{\text{D${}^{3}$ToM-t}} \;\simeq\; \mathrm{FLOPs}_{\text{D${}^{3}$ToM}}.
\label{eq:timeaware}
\end{equation}

We derive the FLOPs expressions for all baseline methods under the same diffusion-MLLM assumptions. To ensure a fair comparison, we match their total effective token budget, meaning the cumulative sequence length over all denoising steps, to that of D$^{3}$ToM. In this way, any observed differences reflect the algorithmic behavior rather than discrepancies in token counts.

\paragraph{Efficientness Results}
We evaluate the computational efficiency of our method using both theoretical FLOPs and measured inference time. The results reported in Table~\ref{tab:efficiency_results} are obtained by running LLaVA-Bench on a single NVIDIA A6000 GPU.

Compared with inner-LLM pruning baselines, D\textsuperscript{3}ToM and D\textsuperscript{3}ToM-t use essentially the same FLOPs and inference time as FastV and remain below PDrop.  At the 10\% retention point, D\textsuperscript{3}ToM-t consumes about 30\% of the reference FLOPs and 42\% of the wall-clock time while delivering higher accuracy.  VisionZip lowers raw cost further by deleting tokens in visual encoder, but its one-shot pruning cannot follow the evolving visual saliency during diffusion process. By adjusting token selection at every denoising step, our decider-guided merging maintains competitive efficiency without the performance limitations.

\subsection{Compatibility with KV-Cache}
\label{sec:ablation_cache}

During cached decoding the model stores key and value tensors  
$K,V\!\in\!\mathbb{R}^{(|V|+|P|)\times d_{\text{model}}}$.  
After importance filtering we obtain the kept index set $\mathcal{K}$  
and its complement $\mathcal{M}$.  
For every $m\!\in\!\mathcal{M}$ we route it to the most similar kept token
\begin{equation}
    \pi(m)=\arg\max_{k\in\mathcal{K}}
    \frac{K_{m}\cdot K_{k}}
         {\|K_{m}\|_2\,\|K_{k}\|_2}.
\end{equation}
We then merge caches and renormalise
\begin{equation}
\begin{aligned}
    K_{\pi(m)} &\leftarrow K_{\pi(m)}+K_{m},\\
    V_{\pi(m)} &\leftarrow V_{\pi(m)}+V_{m},
\end{aligned}
\end{equation}
and finally drop the positions in $\mathcal{M}$,  
so subsequent attention operates on the shorter cache of length $L_{\text{kept}}$.  
The procedure is vectorised and does not modify model parameters.

Table~\ref{tab:ablation_cache} demonstrates that integrating \textsc{D\textsuperscript{3}ToM} with the Prefix-DLM cache consistently shortens the cached sequence while retaining alomst full of model accuracy. These confirm decider-guided token merging and KV caching act on distinct parts and can therefore be combined without compromising prediction quality.

\section{Conclusion}

We introduce D\textsuperscript{3}ToM, a simple yet effective approach that physically shortens the visual token sequence in Diffusion MLLMs, thereby reducing the computational burden of denoising inference. By dynamically focusing computation on the most salient visual tokens, D\textsuperscript{3}ToM preserves generation quality throughout the diffusion process. Empirical results confirm that the proposed merging strategy accelerates inference while maintaining competitive accuracy across diverse multimodal benchmarks.

\section*{Acknowledgments}
The work was supported by the National Natural Science Foundation of China (Grant No. 62471287).

\bibliography{aaai2026}

\clearpage
\appendix

\begingroup
\def\thefootnote{\fnsymbol{footnote}}
\setcounter{footnote}{0}

\twocolumn[
\vskip 0.625in minus 0.125in
\begin{center}
{\LARGE\bf Supplementary for D\textsuperscript{3}ToM:\\[0.4em]
Decider-Guided Dynamic Token Merging for Accelerating Diffusion MLLMs\par}

\vskip 0.1in plus 0.5fil minus 0.05in

{\Large\textbf{
Shuochen Chang,~
Xiaofeng Zhang\footnotemark[1],~
Qingyang Liu,~
Li Niu\footnotemark[2]
}\par}

\vskip .2em plus 0.25fil

{\normalsize
MoE Key Lab of Artificial Intelligence, Shanghai Jiao Tong University\\
\{csc1332741686, framebreak, narumimaria, ustcnewly\}@sjtu.edu.cn
\par}

\vskip 1em plus 2fil
\end{center}
]
\endgroup

\begingroup
\def\thefootnote{\fnsymbol{footnote}}
\footnotetext[1]{Project leader.}
\footnotetext[2]{Corresponding author.}
\endgroup

\section{Detailed Experimental Settings}          
\subsection{Implementation Details}
Our method is implemented on the LaViDa-LLaDA-8B models. Evaluation and inference is performed on an NVIDIA RTX A6000(48G) GPU. To ensure a fair comparison of computational cost with baseline methods, we used total denoising FLOPs and total evaluation time. Baselines are configured for comparable FLOPs: FastV prunes same propotion of tokens at same layer; PDrop keeps 1.5$\times$ tokens as we kept after merge at layer 7 then pruning as same propotion with layer 7 at layers 15 and 23, respectively; VisionZip maintain a consistent proportion of input tokens with our method. To compare Inference time, All methods are evaluated on the full LLaVA-Bench set under identical decoding parameters. We report generation length $N_{output}$ used to produce results in the main paper in Table\ref{tab:1}.
\begin{table}[h!]
\centering

\label{tab:dataset_settings}
\begin{tabular}{lc@{\quad}lc}
\toprule
\textbf{Dataset} & \textbf{$|O|$} & \textbf{Dataset} & \textbf{$|O|$} \\
\midrule
MMMU & 16 & AI2D & 16 \\
SQA & 16 & LLaVA-Bench & 64 \\
MMBench & 16 &  GQA & 16 \\
MathVision & 100 \\

\bottomrule
\end{tabular}
\caption{Settings to produce results of Table 1 at Sec4.2 in main paper. By default, we use the denoising timestep $T$ corresponding to the $|O|$. }
\label{tab:1}
\end{table}
\subsection{Datasets}
We list all evaluation on these multimodal (vision-language) understanding benchmarks. We use official implementation of lmms-eval to achieve results,

\subsubsection{Comprehensive reasoning}
\paragraph{MMMU.} The MMMU~\cite{DBLP:conf/cvpr/YueNZ0LZSJRSWYY24} benchmark is designed to access a model's proficieny in tasks requiring expert-level and domain-specific understanding and reasoning. Ti consists of over 10k questions derived from college-level examinations and tutorials across science, engineering and humanities. The questions involve 30 distinct image types like diagrams and chemical structures, challenge models on both advanced perception and deep knowledge. We evaluate all methods on the validation splits.

\paragraph{MMBench.} MMBench~\cite{DBLP:conf/eccv/LiuDZLZZYWHLCL24} implements a multi-level framework to evaluate cross-specific task dimentions. The assessment is structed into 3 different levels. The hierarchical design enables a comprehensive precise analysis of the strengths and weekness across skills. We evaluate methods' performance on dev splits.

\paragraph{LLaVA-Bench.} LLaVA-Bench~\cite{DBLP:conf/cvpr/LiuLLL24} is a cruated benchmark designed to access a model's instruction-following fidelity and robustness to diverse prompt styles. The benchmark consists various categories, including conversation, detailed description, and complex reasoning. The benchmark rigorously tests a model's weaknesses in open-ended visual question answering and detailed image interpretation. We evaluate our method's performance on the LLaVA-Bench full benchmark.

\subsubsection{Visual Question Answering}

\paragraph{AI2D.} The AI2D~\cite{DBLP:journals/lre/HiippalaAHKLOTS21} benchmark targets the complex challenge of diagrammatic reasoning and question answering. It provides 5k diagrams sourced from textbooks, which are rich in structural elements like text labels, arrows, and pointers. The benchmark provides over 15k multiple-choice questions require models to parse this structure, integrate textual and visual information, and perform non-trivial reasoning about the diagram's components and their relationships. We evaluate our method's performance on the AI2D test split.

\paragraph{SQA.} ScienceQA~\cite{DBLP:journals/jodl/SaikhGMEB22} targets complex and multi-step reasoning with scientific domains. The benchmark features a dataset of questions across nature, language and social science. ScienceQA contains results in 26 topics, 127 categories, and 379 skills. The benchmark is explicitly designed to evaluate multimodal understanding, sophisticated reasoning chains and interpretability of model-generated answers. We evaluate methods on the image split to achieve visual ability.

\paragraph{GQA.} The GQA~\cite{DBLP:conf/cvpr/HudsonM19} is composed of three parts: scene graphs, questions, and images. GQA offers a targeted evaluation methodology for visual question answering, grounded in detailed scene graph representations of images. The questions contained in GQA are generated to test model's ability to reason about object attributes, the understanding of visual scenes and image spatial relationships. We evaluate all methods' performance on GQA\_lite split.

\subsubsection{Visual Question Answering}

\paragraph{MathVision} MathVision~\cite{DBLP:journals/access/AhmadAARABS25} benchmark is a carefully curated of 3,040 image-grounded mathematical problems collected from real mathematics competitions. It spanning 16 distinct mathematical disciplines—including algebra, analytic geometry, combinatorics and probability—and organized into five clearly defined difficulty tiers. MathVision offers fine-grained test bed for evaluating MLLMs by markedly widening the topical coverage and problem diversity beyond earlier resources such as MathVista. We conduct our evaluation in Sec.4.2 on the test split.


\section{Detailed Theoretical FLOPs Analysis}
\label{app:flops_detailed}

This appendix provides the detailed theoretical FLOPs derivations for the baseline methods, as referenced in the main text. The analysis herein uses the same core parameters and assumptions established in our efficiency evaluation.

\subsection{Shared Notation}
\label{app:flops_notation}

For our theoretical analysis, we first establish the core parameters of the model and its inputs. We consider a Transformer architecture characterized by a hidden dimension of size $d$, an intermediate feed-forward network (FFN) dimension of size $m$, and a depth of $L$ layers. The generative process operates over $T$ distinct denoising steps.

At each step, the model processes a sequence comprising three components: a set of visual tokens, a set of prompt tokens, and the tokens being generated. We denote their respective counts as $|V|$, $|P|$, and $|O|$. The total sequence length, $N$, processed by the model in a single step is therefore the sum of these components:
\[
N = |V| + |P| + |O|.
\]
The computational cost for a single Transformer layer that processes a sequence of length $n$, which we denote by $\mathrm{FLOPs_{layer}}(n)$, is determined by its self-attention and FFN modules. The cost is formulated as follows:
\begin{equation}
\label{eq:layer_cost_app}
\mathrm{FLOPs_{layer}}(n) =
\underbrace{4\,n\,d^{2} + 2\,n^{2}d}_{\text{self-attention}}
\;+\;
\underbrace{3\,n\,d\,m}_{\text{FFN}}.
\end{equation}
The token reduction methods described are controlled by hyperparameters that determine their aggressiveness. For a standardized comparison, we define a primary controlling parameter, the \textbf{average visual-token reduction ratio}, denoted by $\alpha \in [0, 1)$, where a higher value signifies a more aggressive reduction.

\paragraph{Scope of FLOPs Reporting.}
We report all Floating Point Operations (FLOPs) in this paper as theoretical calculations. This approach allows us to precisely evaluate the algorithmic efficiency of each method, independent of hardware-specific optimizations or system-level overheads like I/O.

Our calculations are based on a fixed LaViDa--LLaDA--8B model configuration. We hold the following architectural and sequence parameters constant throughout the analysis:
\[
d = 4096,\quad m = 12288,\quad L = 32,\quad T = 32,
\]
\[
|V| = 1000,\quad |P| = 64,\quad |O| = 64.
\]
These values result in a total input sequence length of $N = |V| + |P| + |O| = 1128$ tokens for each diffusion step. In our experiments, we then analyze the impact of varying two key hyperparameters: the average visual-token reduction ratio, $\alpha$, and the merge layer index, $l^*$.


\subsection{Vanilla}
\label{app:flops_baseline}

In the baseline configuration, no token reduction is applied. Consequently, the model processes a constant sequence length $N$ in every layer and at each denoising step. The total theoretical cost is therefore:
\begin{equation}
\label{eq:baseline_app}
\mathrm{FLOPs_{baseline}}
  = T \times L \times \mathrm{FLOPs_{layer}}(N).
\end{equation}

\subsection{FastV}
\label{app:flops_fastv}

The FastV~\cite{DBLP:conf/eccv/ChenZLBLZC24} method prunes a fixed fraction $R$ of visual tokens at a specific layer $K$ (where $0 \le K < L$). This operation occurs once per denoising step. Layers with an index less than $K$ process the full sequence of length $N$, while layers with an index of $K$ or greater operate on the reduced sequence. Let $N_{\text{p}} = N - R\,|V|$ denote the sequence length after pruning. The ranking step, which determines which tokens to discard, has a cost of $2d|V|$ FLOPs per denoising step. The total cost for FastV, accumulated over $T$ denoising steps, is composed of the computation for the initial $K$ layers, the subsequent $L-K$ layers on the pruned sequence, and the token ranking overhead:
\begin{equation}
\label{eq:fastv_app}
\begin{split}
\mathrm{FLOPs_{FastV}}
&=
T\times
\Bigl[
      K \times\mathrm{FLOPs_{layer}}\!\bigl(N\bigr)
\\&\quad
      + (L-K)\times\mathrm{FLOPs_{layer}}\!\bigl(N_{\text{p}}\bigr)
\\&\quad
      + 2\,d\,|V|
\Bigr].
\end{split}
\end{equation}

\subsection{PyramidDrop}
\label{app:flops_pdrop}

PyramidDrop~\cite{xing2024pyramiddrop} divides the decoder into four consecutive stages of equal size. For a model with $L=32$ layers (indexed 0--31), the stages contain $L_0=L_1=L_2=L_3=8$ layers each. Let $V_0 = |V|$ be the initial number of visual tokens. Given a user-specified average reduction ratio $\alpha$, the number of visual tokens retained after the first stage is $V_1 = 1.5(1-\alpha)|V|$. Subsequent reduction steps use a constant retention factor $\beta = V_1 / |V| = 1.5(1-\alpha)$, resulting in $V_2 = \beta V_1$ and $V_3 = \beta V_2$ for the remaining stages. The sequence length within stage $i \in \{0, 1, 2, 3\}$ is therefore $n_i = |P| + |O| + V_i$. A ranking step, with a cost proportional to the current number of visual tokens, is performed at the transition to stages 1, 2, and 3. The total cost is the sum of computations within each of the four stages plus the overhead of the three ranking steps, accumulated over $T$ denoising steps:
\begin{equation}
\label{eq:pdrop_app}
\begin{split}
\mathrm{FLOPs_{PDrop}}
&=
T\times
\Bigl[
      \sum_{i=0}^{3}
        L_i \times\mathrm{FLOPs_{layer}}\!\bigl(n_i\bigr)
\\&\quad
      + 2d\,(V_0 + V_1 + V_2)
\Bigr].
\end{split}
\end{equation}

\subsection{VisionZip}
\label{app:flops_vzip}

VisionZip~\cite{DBLP:conf/cvpr/YangCTWL0J25} performs a single, one-time pruning step, discarding a fraction $R$ of visual tokens \emph{before} they are processed by any of the Transformer layers. The resulting sequence length, $N_{\text{z}} = |P| + |O| + (1-R)|V|$, remains constant throughout all $L$ layers and $T$ denoising steps. No subsequent ranking or merging operations occur. The total theoretical cost is therefore:
\begin{equation}
\label{eq:vzip_app}
\mathrm{FLOPs_{VZip}}
  = T \times L \times
    \mathrm{FLOPs_{layer}}\!\bigl(N_{\text{z}}\bigr).
\end{equation}

\subsection{Comparison at an Equal Average Reduction Ratio}
\label{app:flops_equal_budget}

As established in the main text, D\textsuperscript{3}ToM-t use a time-varying schedule $\{\alpha_t\}$ is approximately equal to D\textsuperscript{3}ToM when using a constant average ratio. Therefore, to ensure a fair and clear comparison, we normalize all baselines against a single, constant average reduction ratio, $\alpha$, which corresponds to the mean of our method's schedule. Specifically, for \textbf{FastV}, we set its pruning layer $K$ to match D\textsuperscript{3}ToM's merge layer, $l^*$, and its reduction fraction is set to $R = \alpha$. For \textbf{PyramidDrop}, its own internal average reduction ratio is set to be equal to $\alpha$. For \textbf{VisionZip}, the one-shot reduction fraction is also set to $R=\alpha$.

By coupling all methods to the single parameter $\alpha$, we ensure that each baseline processes, on average, the same number of visual tokens per layer. Thus, any observed differences in the total FLOPs can be attributed directly to the intrinsic costs of each method's reduction mechanism (e.g., ranking overhead) rather than to discrepancies in their overall token processing budgets.

\section{Ablation Studies}                        

\subsection{Impact of Merge Layer $l^{\ast}$ and Merge Ratio $\alpha$ on FLOPs}
\label{app:abl_l_alpha}

This section analyzes how the merge layer index, $l^*$, and the merge ratio, $\alpha$, jointly affect the total computational cost in FLOPs.

\begin{figure}[h!]
    \centering
    \includegraphics[width=\linewidth]{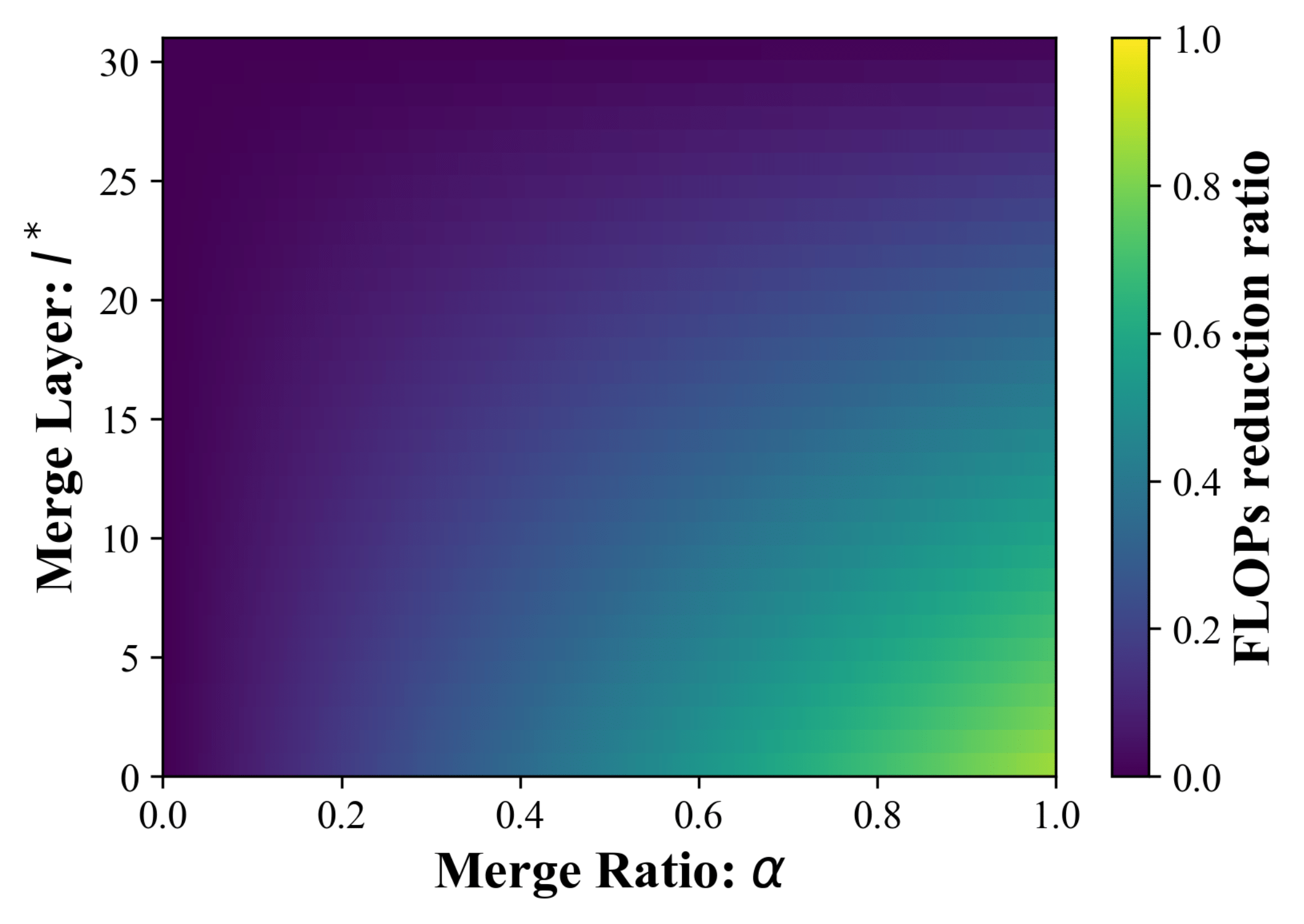}
    \caption{FLOPs reduction (\%) as a function of merge layer $l^{\ast}$ and merge ratio $\alpha$. Higher is better.}
    \label{fig:flops_surface}
\end{figure}

\begin{figure}[h!]
  \centering
  \includegraphics[width=\linewidth]{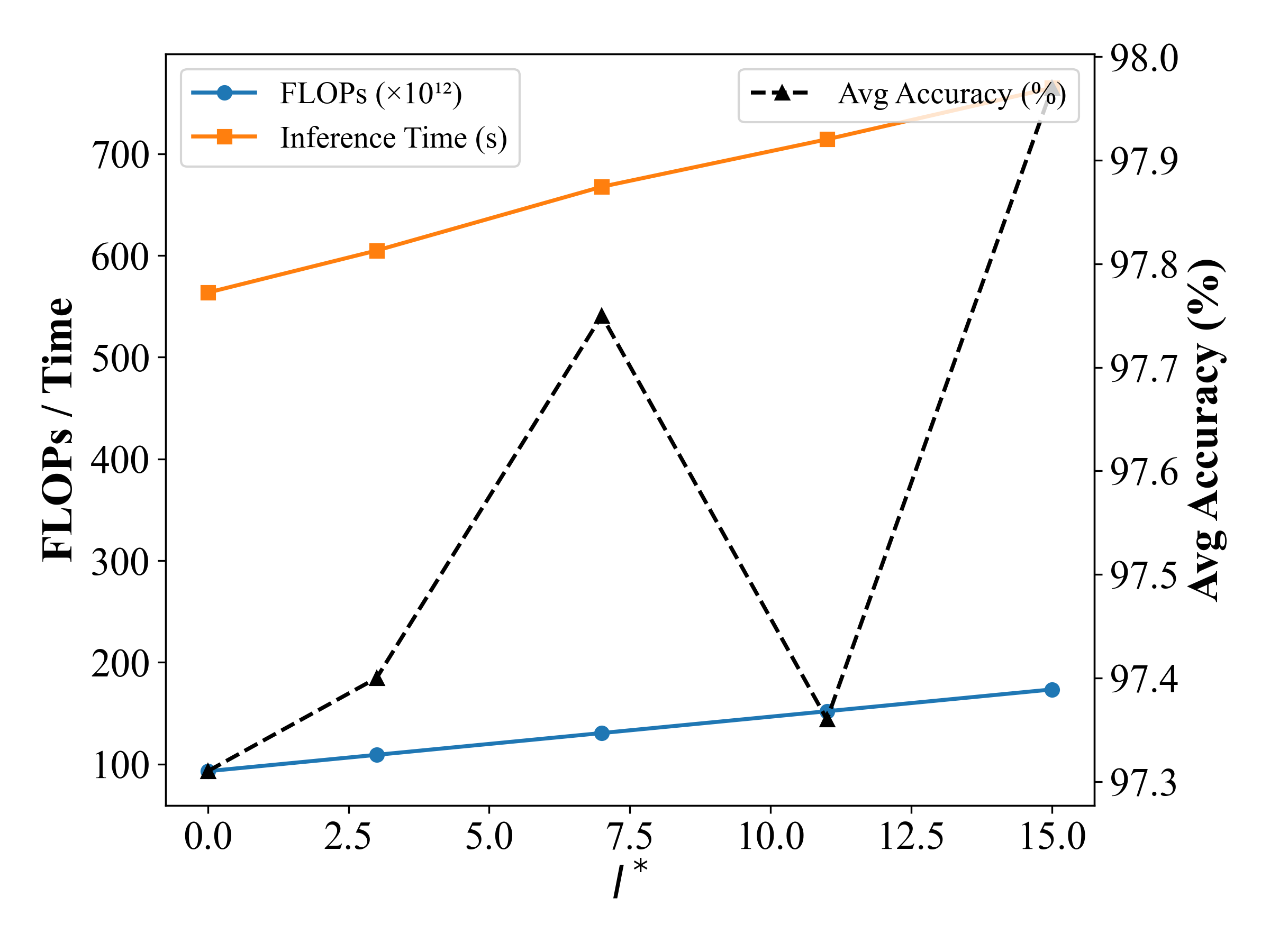}
  \caption{Ablation study on the merge layer~$l^{\ast}$ for fixed merge ratio $\alpha=0.75$. The solid lines show theoretical FLOPs (\(\times10^{12}\)) and single-GPU inference time (s) as $l^{\ast}$ varies, and the dashed line shows average accuracy (\%) across four benchmarks.}
  \label{fig:merge_layer_ablation_plot}
\end{figure}

\begin{figure}[h!]
  \centering
  \includegraphics[width=\linewidth]{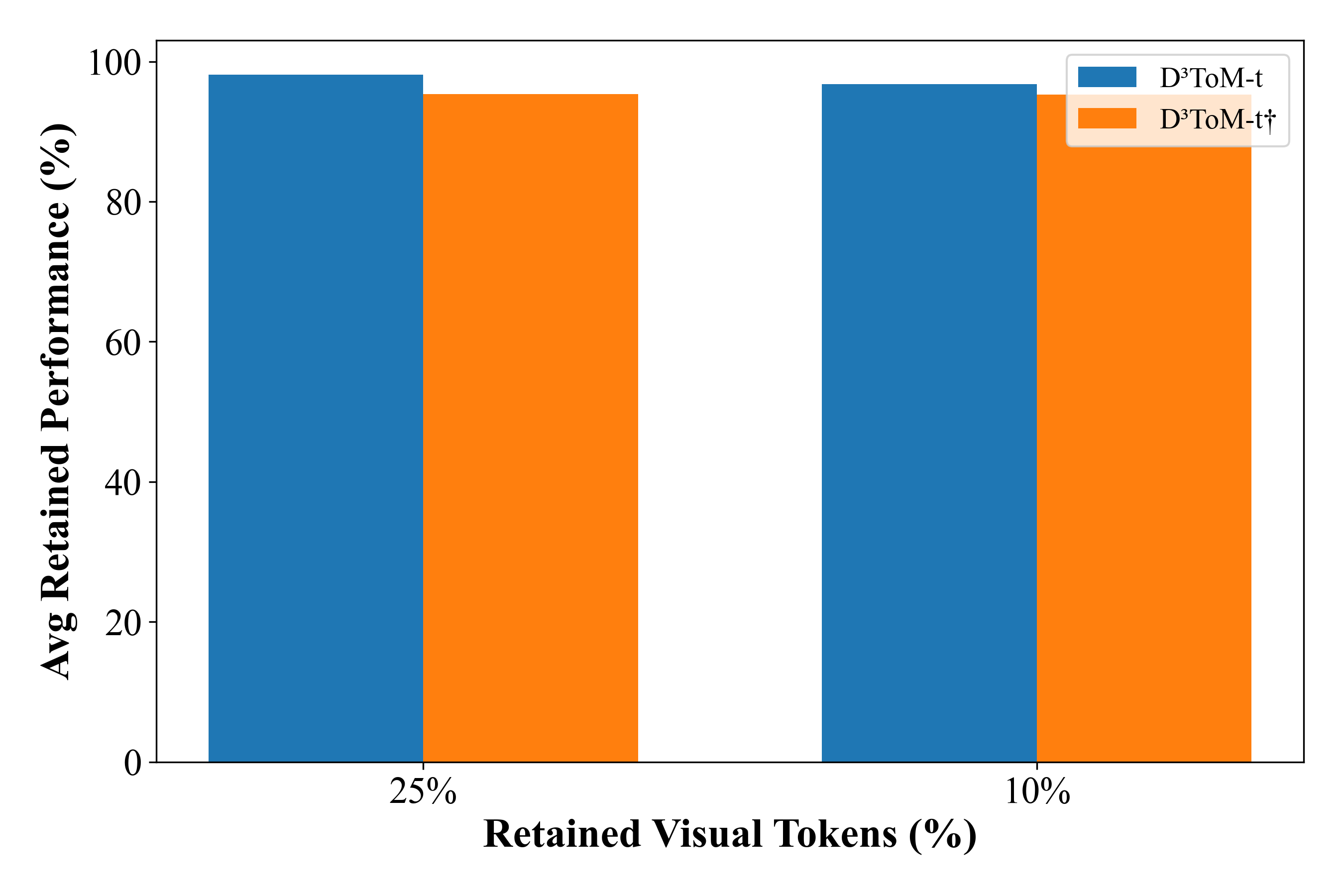}
  \caption{Comparison of average retained performance (\%) for D$^{3}$ToM-t (low\,$\rightarrow$\,high) versus D$^{3}$ToM-t$^{\dagger}$ (high\,$\rightarrow$\,low) merge schedules under 25\% and 10\% token retention.}
  \label{fig:schedule_ablation_plot}
\end{figure}

\begin{table*}[h!]
\centering
\small
\setlength{\tabcolsep}{5pt} 
\begin{tabular}{l |cc | cccc | c}
\toprule
\multirow{2}{*}{\textbf{Method}} &
\multirow{2}{*}{\textbf{FLOPs$\downarrow$}} &
\multirow{2}{*}{\textbf{Time(s)$\downarrow$}} &
\multicolumn{4}{c|}{\textbf{Benchmark Accuracy (\%)}} &
\multirow{2}{*}{\textbf{Avg$\uparrow$}} \\
\cmidrule{4-7} 
& & & \textbf{MMMU}$\uparrow$ & \textbf{LLaVA-Bench}$\uparrow$ & \textbf{SQA}$\uparrow$ & \textbf{AI2D}$\uparrow$ & \\
\midrule
\multicolumn{8}{c}{\textbf{Upper Bound, Retain 100\% Tokens}}\\
\midrule
LaViDa & 262.60~(100\%) & 1156.65s~(100\%) & 43.78 & 71.60 & 72.34 & 69.46 & 100\% \\
\midrule
\multicolumn{8}{c}{\textbf{Retain 25\% Tokens ($\alpha = 0.75$)}}\\
\midrule
$l^{\ast}=0$  & 93.05~(35.43\%) & 563.49s~(48.72\%)  & 42.22 & 70.70 & 71.58 & 66.05 & 97.31\% \\
$l^{\ast}=3$  & 109.12~(41.55\%)& 604.87s~(52.30\%)  & 42.33 & 70.80 & 71.64 & 66.19 & 97.40\% \\
$l^{\ast}=7$  & 130.55~(49.71\%) & 667.56s~(57.71\%)  & 42.11 & 71.20 & 72.24 & 66.34 & 97.75\% \\
$l^{\ast}=11$ & 151.98~(57.88\%) & 714.16s~(61.74\%)  & 42.11 & 70.20 & 71.84 & 66.61 & 97.36\% \\
$l^{\ast}=15$ & 173.41~(66.04\%) & 764.76s~(66.12\%)  & 42.67 & 71.10 & 72.04 & 66.71 & 97.97\% \\
\midrule
\multicolumn{8}{c}{\textbf{Retain 10\% Tokens ($\alpha = 0.90$)}}\\
\midrule
$l^{\ast}=0$  & 60.16~(22.91\%)  & 523.90s~(45.29\%)  & 42.00 & 70.20 & 70.78 & 65.74 & 96.62\% \\
$l^{\ast}=3$  & 79.35~(30.32\%)  & 550.37s~(47.58\%)  & 42.67 & 70.10 & 71.44 & 65.87 & 97.19\% \\
$l^{\ast}=7$  & 104.93~(39.96\%) & 601.18s~(51.98\%)  & 42.44 & 70.30 & 71.67 & 66.27 & 97.40\% \\
$l^{\ast}=11$ & 130.51~(49.70\%) & 655.83s~(56.70\%)  & 41.78 & 70.30 & 71.54 & 66.45 & 97.04\% \\
$l^{\ast}=15$ & 156.09~(59.44\%) & 701.24s~(60.63\%)  & 43.22 & 71.10 & 71.68 & 66.48 & 98.21\% \\
\bottomrule
\end{tabular}
\caption{Ablation study on the merge layer index $l^{\ast}$ for fixed merge ratios $\alpha=0.75$ and $\alpha=0.90$. Each row lists the theoretical FLOPs ($\times10^{12}$), single-GPU inference time (s), and accuracy (\%) on four benchmarks. Inference time is measured on the LLaVA-Bench benchmark with settings $T=32$ and $N_{output}=64$.}
\label{tab:merge_layer_ablation}
\end{table*}

Figure~\ref{fig:flops_surface} plots the theoretical FLOPs reduction as a function of the merge layer, $l^*$, and the merge ratio, $\alpha$. The reduction in computational cost increases monotonically with a higher merge ratio and an earlier merge layer. However, the gains are not uniform across the parameter space. The most significant efficiency improvements are realized when merging occurs within the initial layers (e.g., $l^* < 10$), as deferring the merge to deeper layers yields progressively smaller marginal benefits. This suggests an early, but not necessarily immediate, merge layer may provide a strong balance between efficiency and model stability.

Concurrently, the analysis of the merge ratio, $\alpha$, shows that values below approximately 0.5 result in only modest computational savings. The most substantial efficiency gains begin to accumulate only as $\alpha$ surpasses this threshold. This indicates that the most impactful and practical range for token merging starts where the trade-off between the number of tokens removed and the computational resources saved becomes most favorable.


\subsection{Ablation Study on Merge Layer $l^{\ast}$}
\label{app:abl_fixed_alpha}
Table~\ref{tab:merge_layer_ablation} benchmarks D$^{3}$ToM under a constant retention ratio of $\alpha=0.75$ (i.e.\ 25\% of visual tokens retained) while sweeping $l^{\ast}$.  
We report average accuracy (\%) on four representative datasets and theoretical FLOPs ($\times10^{12}$) with end-to-end wall-clock time (seconds) on LLaVA-Bench.  

We select $l^*=3$ as the default merge layer as it represents the optimal Pareto point between computational cost and model accuracy. Merging at earlier layers (e.g., $l^*=0$) causes a notable degradation in performance, while deferring the merge to later layers ($l^* \ge 7$) provides only marginal accuracy improvements at the cost of substantial efficiency losses. Figure~\ref{fig:merge_layer_ablation_plot} shows that theoretical FLOPs and inference time both increase with $l^*$. We choose $l^* = 3$ and exhibits results in main paper.

\subsection{Ablation Study on Timestep-Dependent Merge Schedule}
\label{app:timestep_ablation}

This ablation isolates the influence of the merge-ratio direction.  
D$^{3}$ToM-t applies a low\,$\rightarrow$\,high linear schedule so that early diffusion steps retain most visual tokens while later steps merge aggressively.  
D$^{3}$ToM-t$^{\dagger}$ reverses the schedule, beginning with the strongest compression and relaxing it over time.  
Both variants are calibrated such that the mean retention equals the constant 25\% or 10\% settings, leaving the cumulative computational budget unchanged.  
Results are reported on the seven evaluation benchmarks used in the main paper; the last column gives the average performance retained relative to the full-token LaViDa baseline.

Result in Table~\ref{tab:ablation_schedule} demonstrate that D$^{3}$ToM-t outperform the high\,$\rightarrow$\,low linear schedule. Figure~\ref{fig:schedule_ablation_plot} shows that D³ToM-t consistently yields higher average retained performance than D³ToM-t† under both 25\% and 10\% retention settings. 

\begin{table*}[t]
\centering
\small
\begin{tabular}{l|ccccccc|c}
\toprule
\textbf{Method} & \textbf{MMMU} & \textbf{SQA} & \textbf{MMB} & \textbf{MathVision} & \textbf{AI2D} & \textbf{LLaVA-B} & \textbf{GQA} & \textbf{Avg.$\uparrow$} \\
\midrule
\multicolumn{9}{c}{\textbf{Upper Bound, Retain 100\% Tokens}} \\
\cmidrule(lr){1-9}
LaViDa & 43.78 & 72.34 & 74.24 & 20.39 & 69.46 & 71.60 & 66.20 & 100.00\% \\
\midrule
\multicolumn{9}{c}{\textbf{Retain 25\% Tokens}} \\
\cmidrule(lr){1-9}
D$^{3}$ToM-t & \textbf{43.00} & 72.48 & \textbf{73.48} & \textbf{20.07} & \textbf{65.45} & \textbf{72.10} & \textbf{63.60} & \textbf{98.12\%} \\
D$^{3}$ToM-t$^{\dagger}$ & 42.89 & \textbf{72.53} & 71.97 & 18.65 & \textbf{66.06} & 65.40 & 62.40 & 95.33\% \\
\midrule
\multicolumn{9}{c}{\textbf{Retain 10\% Tokens}} \\
\cmidrule(lr){1-9}
D$^{3}$ToM-t & \textbf{42.89} & \textbf{72.58} & \textbf{73.48} & 18.68 & \textbf{65.67} & \textbf{70.20} & \textbf{63.40} & \textbf{96.75\%} \\
D$^{3}$ToM-t$^{\dagger}$ & 42.78 & 72.53 & 71.21 & \textbf{18.79} & 65.48 & 66.90 & 61.80 & 95.30\% \\
\bottomrule
\end{tabular}
\caption{Impact of the linear merge-schedule direction.  
D$^{3}$ToM-t adopts a low\,$\rightarrow$\,high retention curve, whereas D$^{3}$ToM-t$^{\dagger}$ reverses it.  
All values are placeholders to be filled with measured accuracies; FLOPs are identical by construction and are therefore omitted.}
\label{tab:ablation_schedule}
\end{table*}

\section{Compatibility with Flash Attention}
\label{app:flash_compat}

Our D\textsuperscript{3}ToM method is designed for full compatibility with memory-efficient attention mechanisms such as FlashAttention. The primary challenge is to obtain the decider-guided importance scores for every visual token without materializing the full $N\times N$ attention matrix. Recall that the importance score for a visual token $j$, denoted $S^{(t)}_j$, is the sum of attention it receives from all decider tokens in the set $\mathcal{D}^{(t)}$, defined as:
\begin{equation}
    S^{(t)}_j = \sum_{d_i\in\mathcal{D}^{(t)}} A_{i,j}.
\end{equation}
To compute these scores efficiently, we propose a **dual-pass extraction** scheme. This method involves two sequential FlashAttention calls on the merge layer $l^{\ast}$, both sharing the same key matrix $K$.

\subsection{Pass 1: Standard Attention Computation}
The first pass executes a standard FlashAttention operation using the original query ($Q$), key ($K$), and value ($V$) matrices. This computes the standard output hidden states, $\mathcal{H}'$, which we cache for the subsequent merge step. The operation is as follows:
\begin{equation}
    \mathcal{H}' = \text{FlashAttention}(Q, K, V).
\end{equation}

\subsection{Pass 2: Decider-Guided Score Extraction}
The second pass is designed specifically to extract the importance scores. We first construct a new query matrix, $Q^{\dagger}$, composed exclusively of the hidden states of the decider tokens:
\begin{equation}
    Q^{\dagger} = Q[\mathcal{D}^{(t)},:] \in\mathbb{R}^{|\mathcal{D}^{(t)}|\times d_{\text{model}}}.
\end{equation}
Next, we create a modified value matrix, $V^{\dagger}$, where every row is set to the first basis vector $e_{1}\in\mathbb{R}^{d_{\text{model}}}$:
\begin{equation}
    \text{row}_{j}(V^{\dagger}) = e_{1}, \quad j=1,\dots,N.
\end{equation}
We then execute FlashAttention with the modified query and value matrices while reusing the original key matrix $K$. This yields an output tensor $O^{\dagger}$:
\begin{equation}
    O^{\dagger} = \text{FlashAttention}(Q^{\dagger}, K, V^{\dagger}) \in\mathbb{R}^{|\mathcal{D}^{(t)}|\times d_{\text{model}}}.
\end{equation}
Because every row in $V^{\dagger}$ is identical, the first channel of the output $O^{\dagger}$ effectively encodes the attention distribution from each decider token to all other tokens. By aggregating this first channel across all decider outputs, we recover the exact importance score $S^{(t)}_j$ for each visual token $j$ as follows:
\begin{equation}
    S^{(t)}_j = \sum_{d_i\in\mathcal{D}^{(t)}} A_{i,j} = \sum_{i=1}^{|\mathcal{D}^{(t)}|} \bigl(O^{\dagger}_{i}\bigr)_{1,j}.
\end{equation}

\begin{figure*}[h]
  \centering
  \subfigure[Accuracy versus retained tokens]{%
    \includegraphics[width=0.48\linewidth]{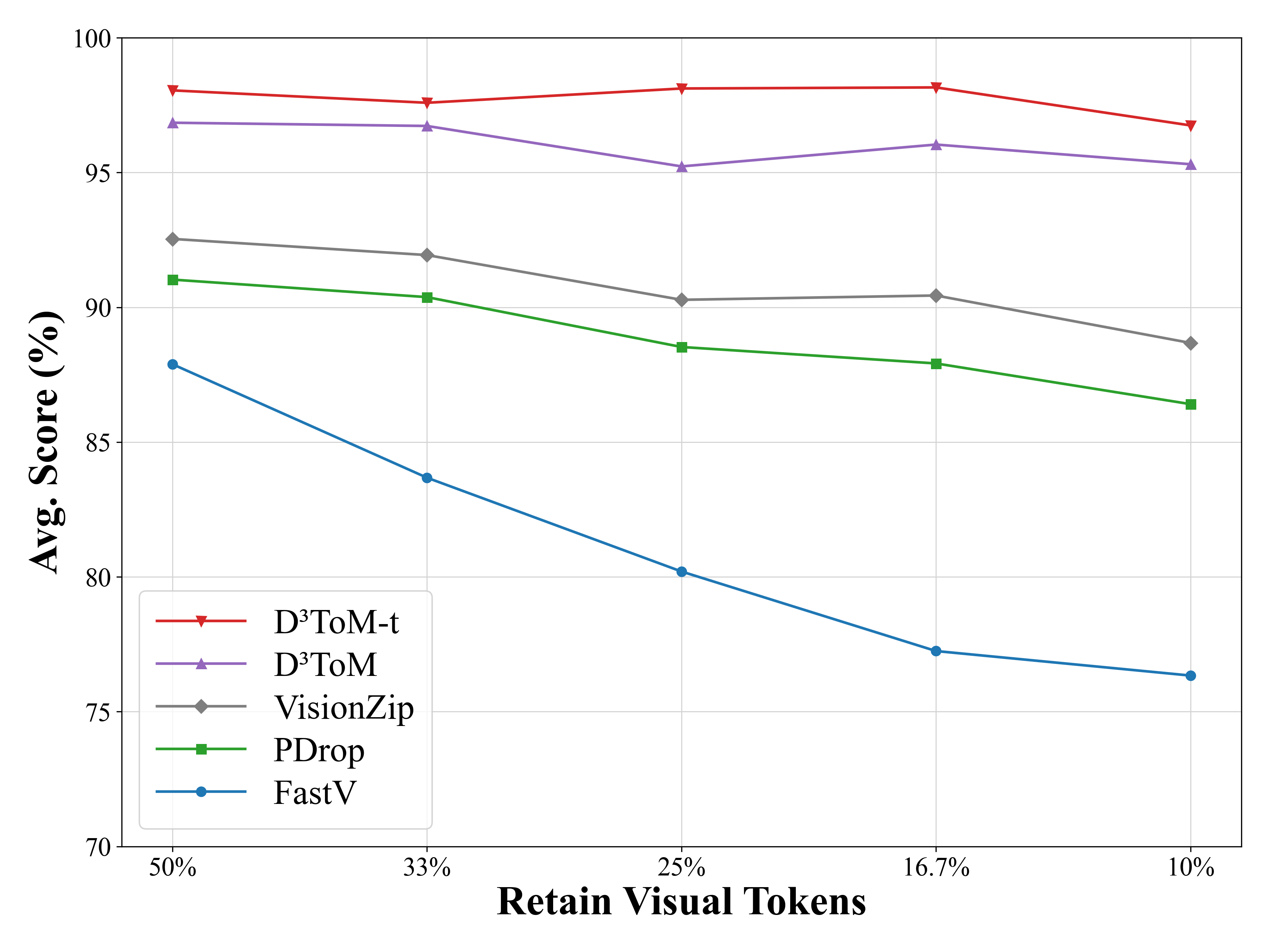}\label{fig:tok_acc}}
  \hfill
  \subfigure[Accuracy versus relative FLOPs]{%
    \includegraphics[width=0.48\linewidth]{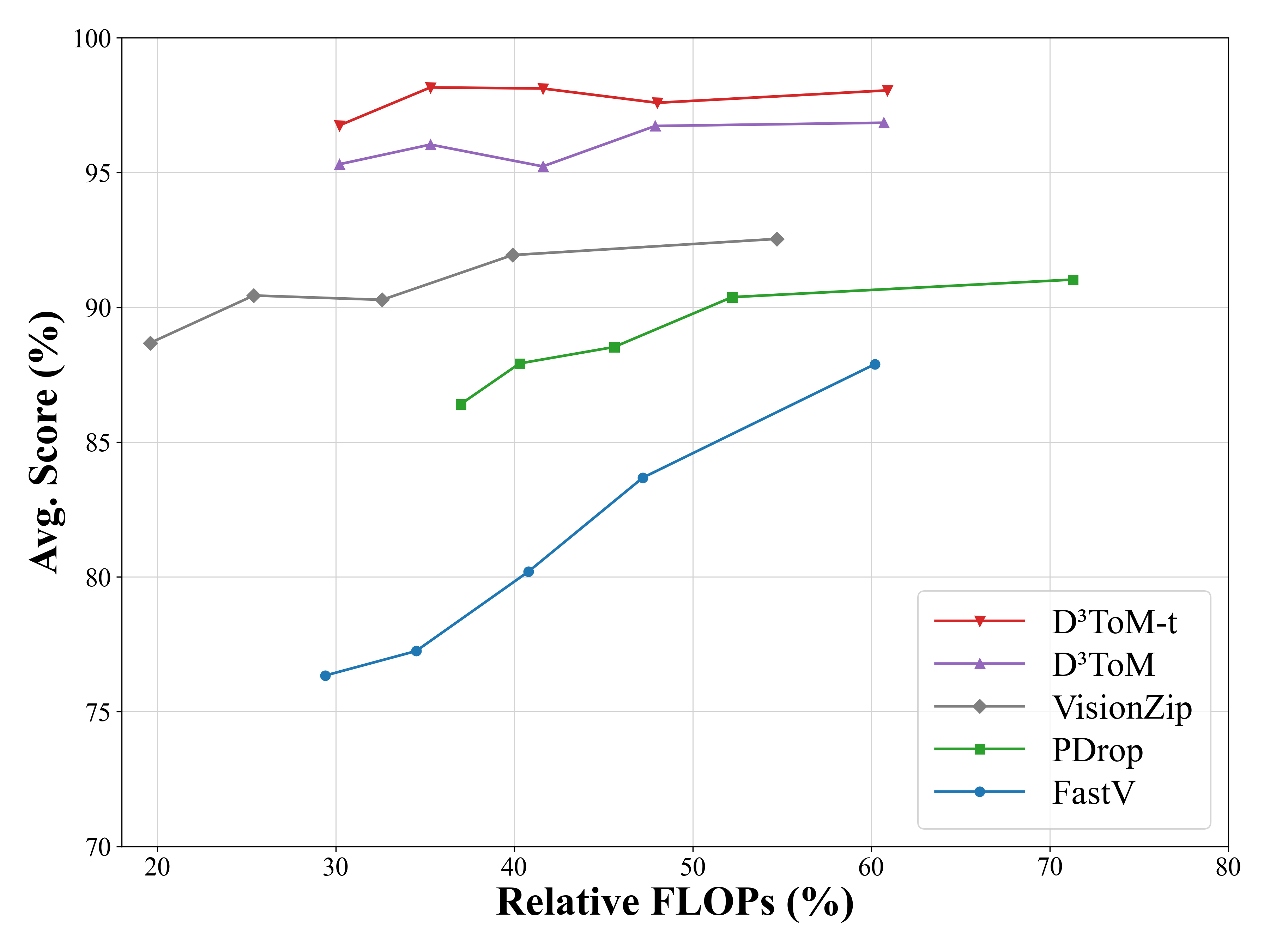}\label{fig:flops_acc}}
  \vskip\baselineskip
  \subfigure[Relative FLOPs versus retained tokens]{%
    \includegraphics[width=0.48\linewidth]{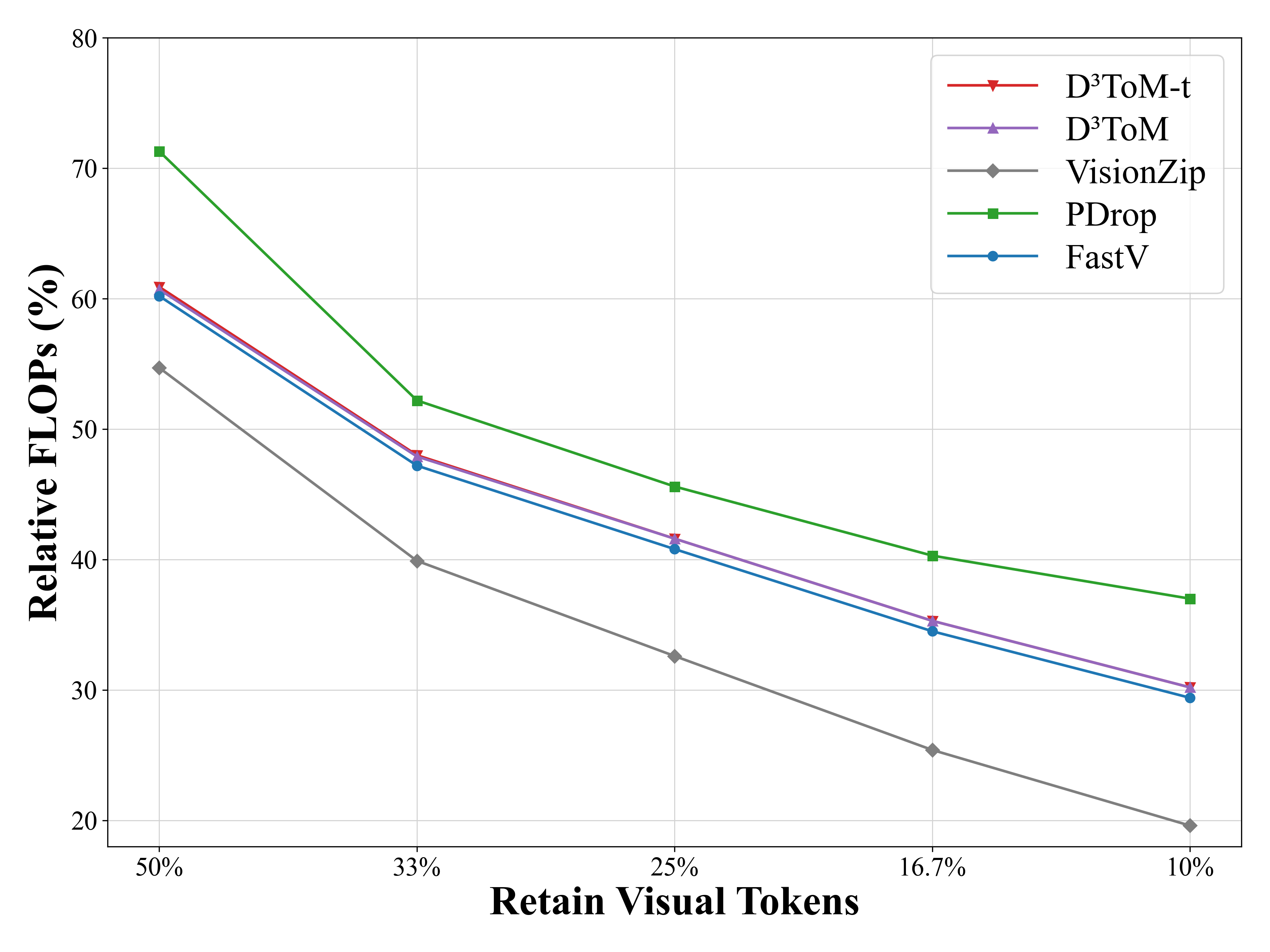}\label{fig:tok_flops}}
  \hfill
  \subfigure[Inference latency versus retained tokens]{%
    \includegraphics[width=0.48\linewidth]{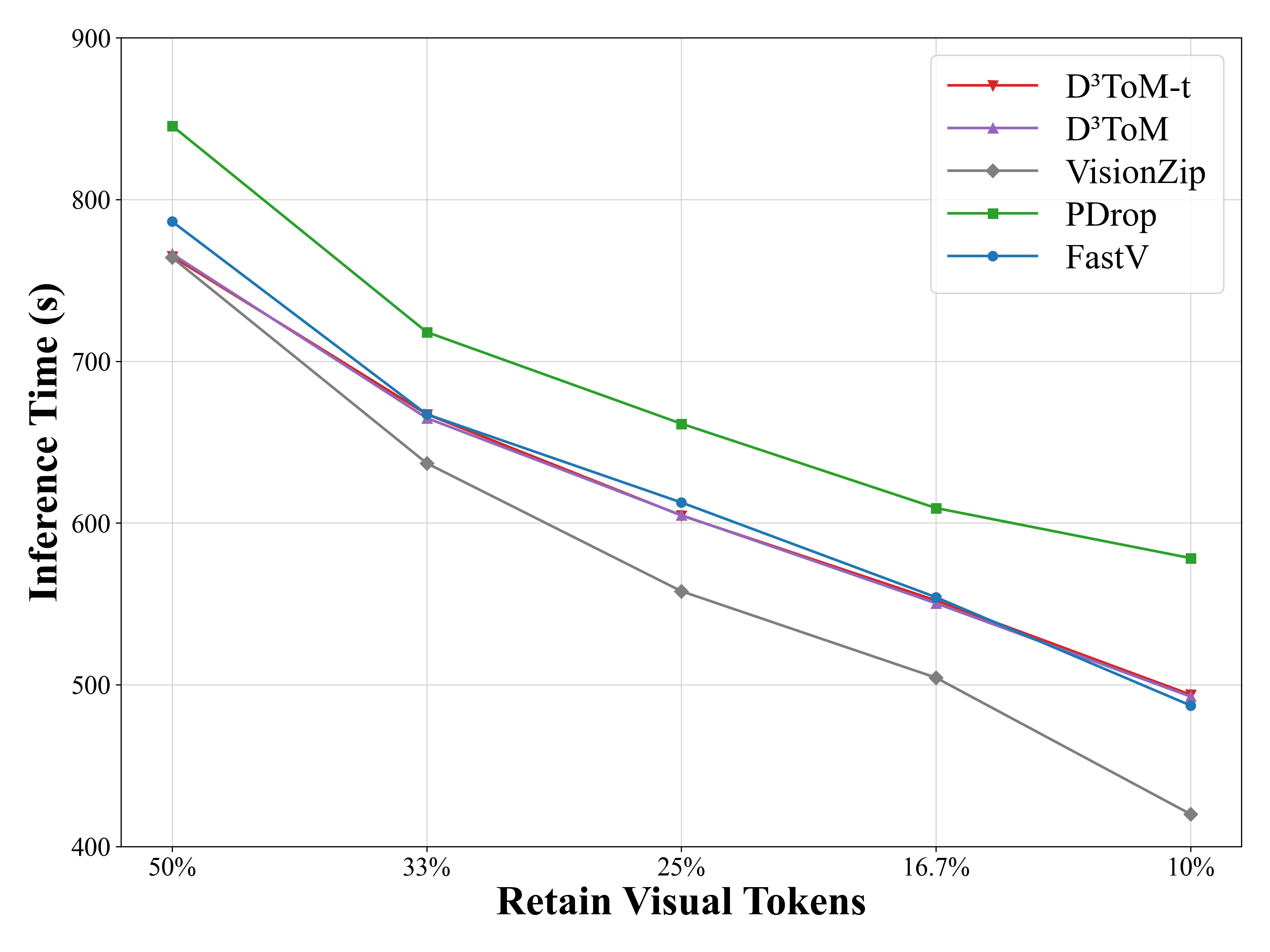}\label{fig:tok_time}}
  \caption{Efficiency plots derived directly from the main-paper tables.  All curves share identical marker, colour, and font conventions for consistency with the quantitative results.}
  \label{fig:extended_eff_plots}
\end{figure*}

\subsection{Token Selection and Merging}
With the importance scores $\{S^{(t)}_j\}$ computed, we perform a top-k selection to partition the visual tokens into a kept set, $\mathcal{V}_{\text{kept}}(t)$, and a merge set, $\mathcal{V}_{\text{merge}}(t)$. The physical merging operation is then applied to the cached hidden states $\mathcal{H}'$ obtained from Pass 1.

\begin{figure*}[h!]
    \centering
    \subfigure[L1, step 8]{\includegraphics[width=0.23\linewidth]{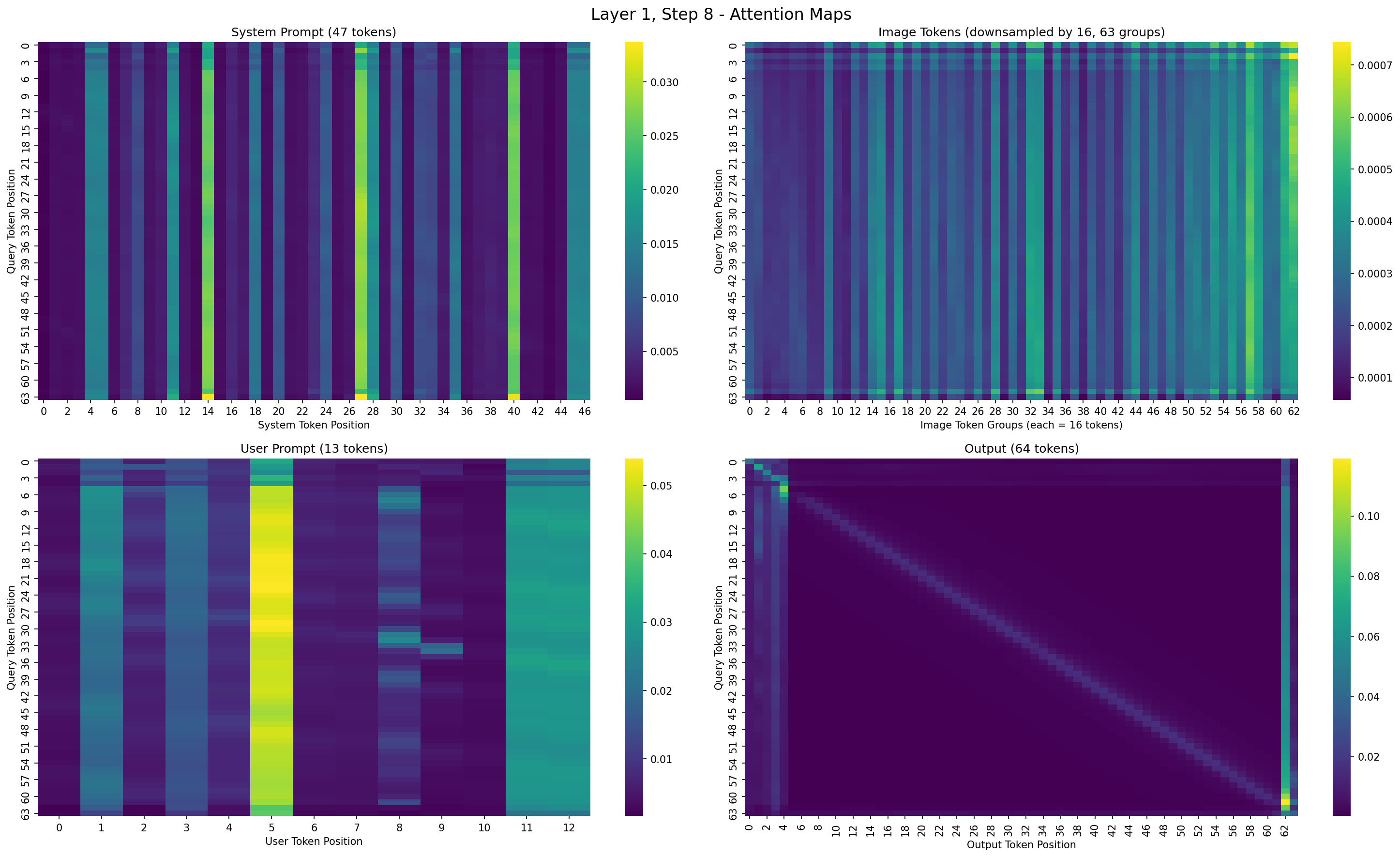}}
    \hfill
    \subfigure[L1, step 16]{\includegraphics[width=0.23\linewidth]{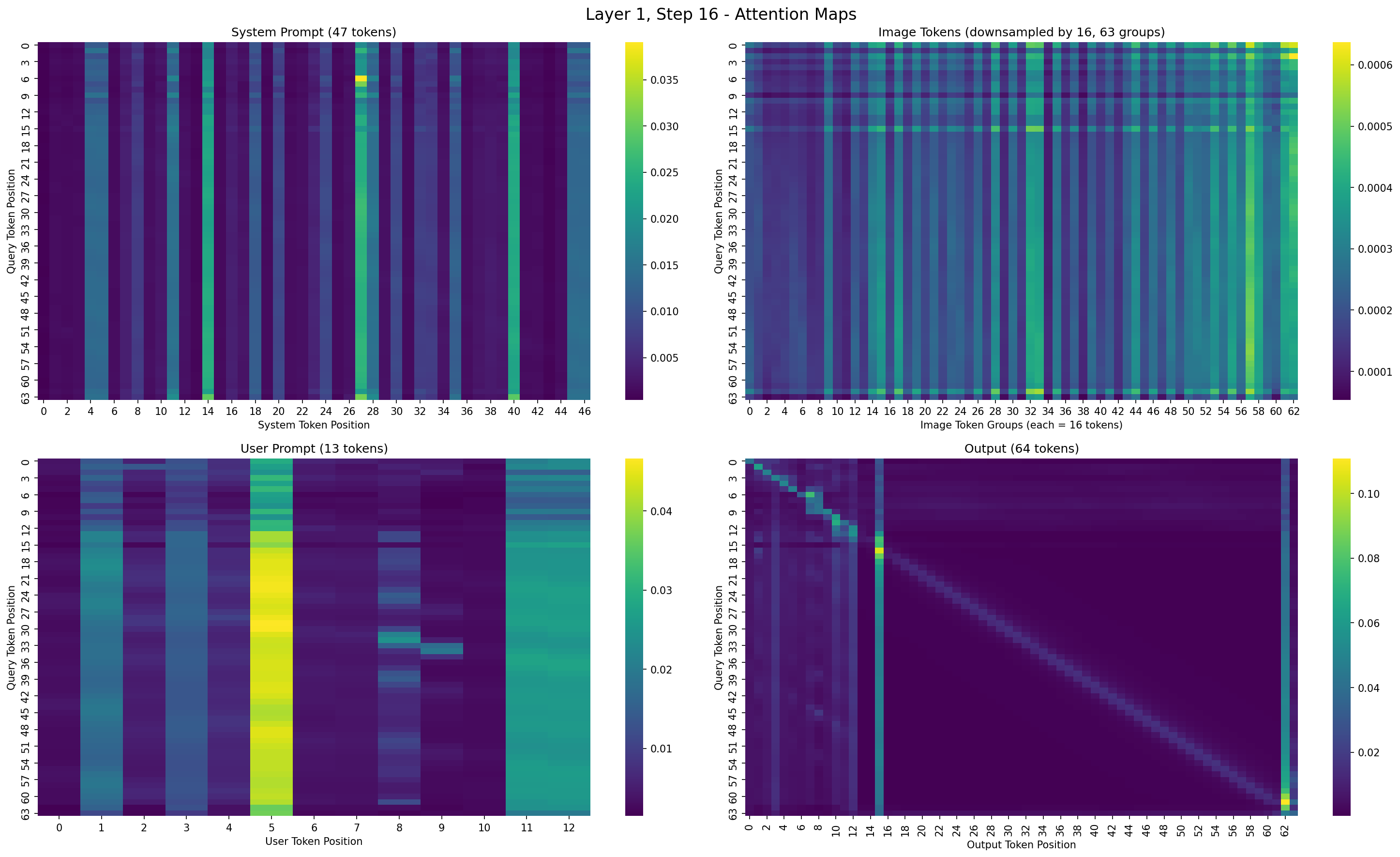}}
    \hfill
    \subfigure[L1, step 24]{\includegraphics[width=0.23\linewidth]{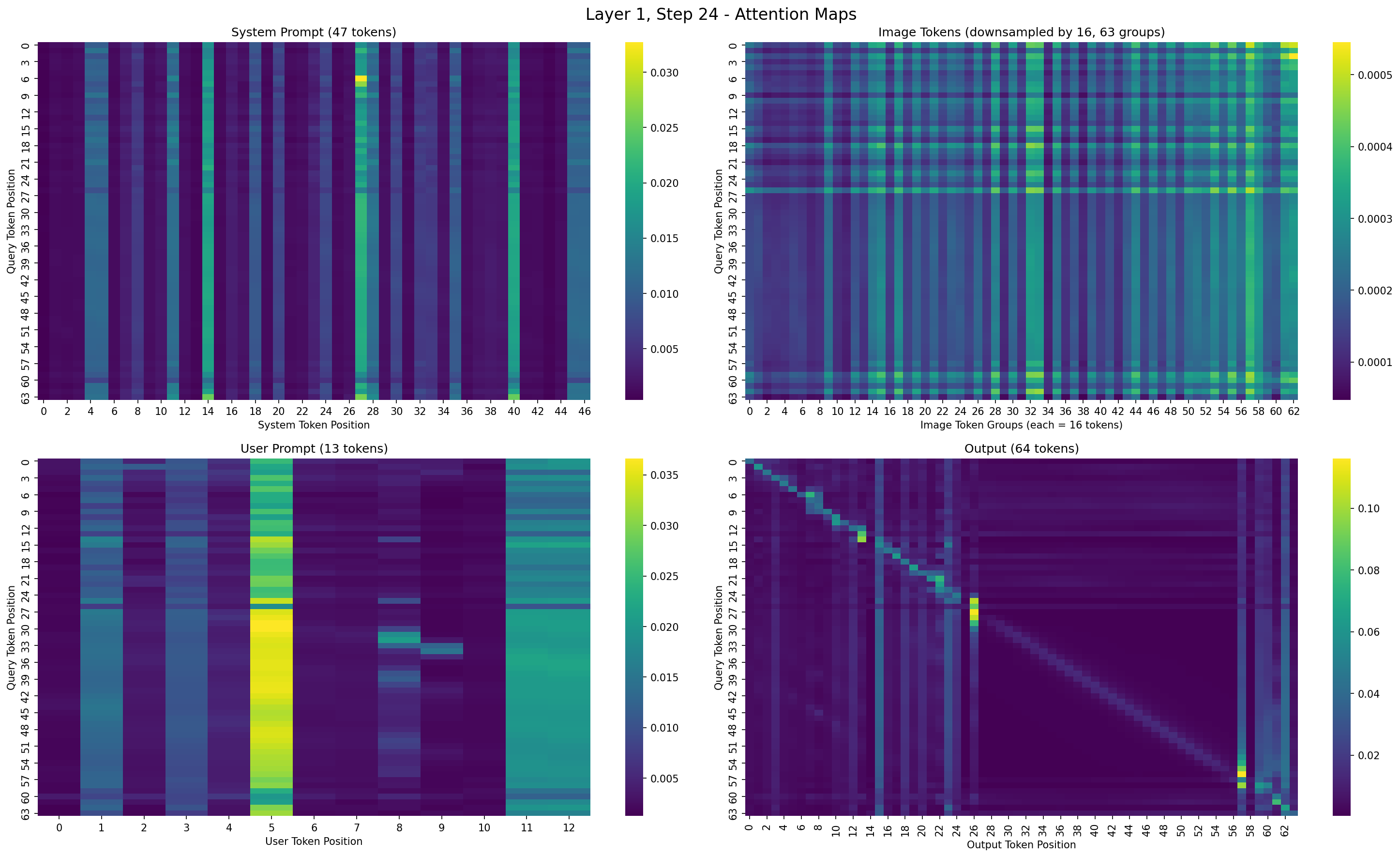}}
    \hfill
    \subfigure[L1, step 32]{\includegraphics[width=0.23\linewidth]{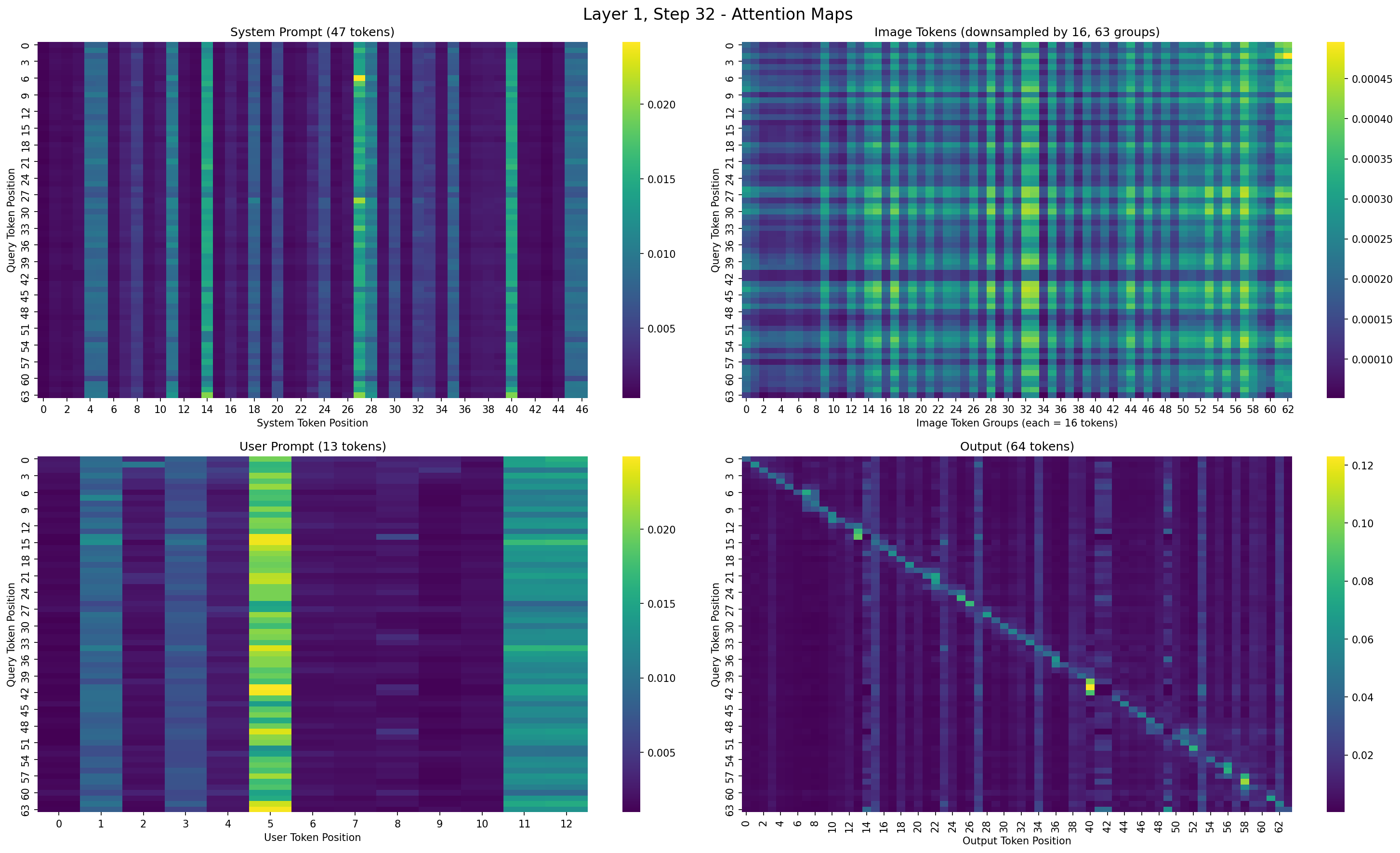}}
    \\[-1.2ex]
    \subfigure[L8, step 8]{\includegraphics[width=0.23\linewidth]{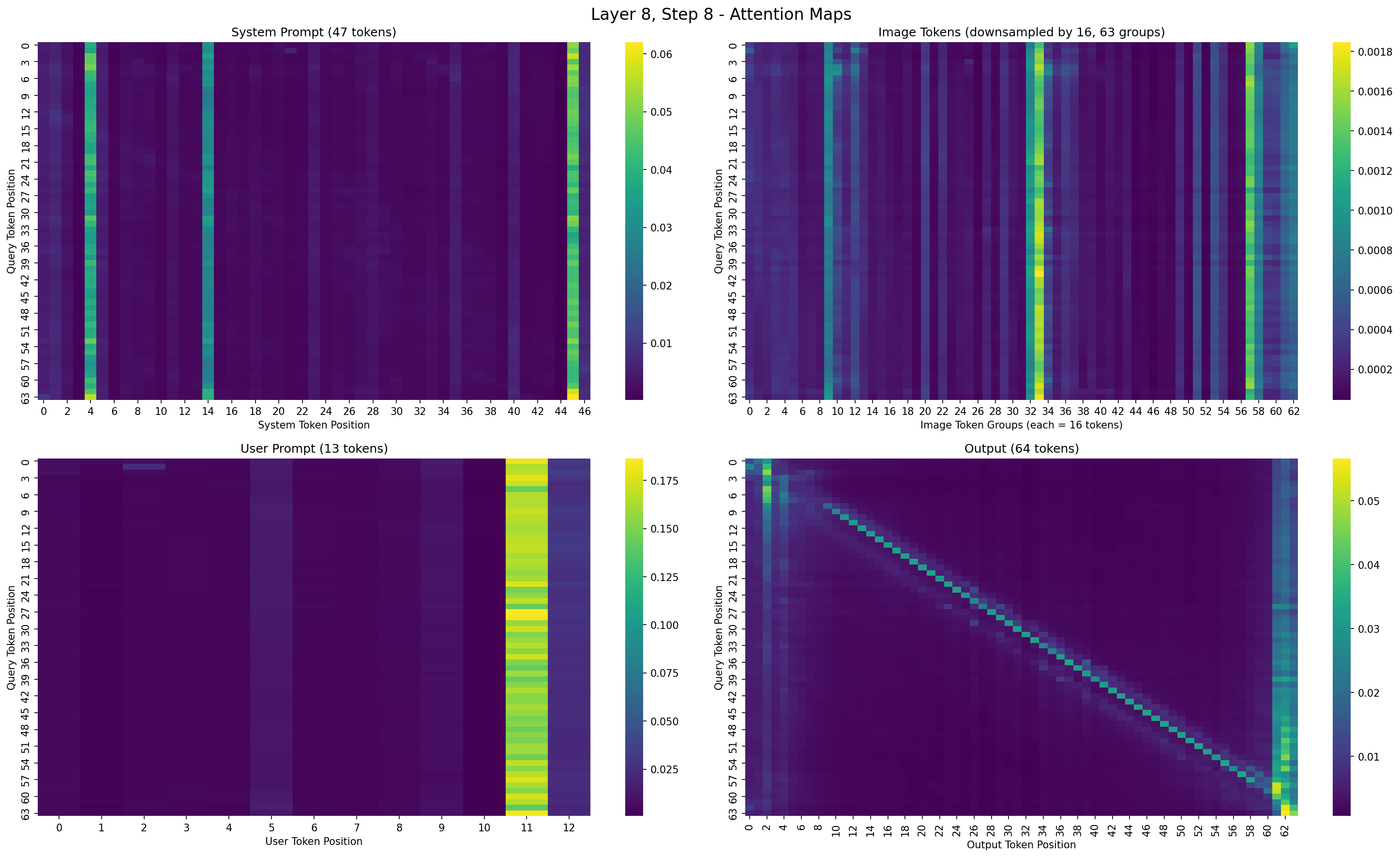}}
    \hfill
    \subfigure[L8, step 16]{\includegraphics[width=0.23\linewidth]{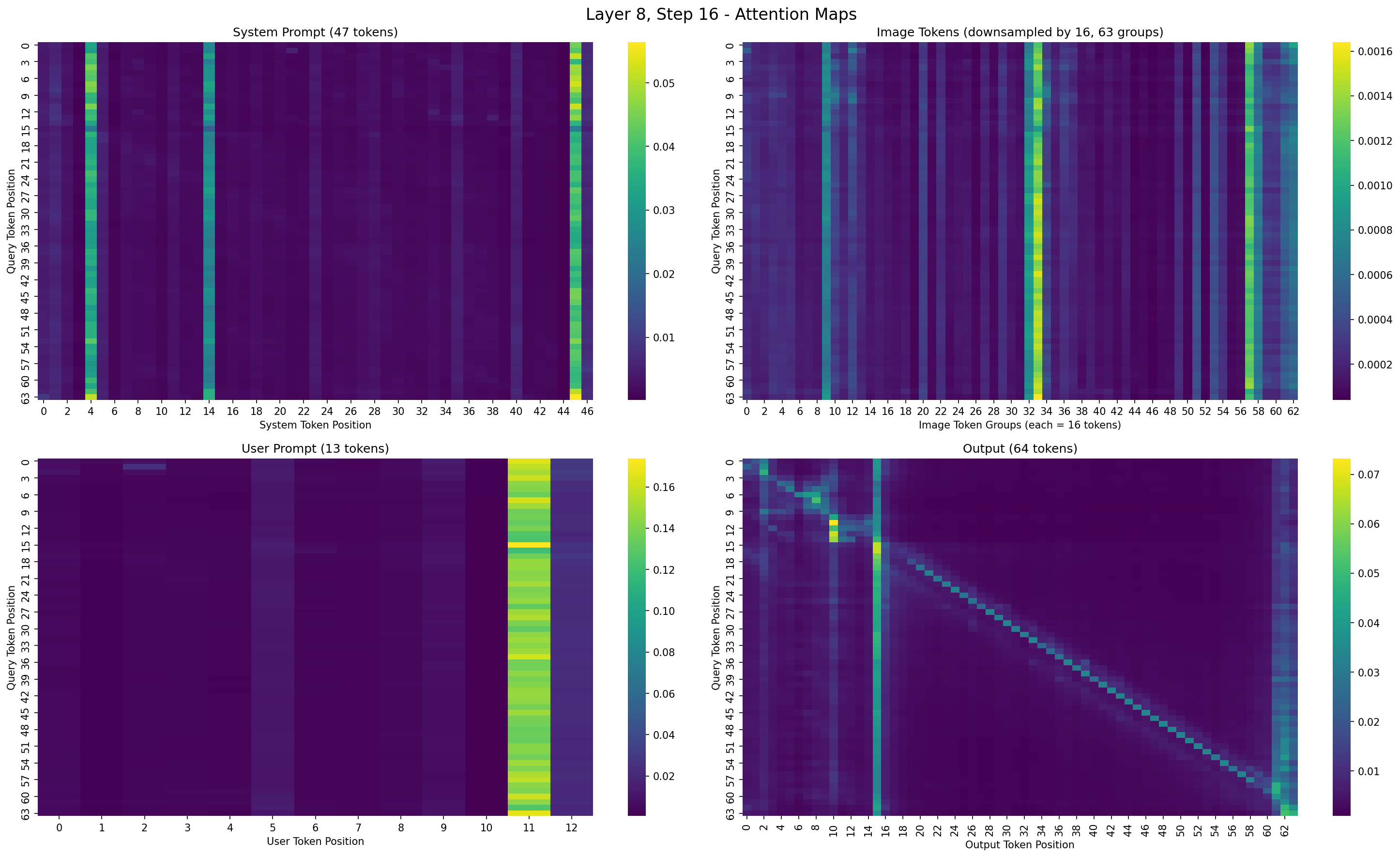}}
    \hfill
    \subfigure[L8, step 24]{\includegraphics[width=0.23\linewidth]{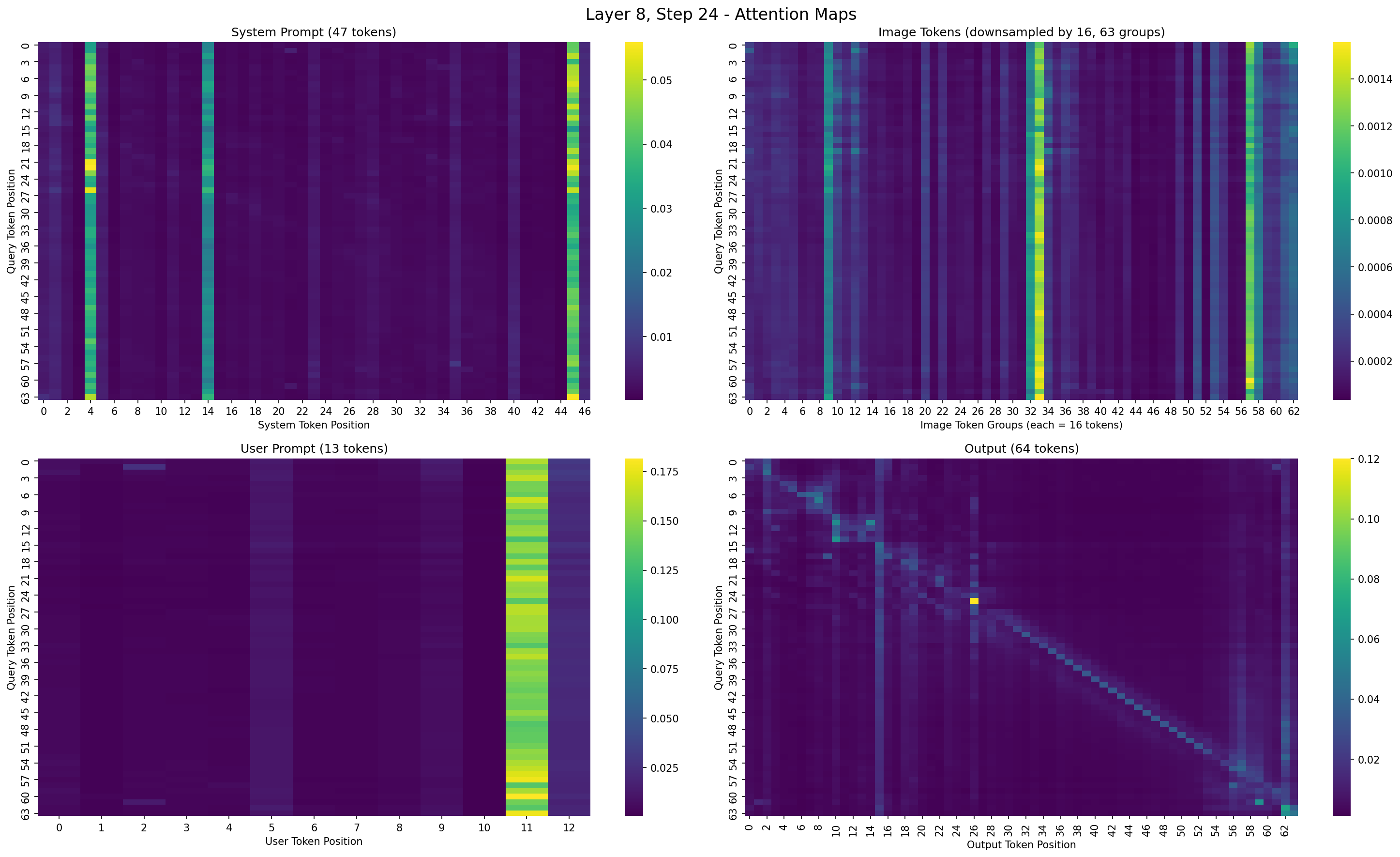}}
    \hfill
    \subfigure[L8, step 32]{\includegraphics[width=0.23\linewidth]{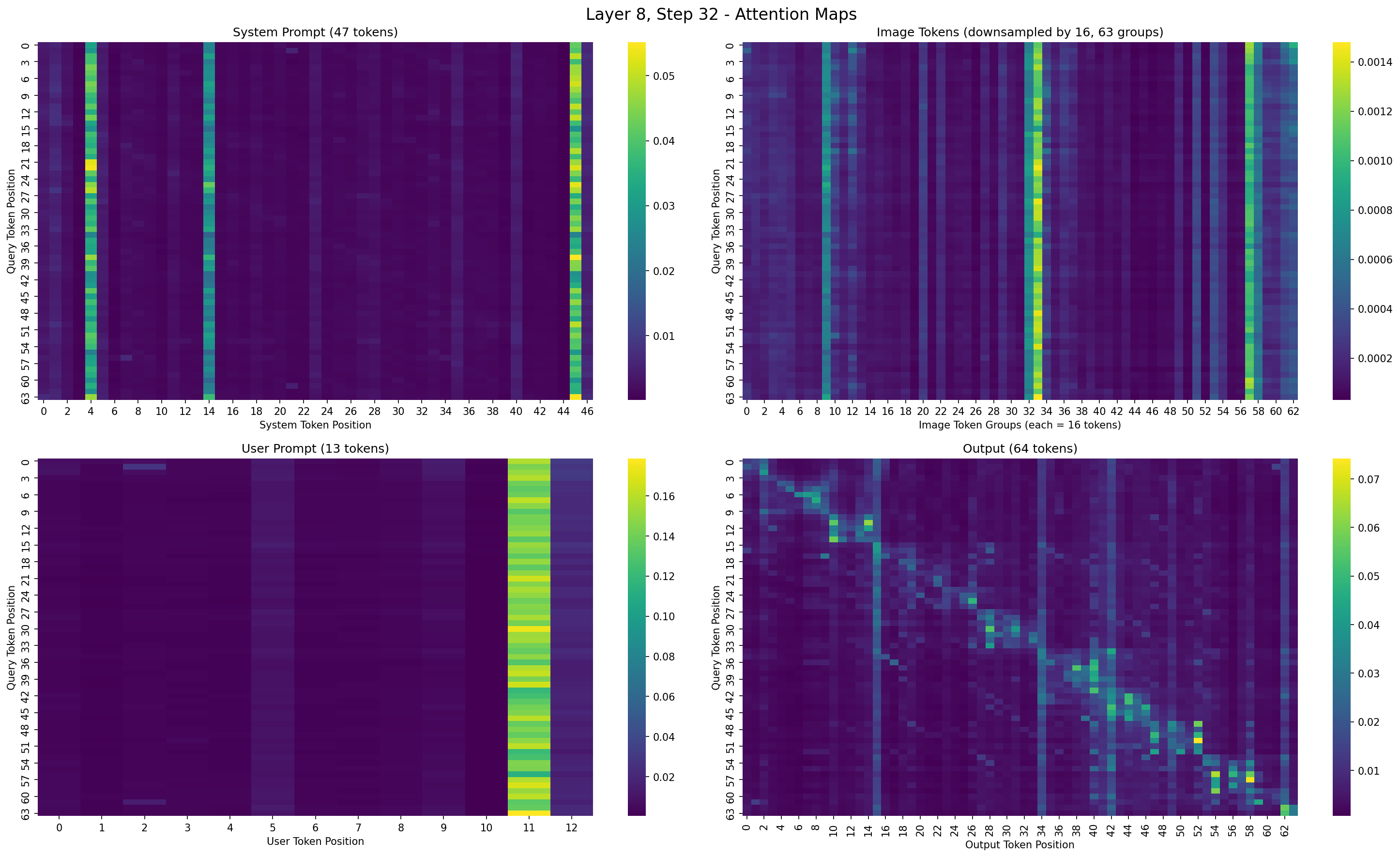}}
    \\[-1.2ex]
    \subfigure[L16, step 8]{\includegraphics[width=0.23\linewidth]{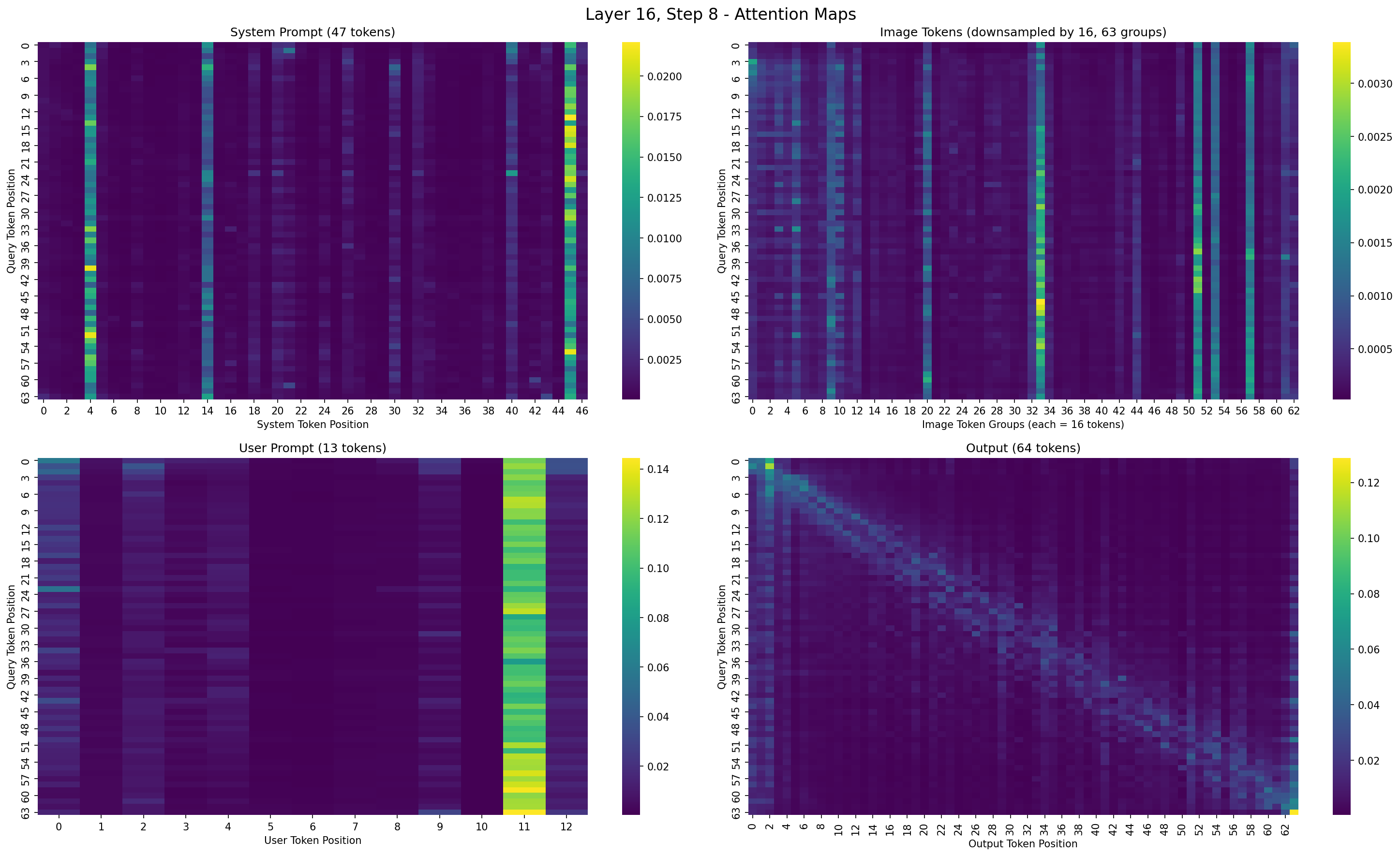}}
    \hfill
    \subfigure[L16, step 16]{\includegraphics[width=0.23\linewidth]{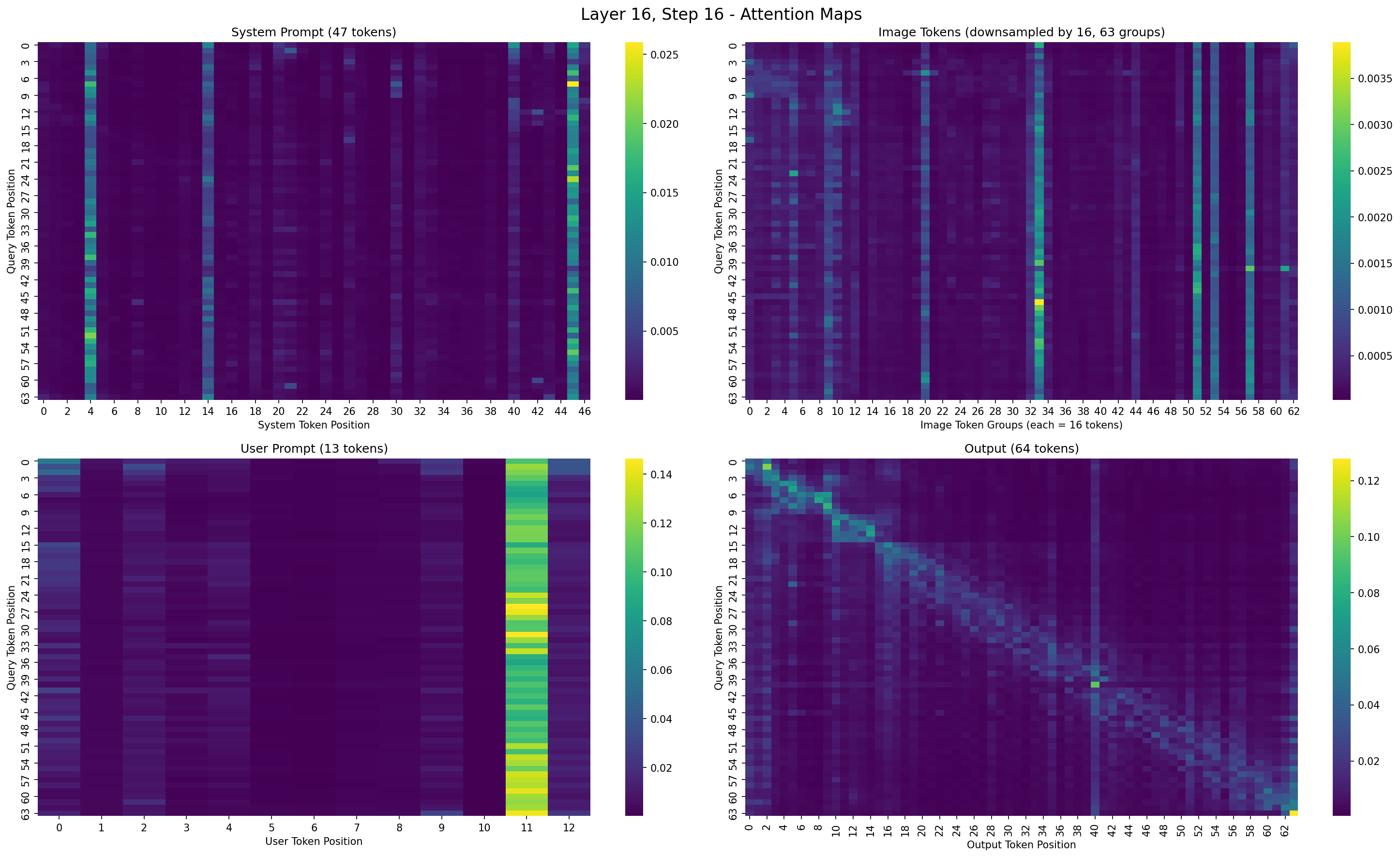}}
    \hfill
    \subfigure[L16, step 24]{\includegraphics[width=0.23\linewidth]{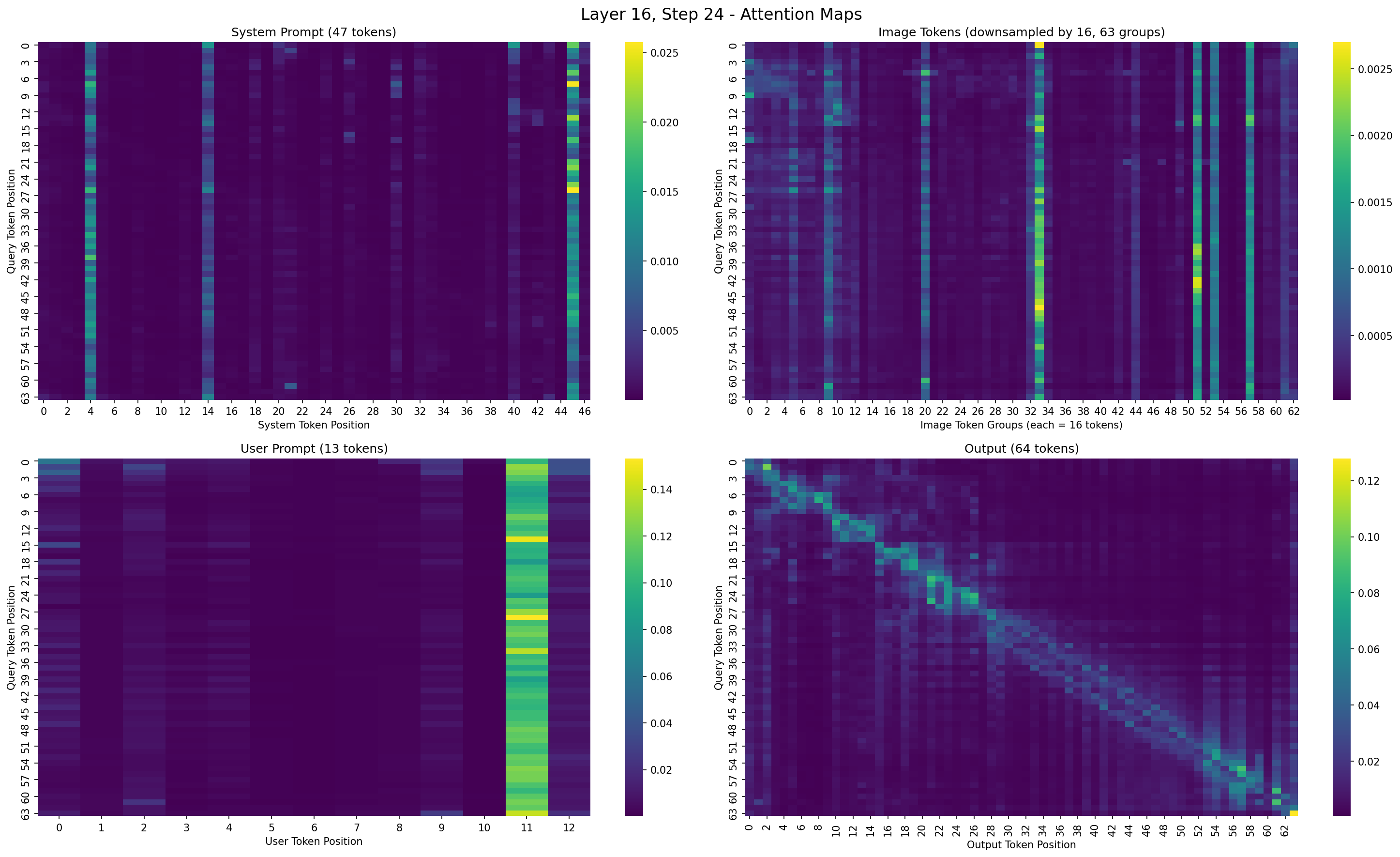}}
    \hfill
    \subfigure[L16, step 32]{\includegraphics[width=0.23\linewidth]{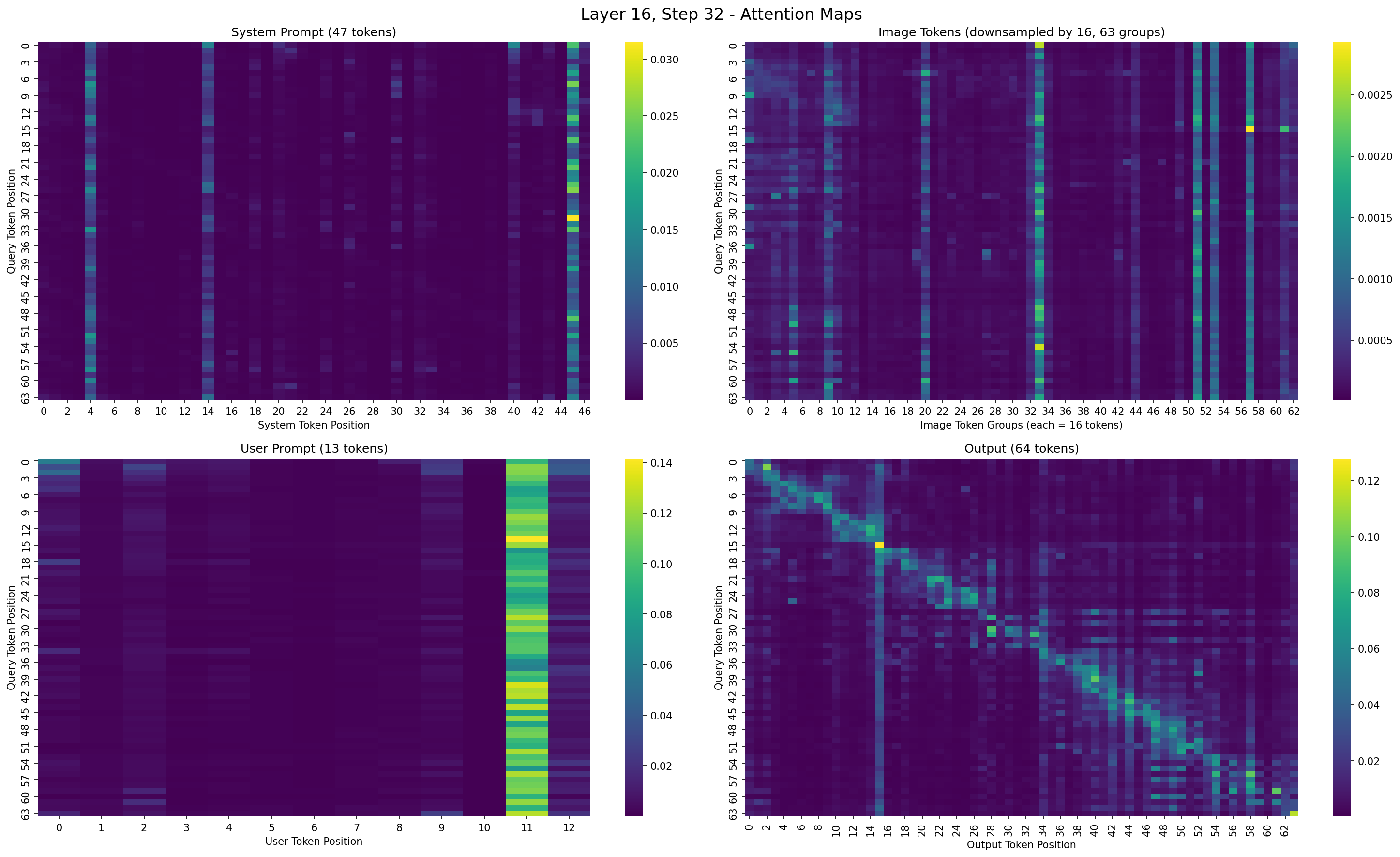}}
    \\[-1.2ex]
    \subfigure[L24, step 8]{\includegraphics[width=0.23\linewidth]{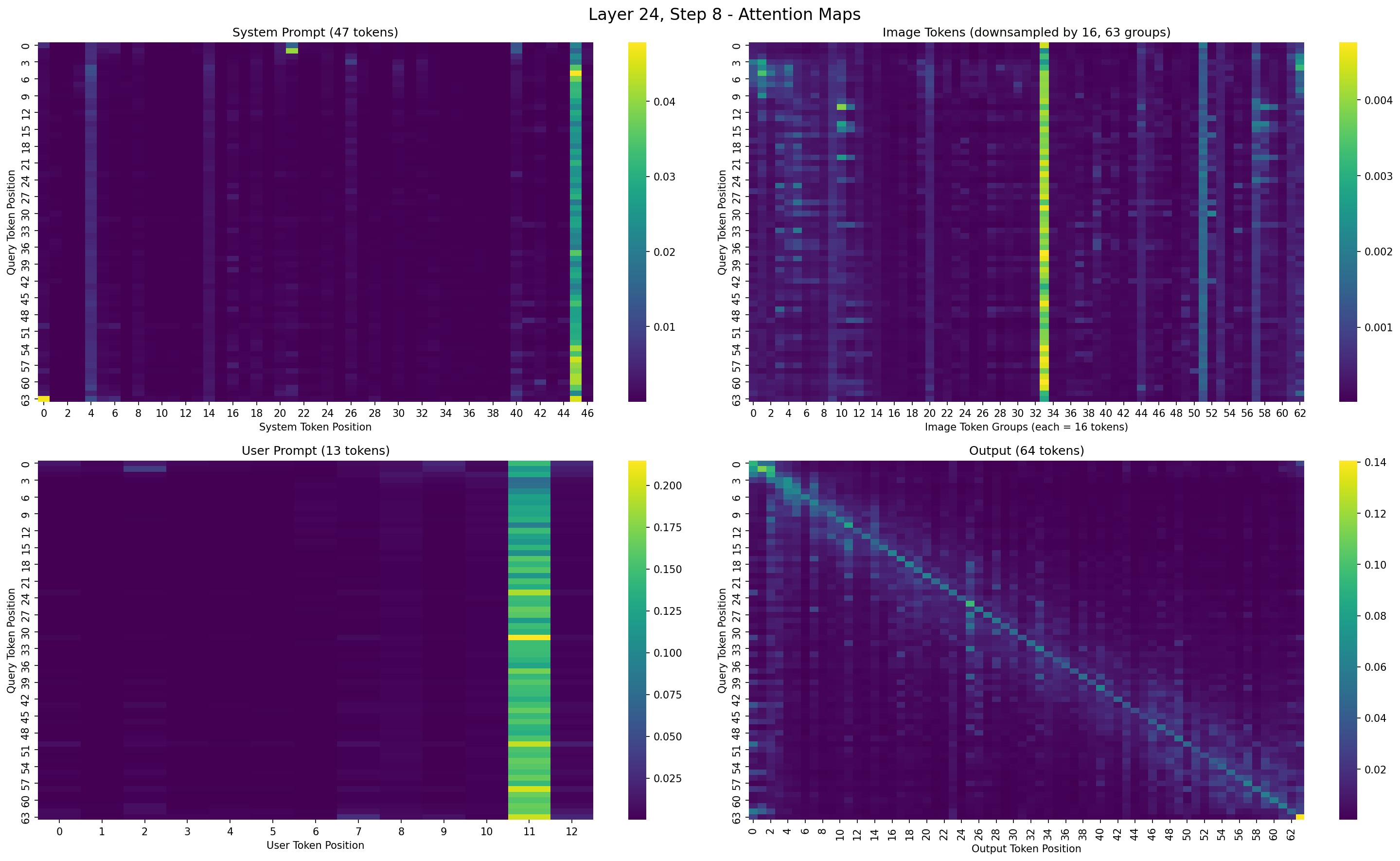}}
    \hfill
    \subfigure[L24, step 16]{\includegraphics[width=0.23\linewidth]{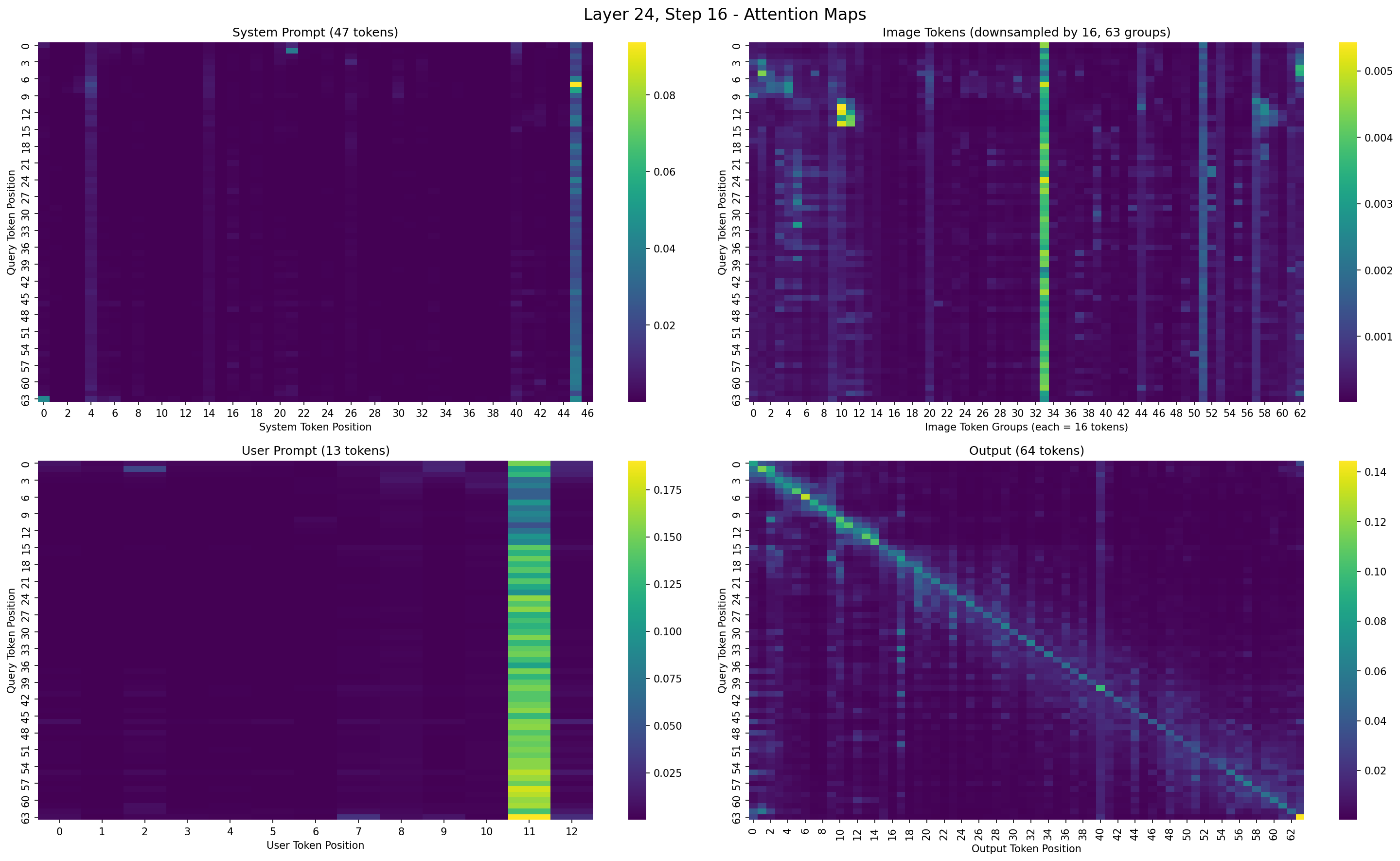}}
    \hfill
    \subfigure[L24, step 24]{\includegraphics[width=0.23\linewidth]{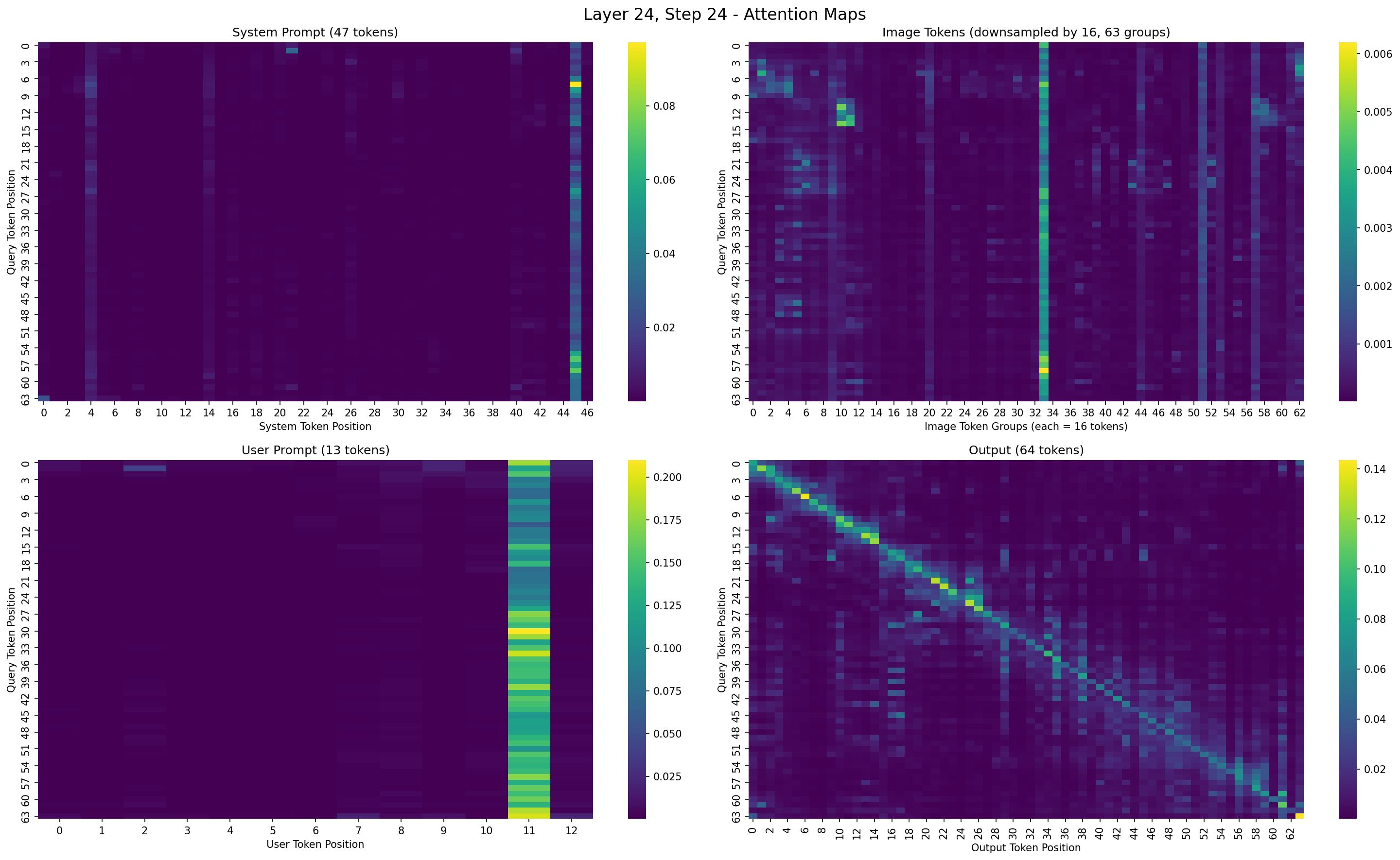}}
    \hfill
    \subfigure[L24, step 32]{\includegraphics[width=0.23\linewidth]{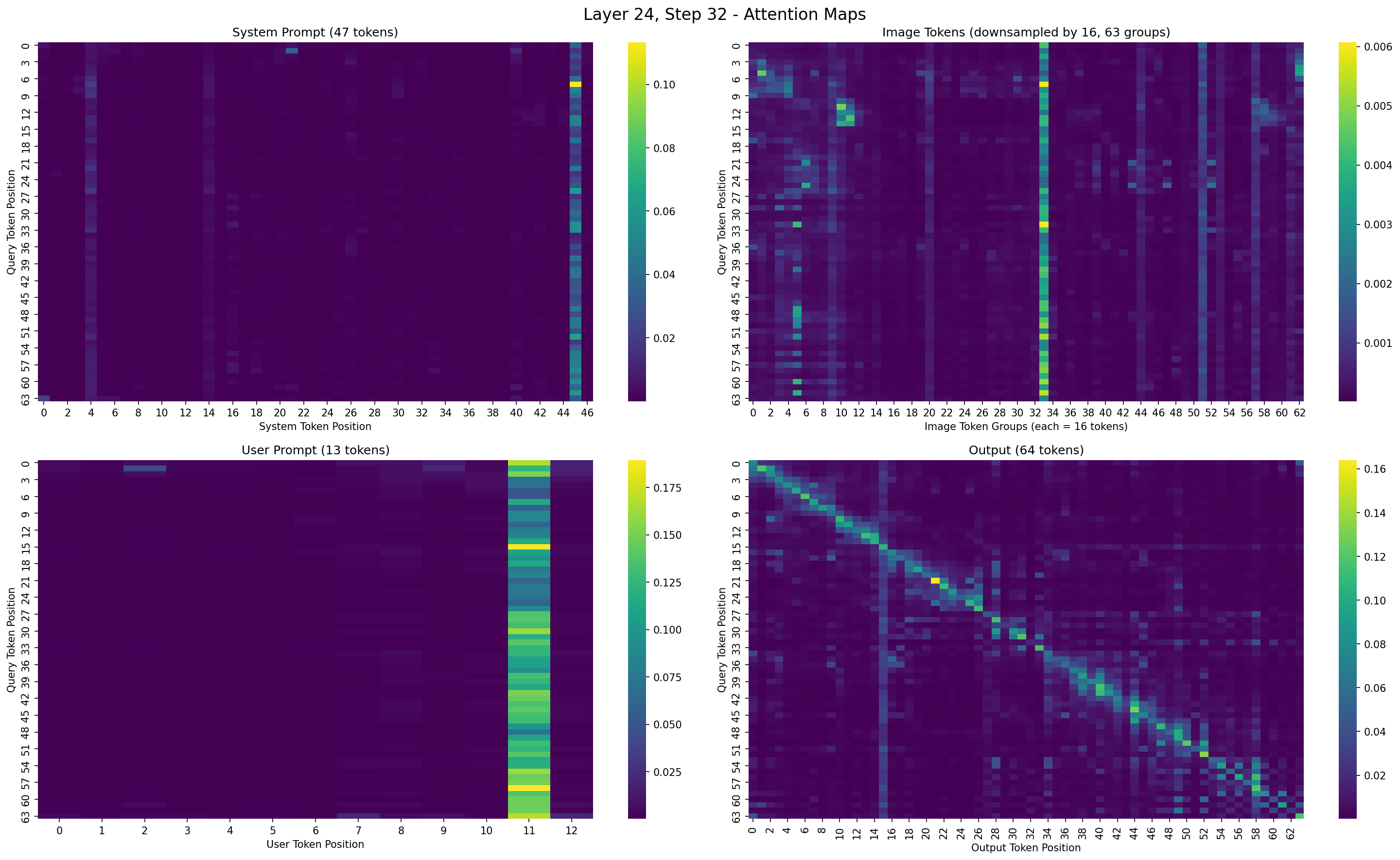}}
    \\[-1.2ex]
    \subfigure[L32, step 8]{\includegraphics[width=0.23\linewidth]{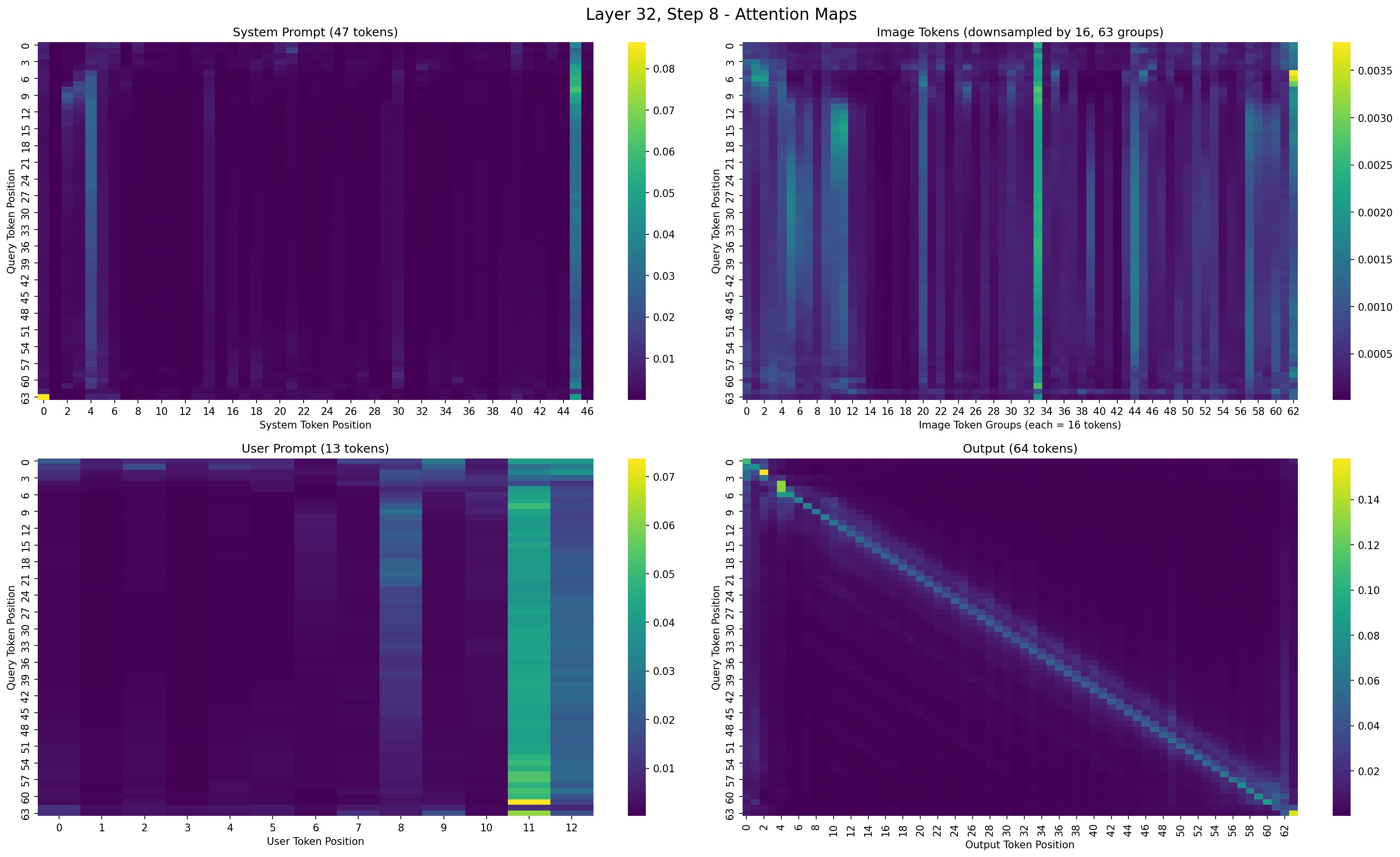}}
    \hfill
    \subfigure[L32, step 16]{\includegraphics[width=0.23\linewidth]{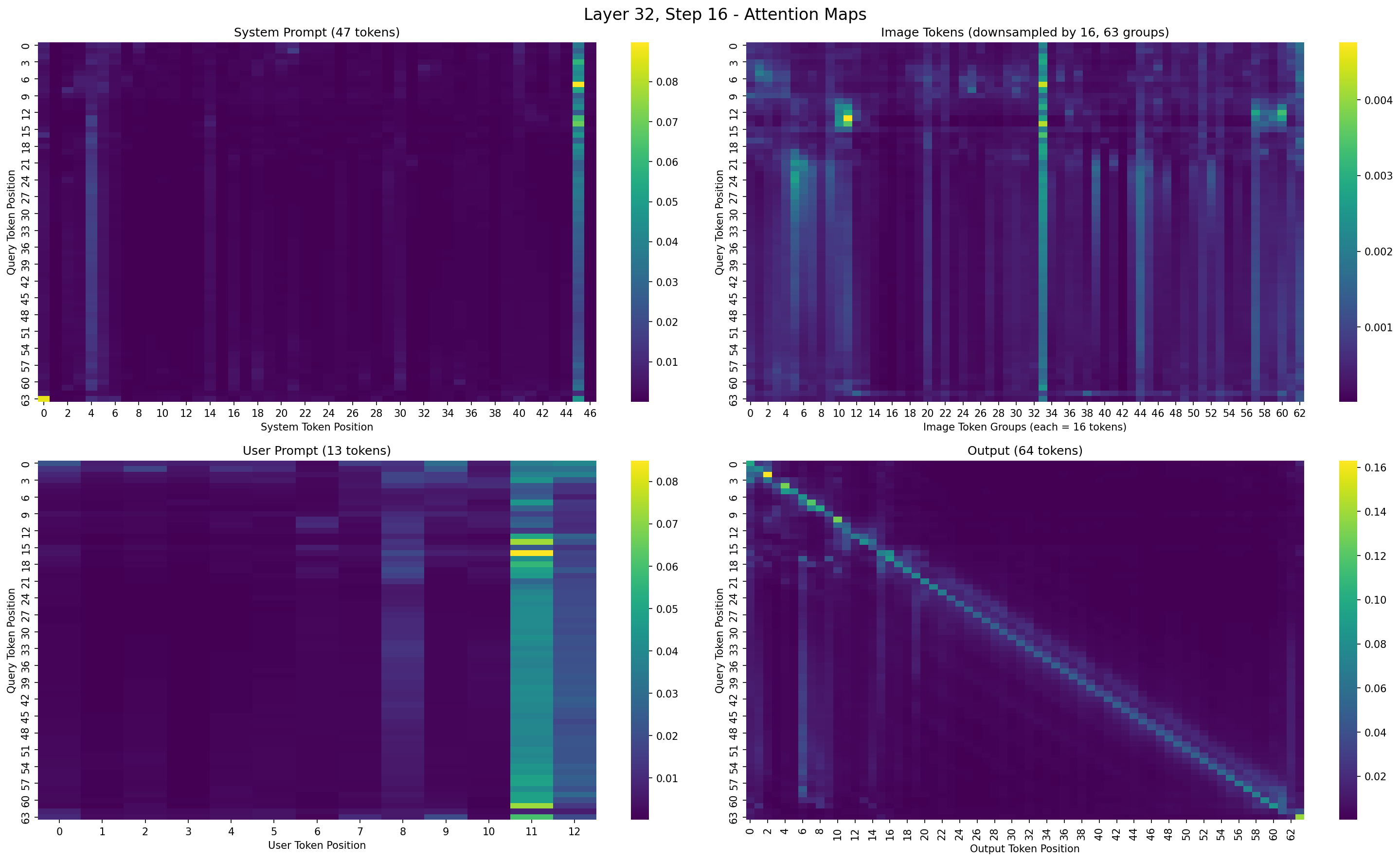}}
    \hfill
    \subfigure[L32, step 24]{\includegraphics[width=0.23\linewidth]{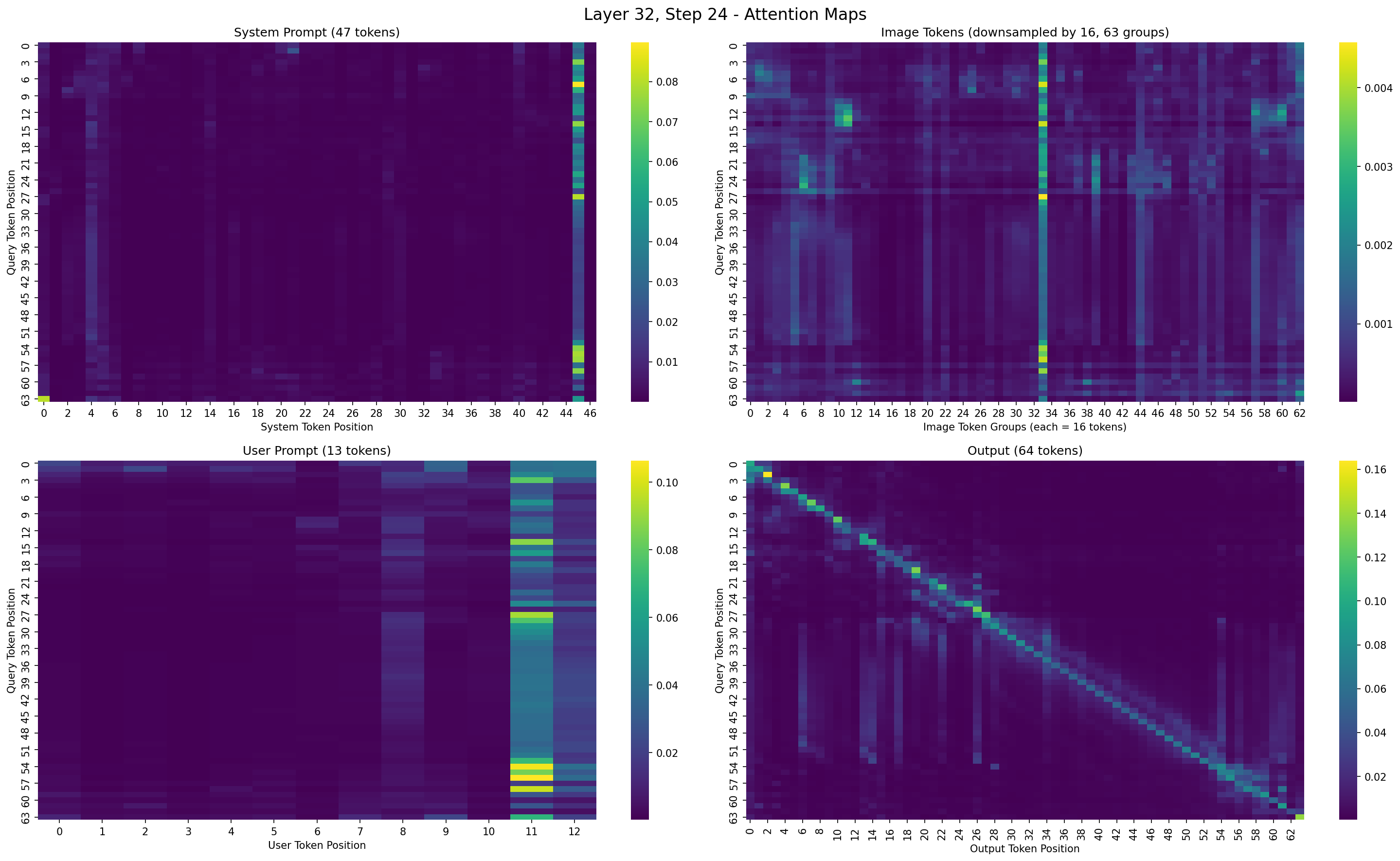}}
    \hfill
    \subfigure[L32, step 32]{\includegraphics[width=0.23\linewidth]{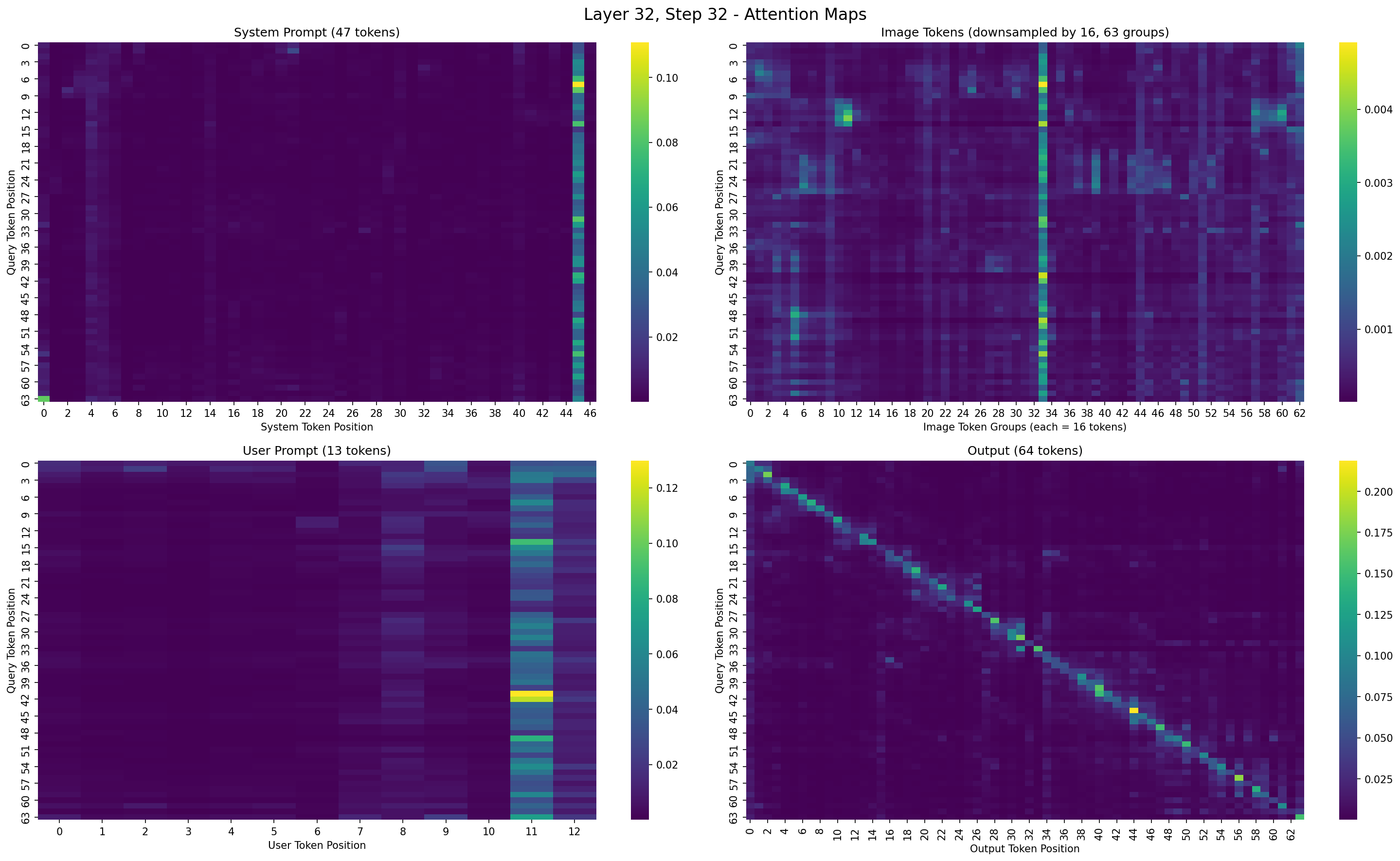}}
    \caption{Query–key attention matrices sampled across depth and time.  
    Dark colours indicate low weights.  
    From left to right the temporal focus sharpens; from top to bottom the spatial focus narrows.  
    The intersection of these trends visualises how the two-level dynamic merge schedule first preserves global context then concentrates computation on the salient image groups identified by the decider tokens.}
    \label{fig:attn_dynamics_grid}
\end{figure*}

\begin{figure*}[h!]
    \centering
    \subfigure[Step 1]{\includegraphics[width=0.115\textwidth]{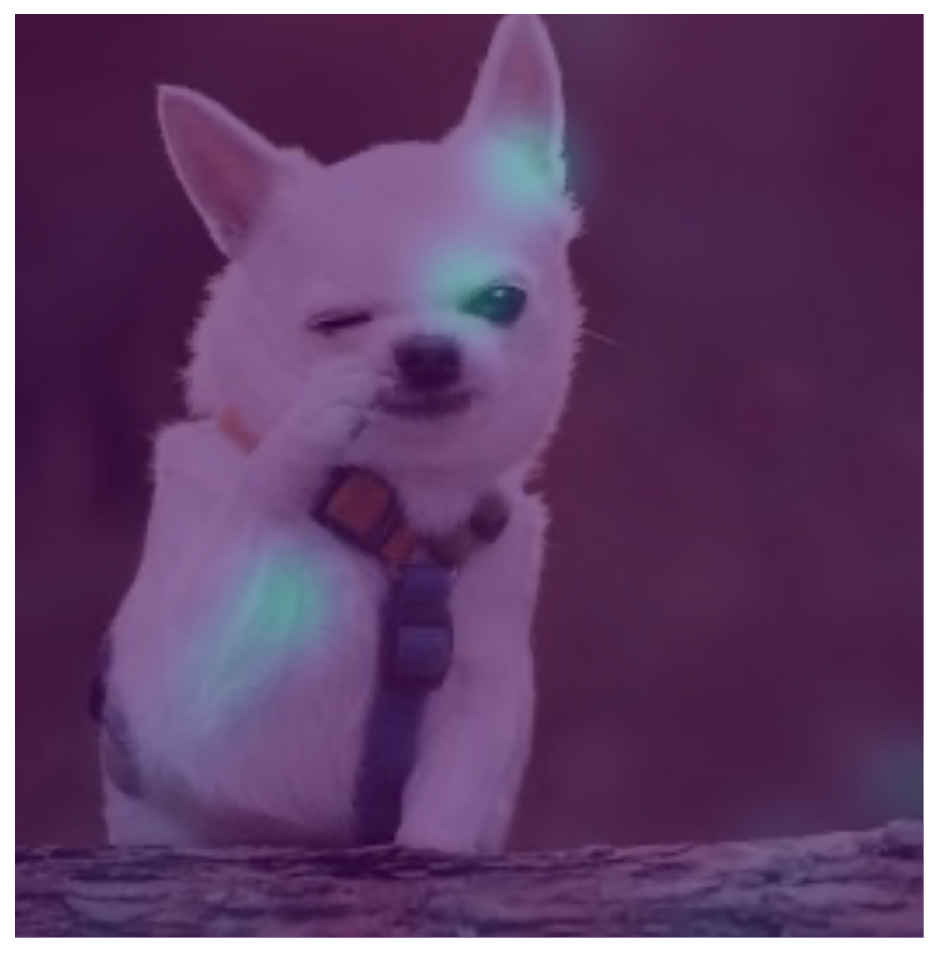}}\hfill
    \subfigure[Step 2]{\includegraphics[width=0.115\textwidth]{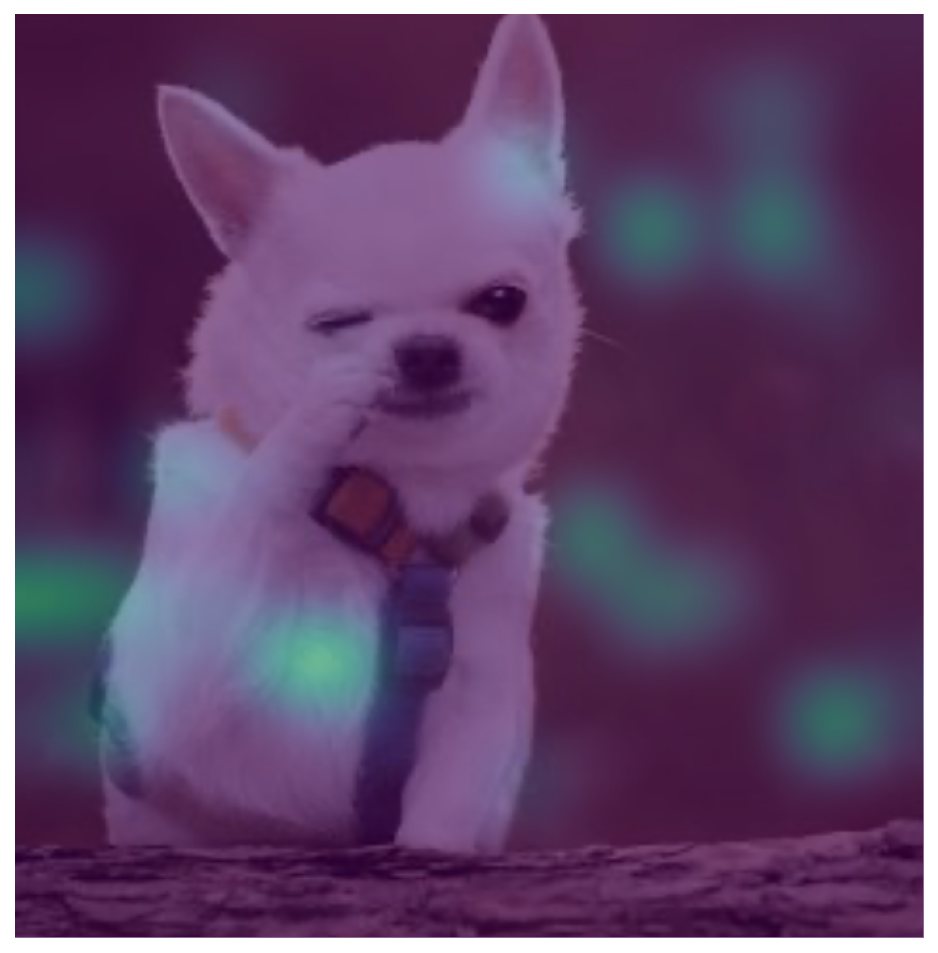}}\hfill
    \subfigure[Step 3]{\includegraphics[width=0.115\textwidth]{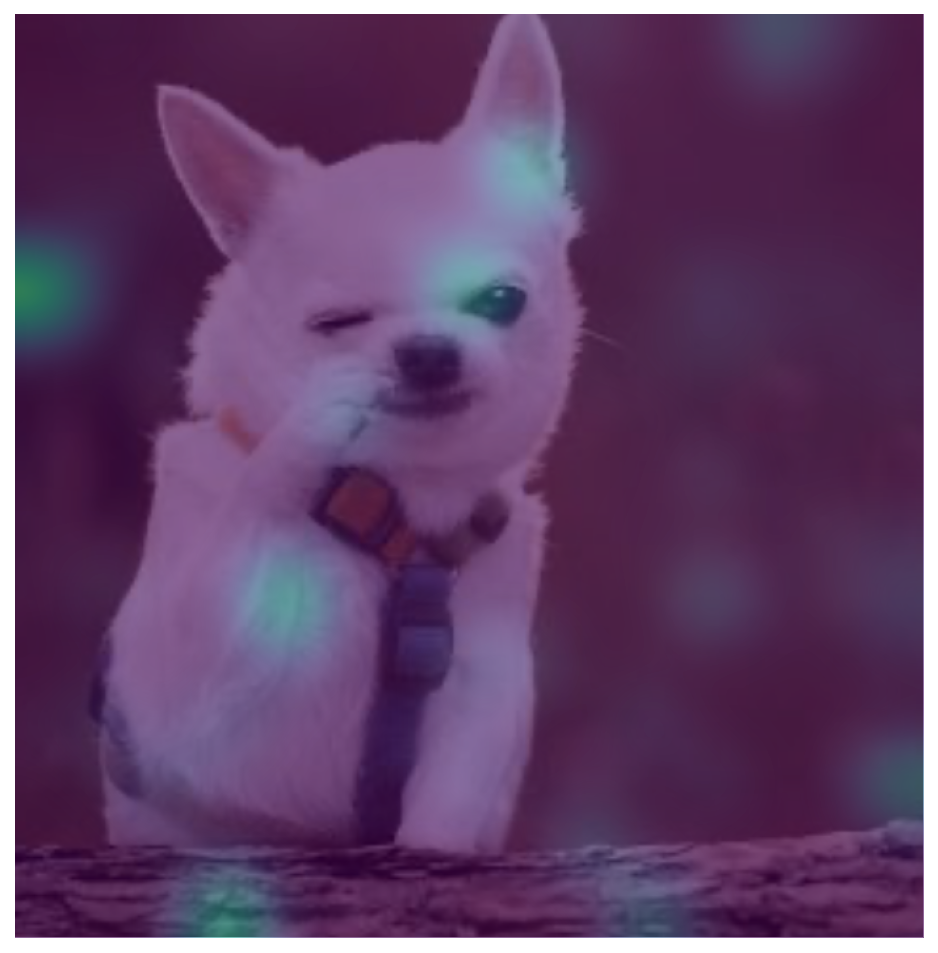}}\hfill
    \subfigure[Step 4]{\includegraphics[width=0.115\textwidth]{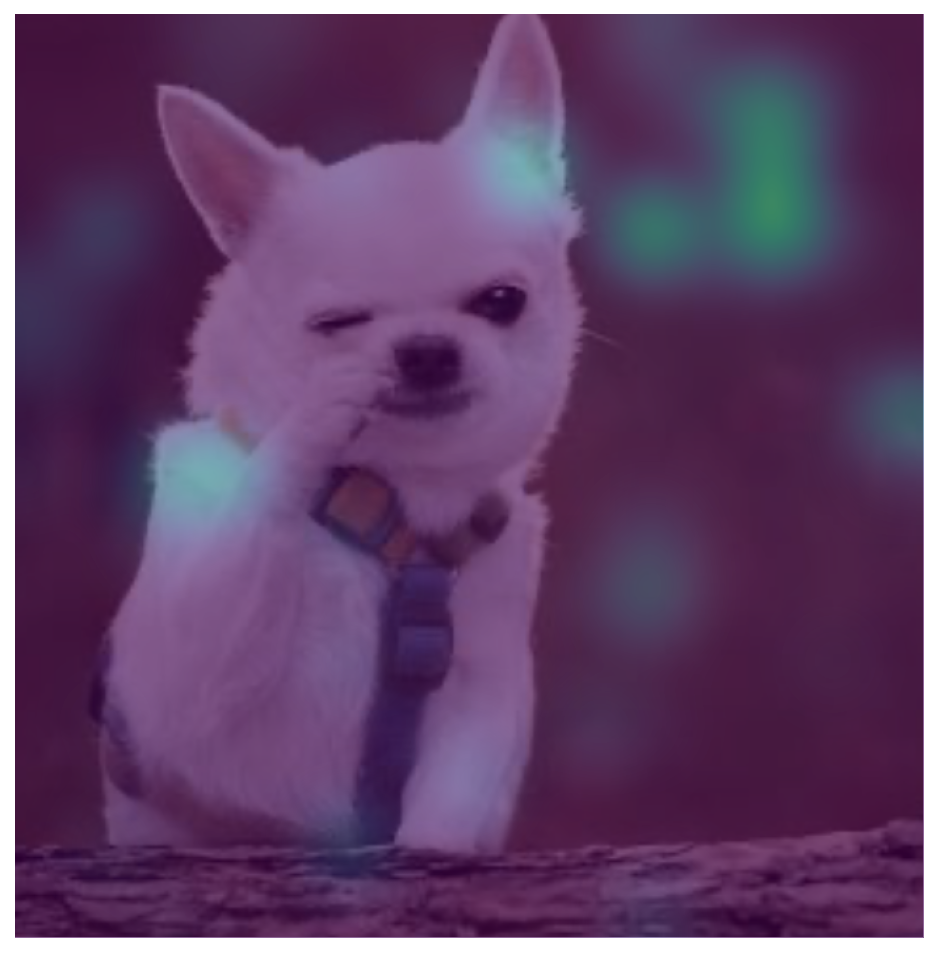}}\hfill
    \subfigure[Step 5]{\includegraphics[width=0.115\textwidth]{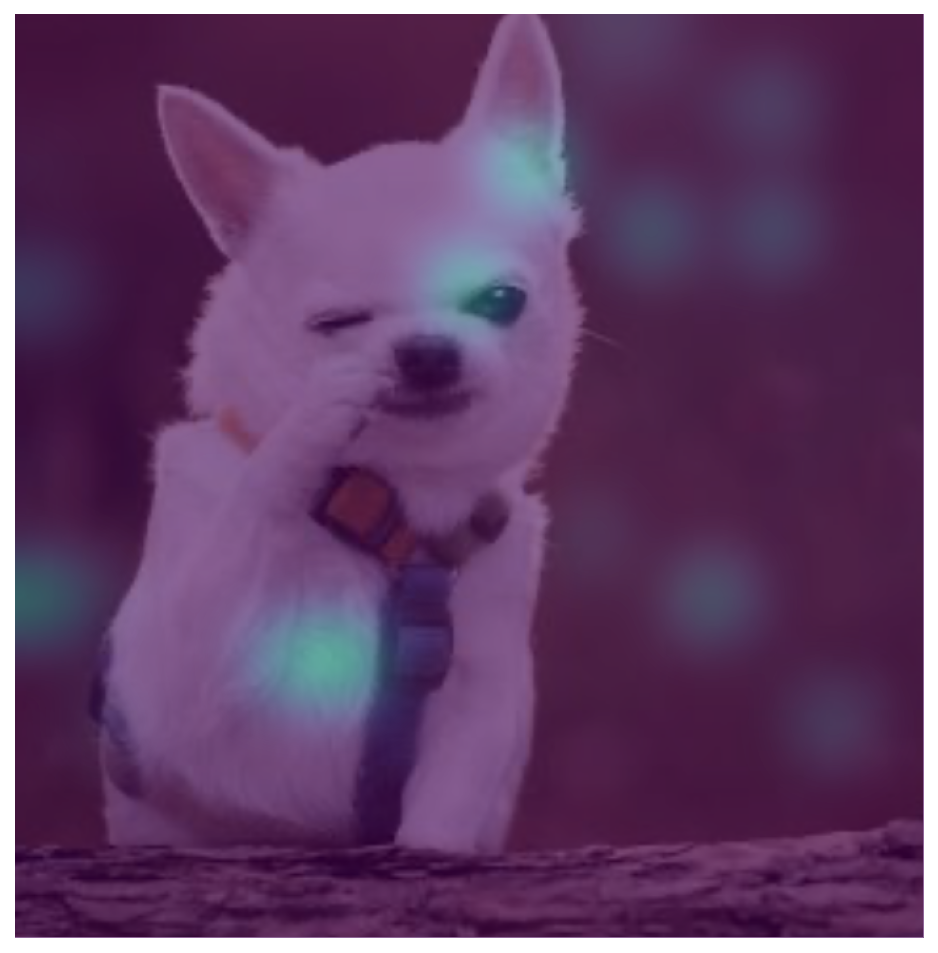}}\hfill
    \subfigure[Step 6]{\includegraphics[width=0.115\textwidth]{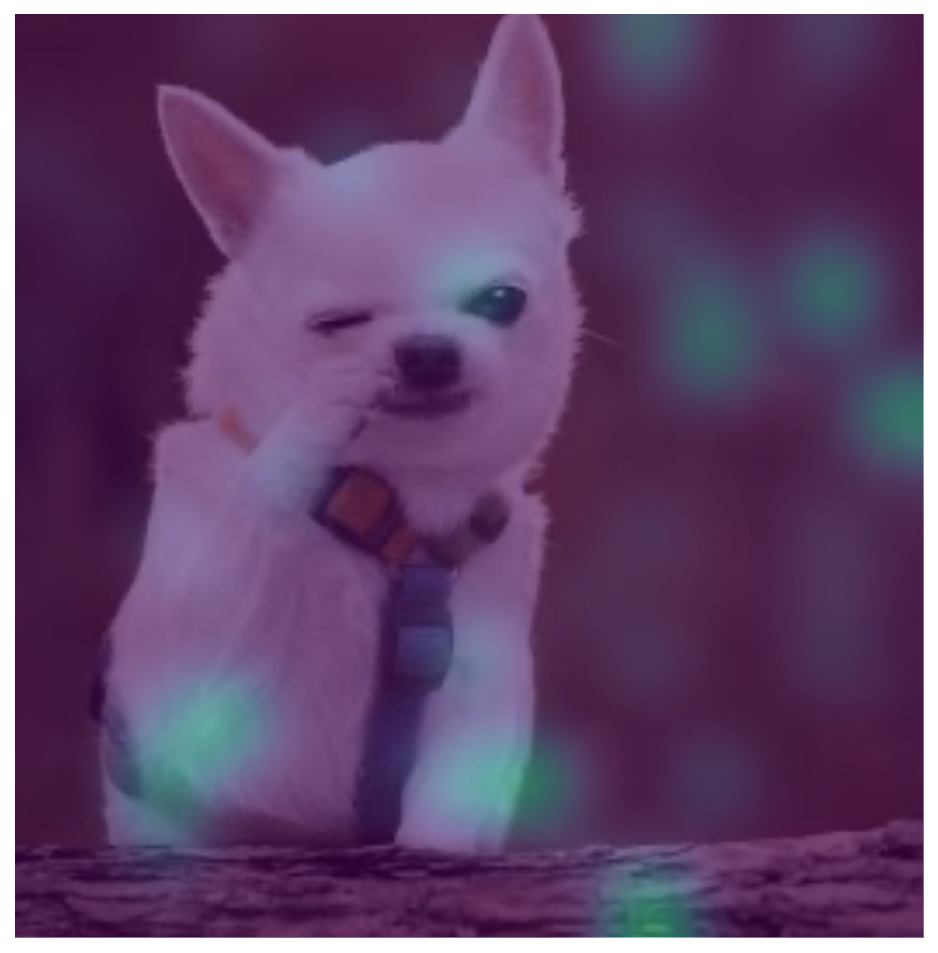}}\hfill
    \subfigure[Step 7]{\includegraphics[width=0.115\textwidth]{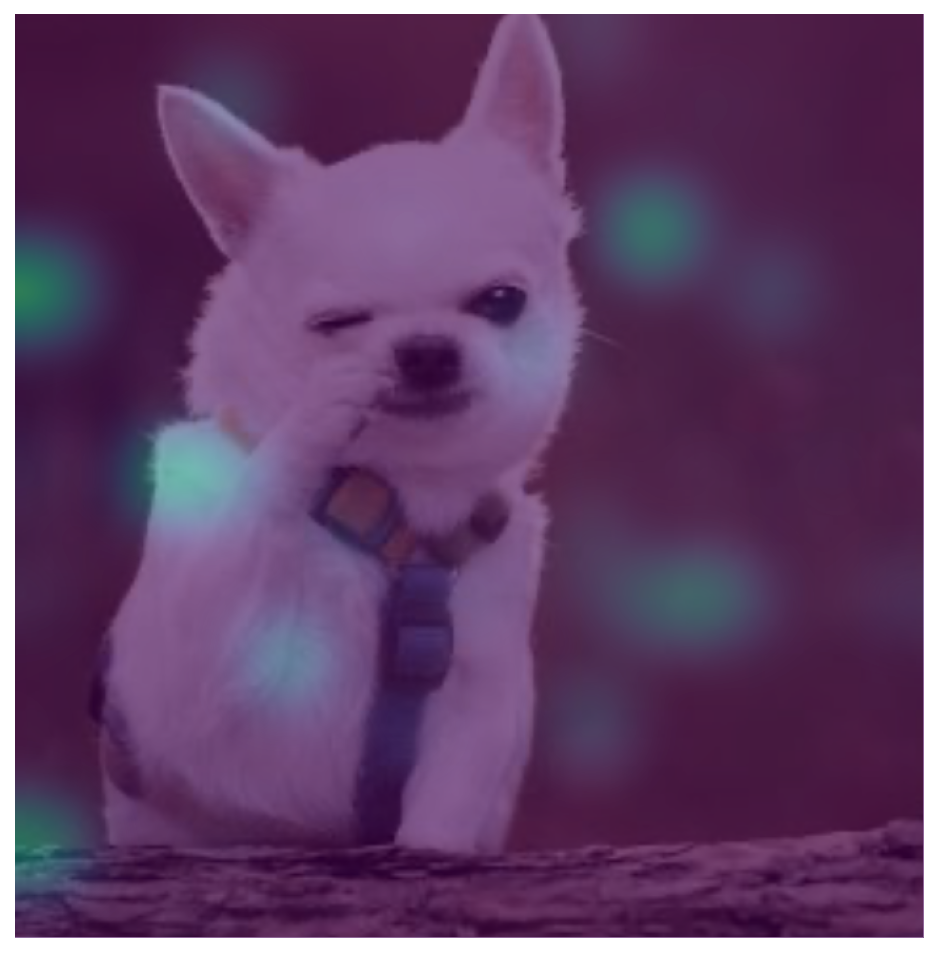}}\hfill
    \subfigure[Step 8]{\includegraphics[width=0.115\textwidth]{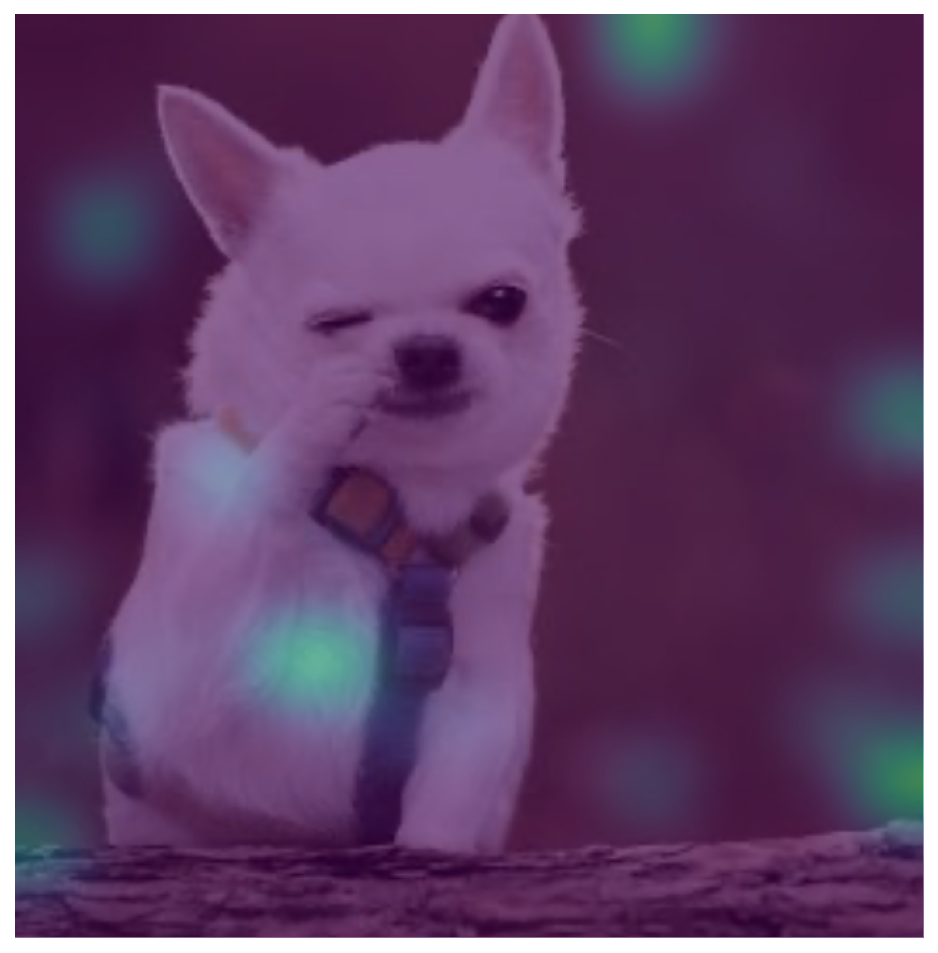}}
    \\[-1ex]
    \subfigure[Step 9]{\includegraphics[width=0.115\textwidth]{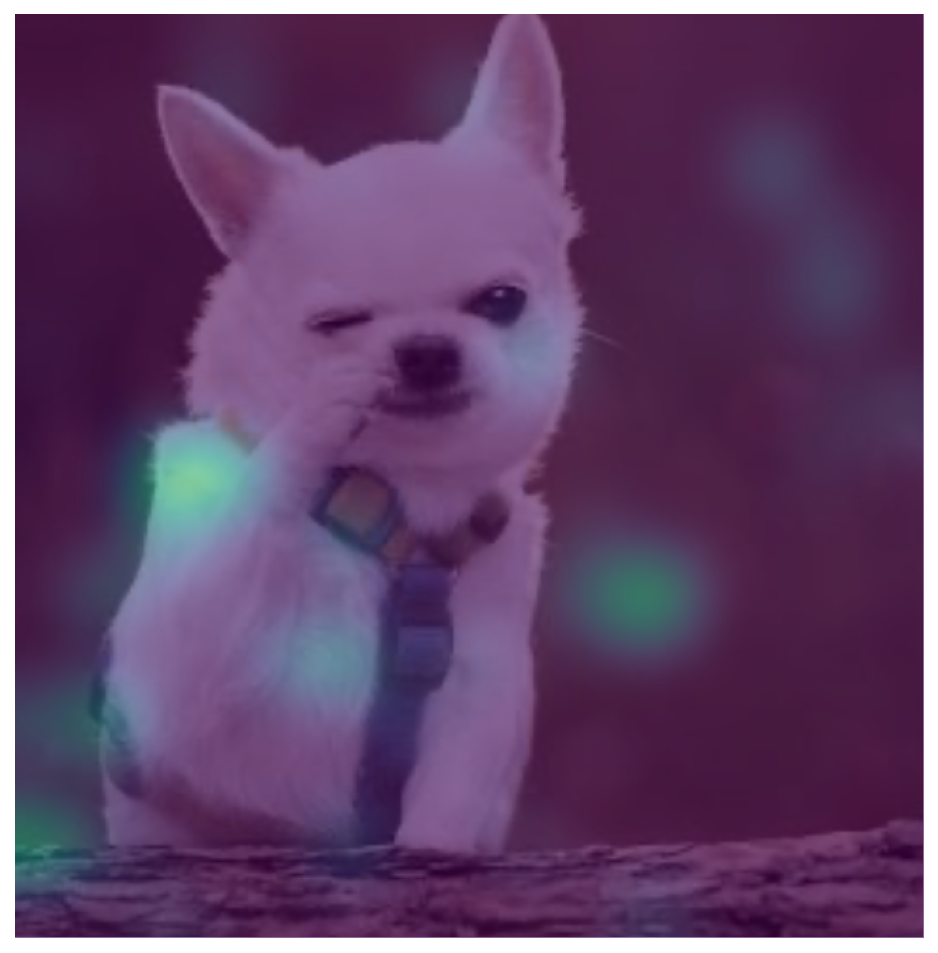}}\hfill
    \subfigure[Step 10]{\includegraphics[width=0.115\textwidth]{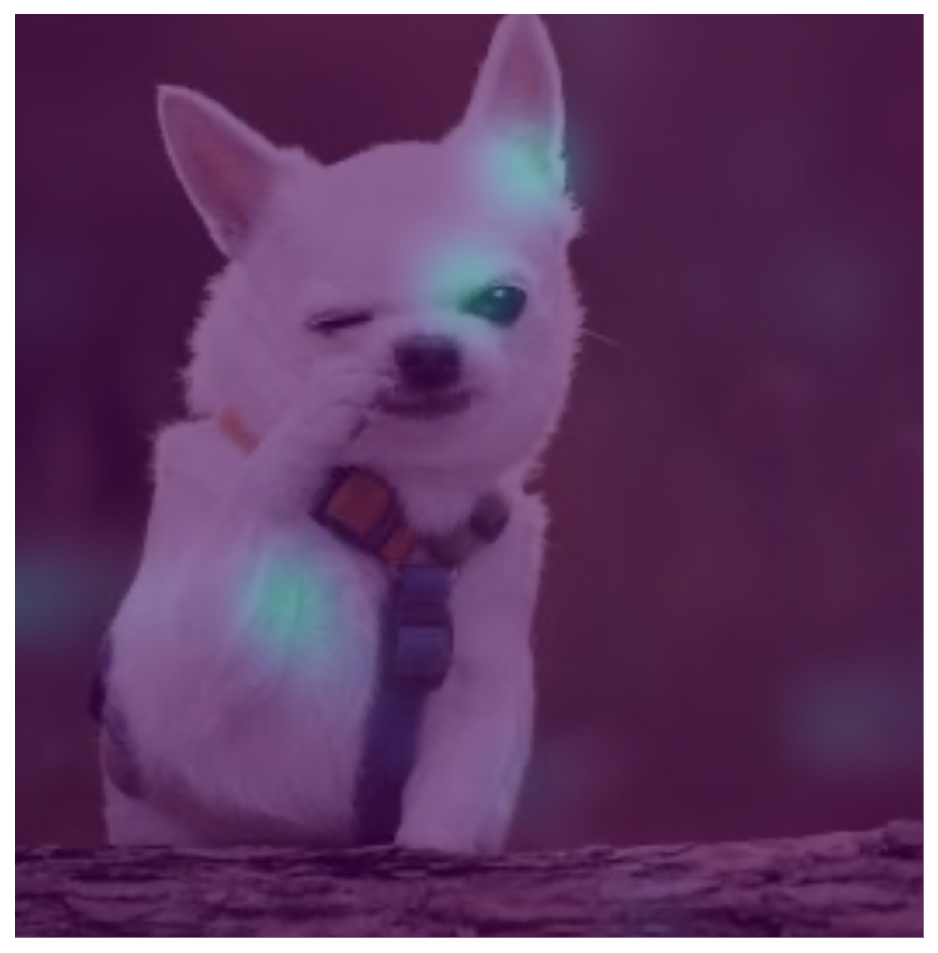}}\hfill
    \subfigure[Step 11]{\includegraphics[width=0.115\textwidth]{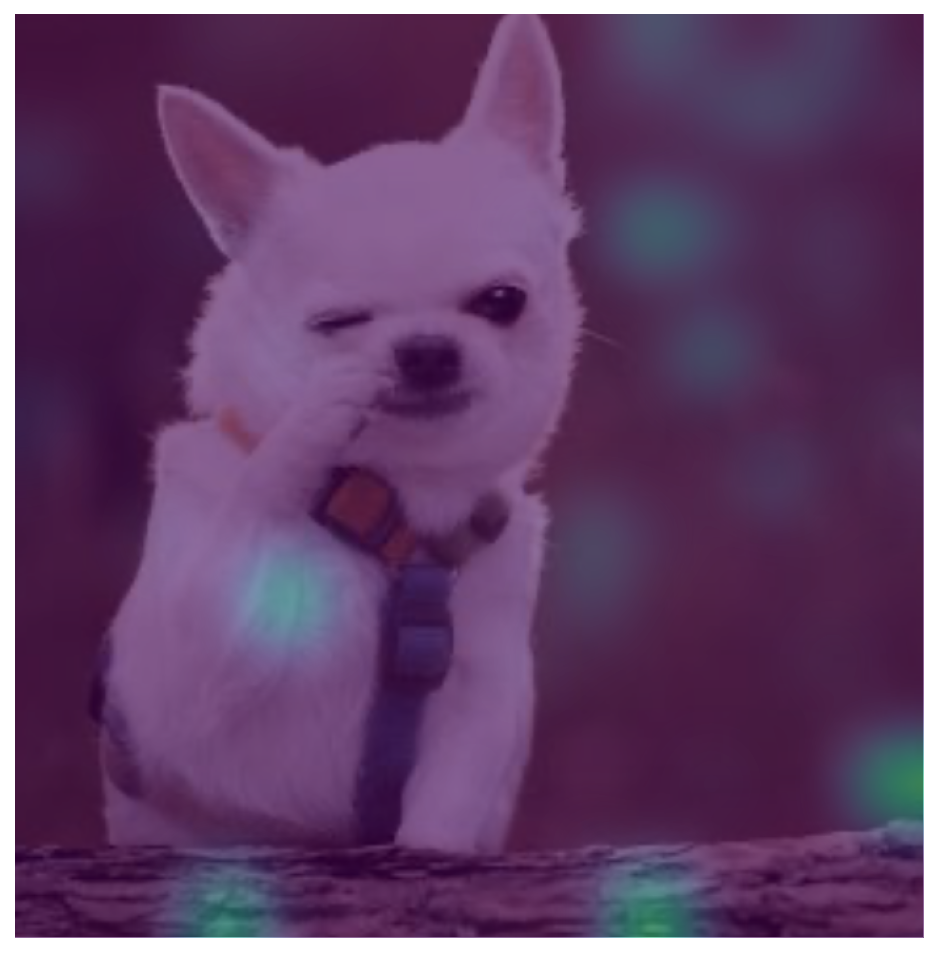}}\hfill
    \subfigure[Step 12]{\includegraphics[width=0.115\textwidth]{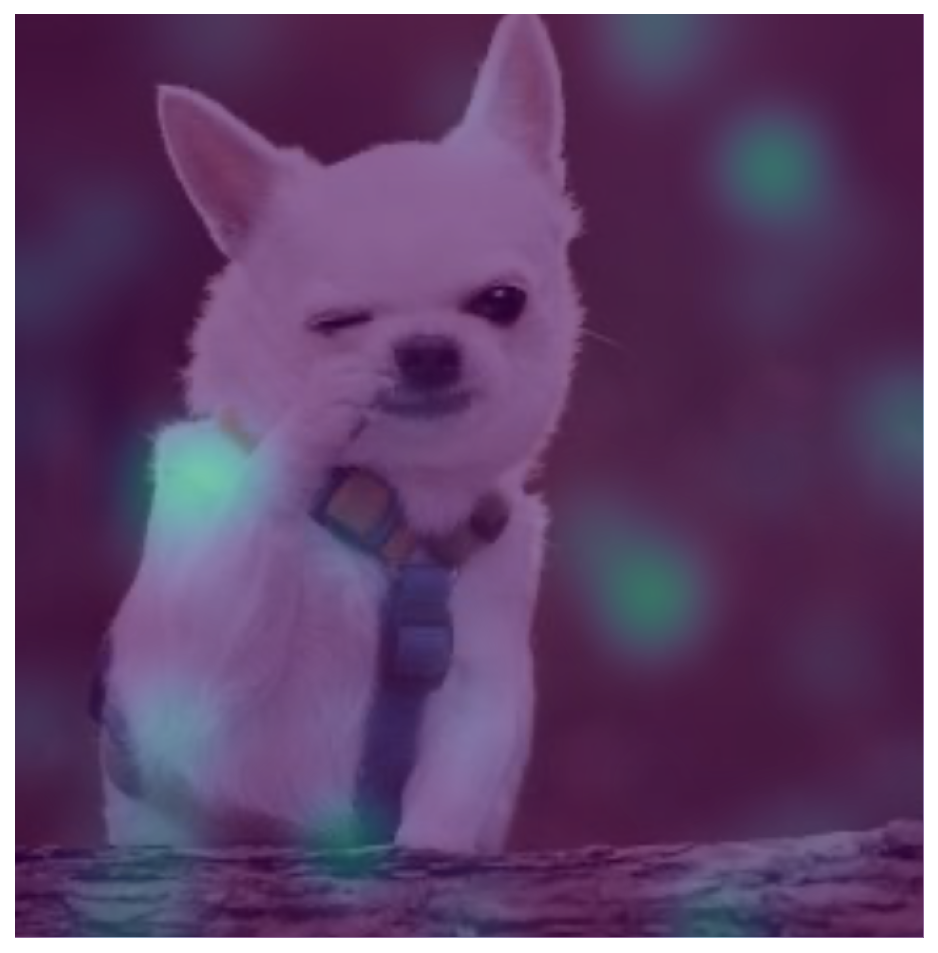}}\hfill
    \subfigure[Step 13]{\includegraphics[width=0.115\textwidth]{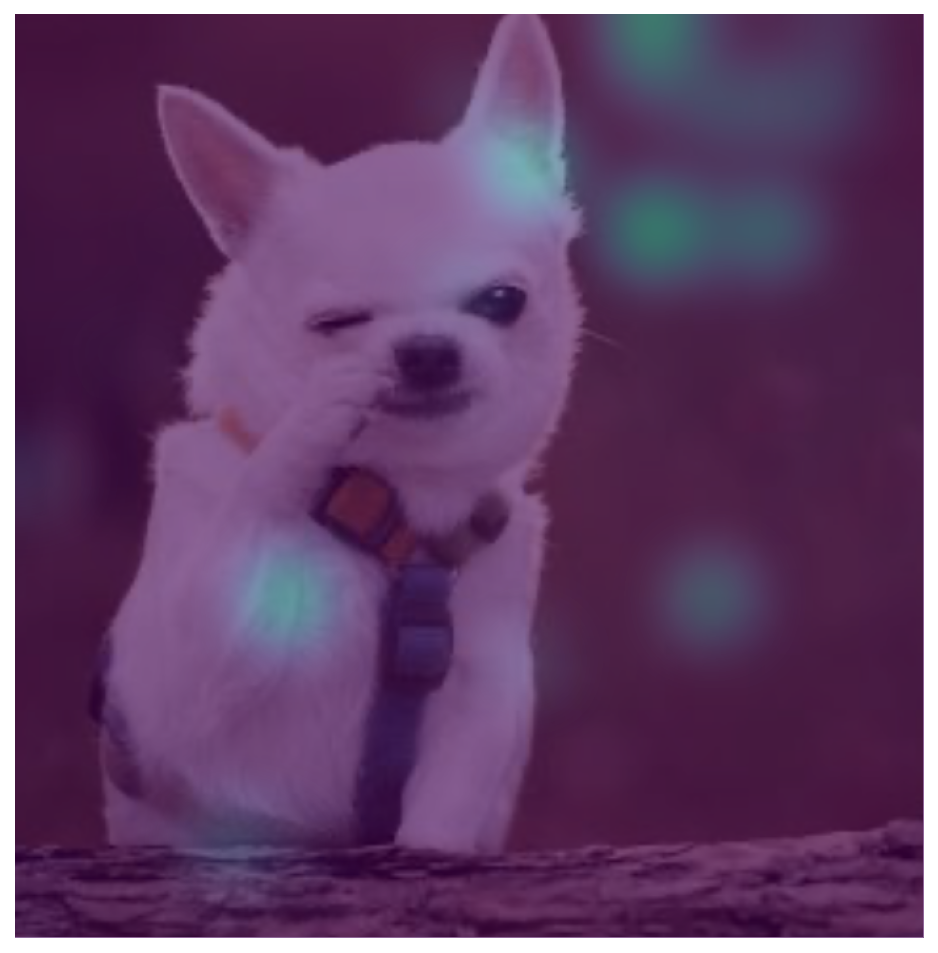}}\hfill
    \subfigure[Step 14]{\includegraphics[width=0.115\textwidth]{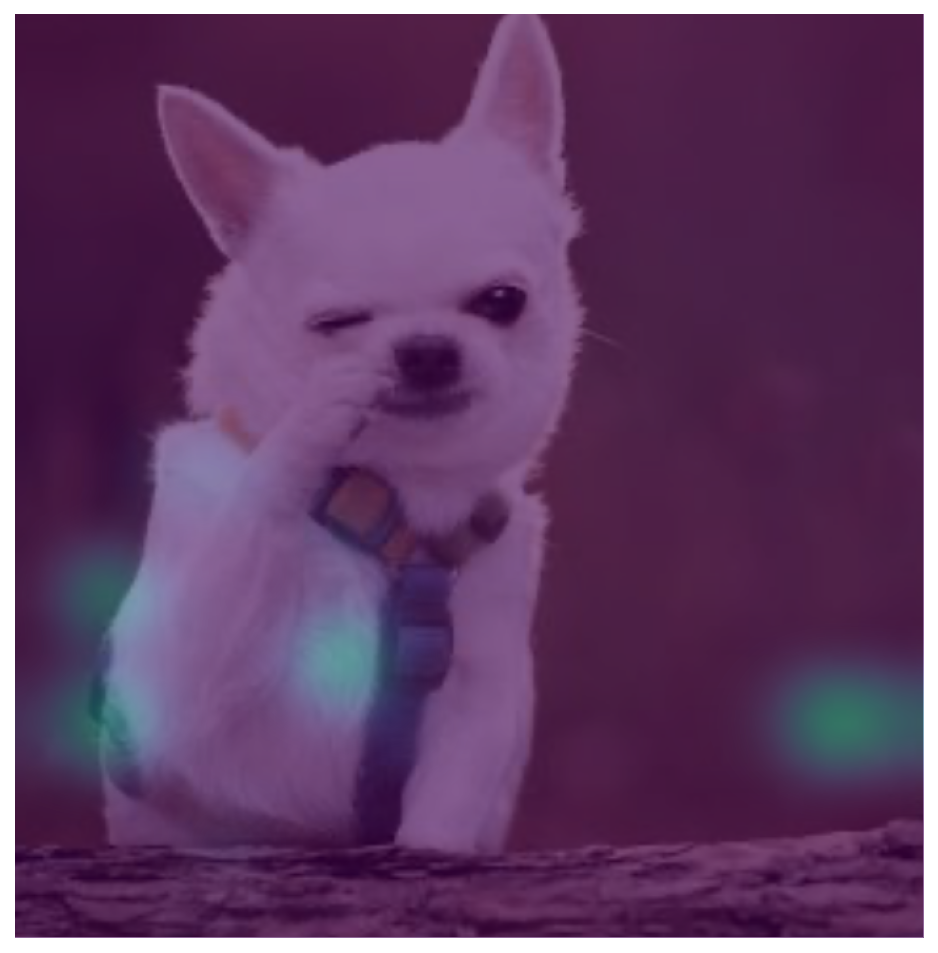}}\hfill
    \subfigure[Step 15]{\includegraphics[width=0.115\textwidth]{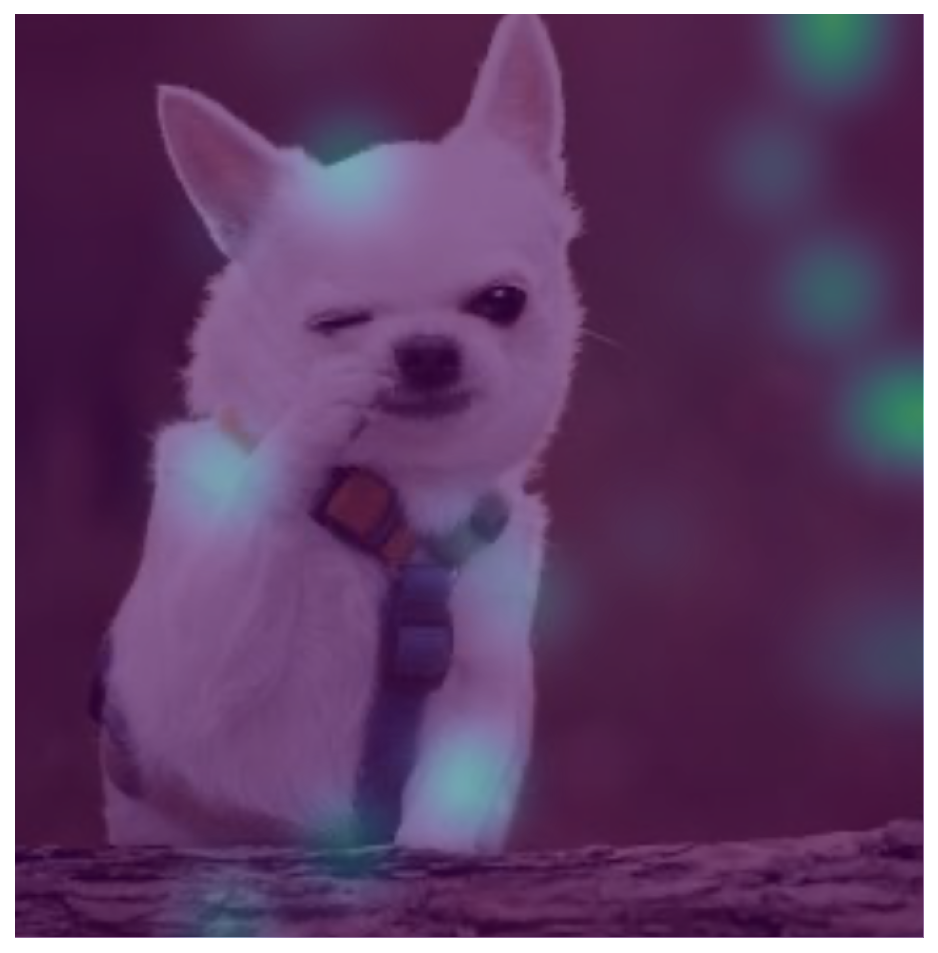}}\hfill
    \subfigure[Step 16]{\includegraphics[width=0.115\textwidth]{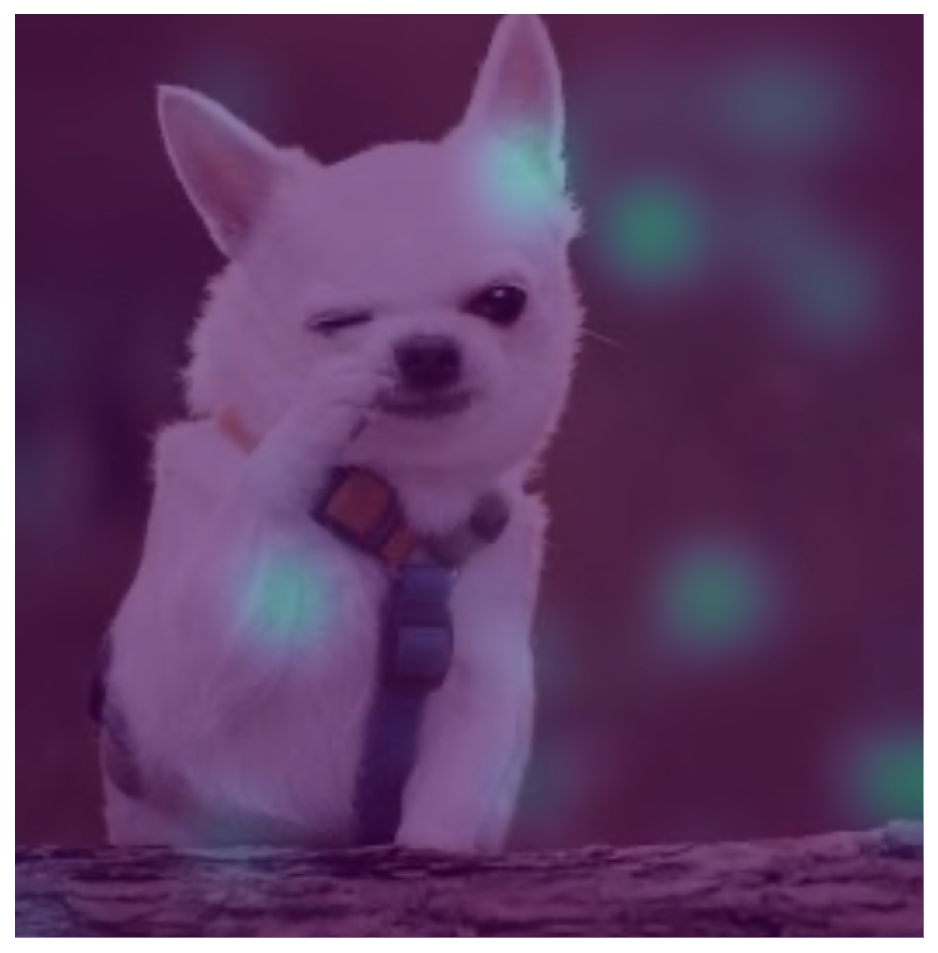}}
    \caption{Visualization of the most important visual token group at each denoising step (Part 1). The highlighted region in each subfigure corresponds to the visual token with the maximum decider-guided importance score.}
    \label{fig:importance_evolution_1}
\end{figure*}

\begin{figure*}[h!]
    \centering
    \ContinuedFloat
    \subfigure[Step 17]{\includegraphics[width=0.115\textwidth]{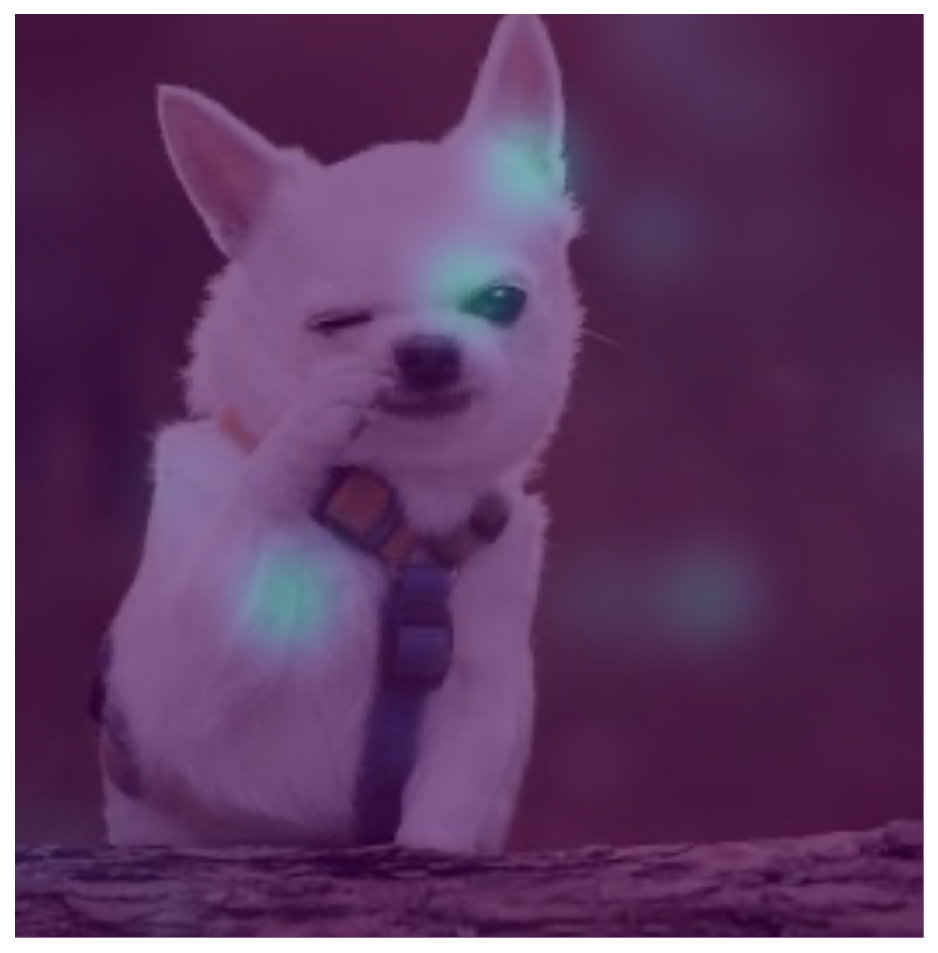}}\hfill
    \subfigure[Step 18]{\includegraphics[width=0.115\textwidth]{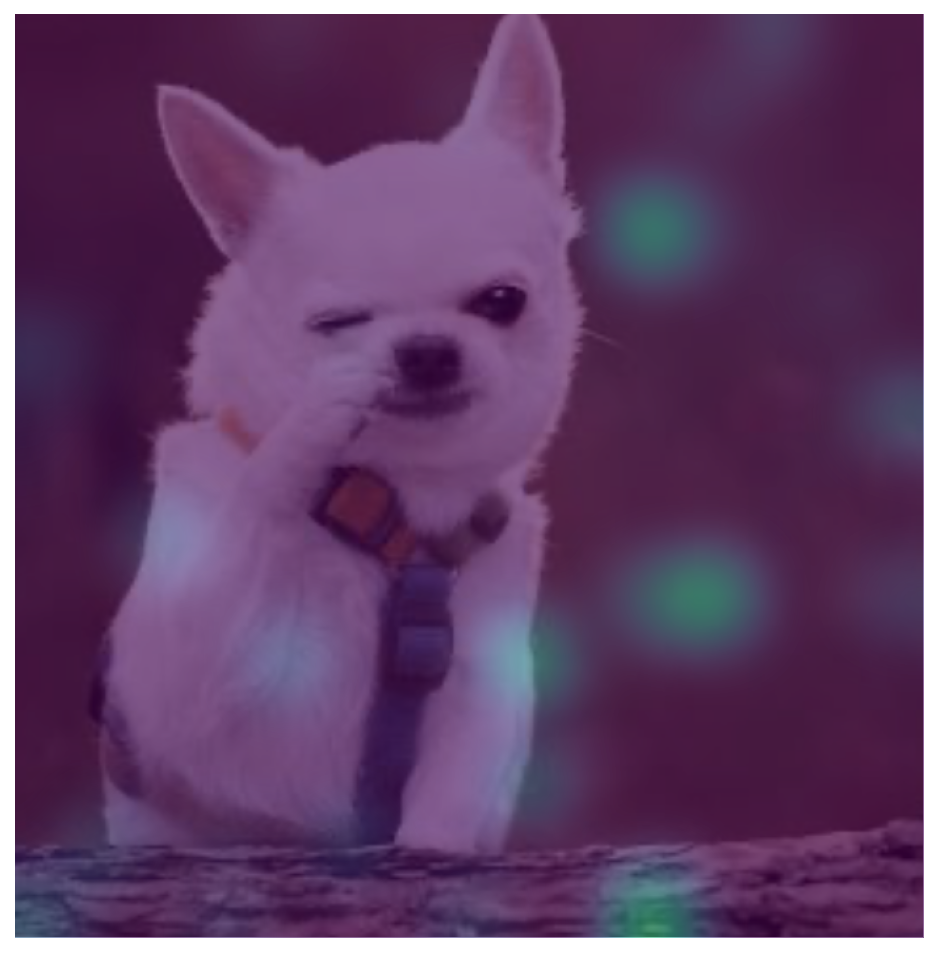}}\hfill
    \subfigure[Step 19]{\includegraphics[width=0.115\textwidth]{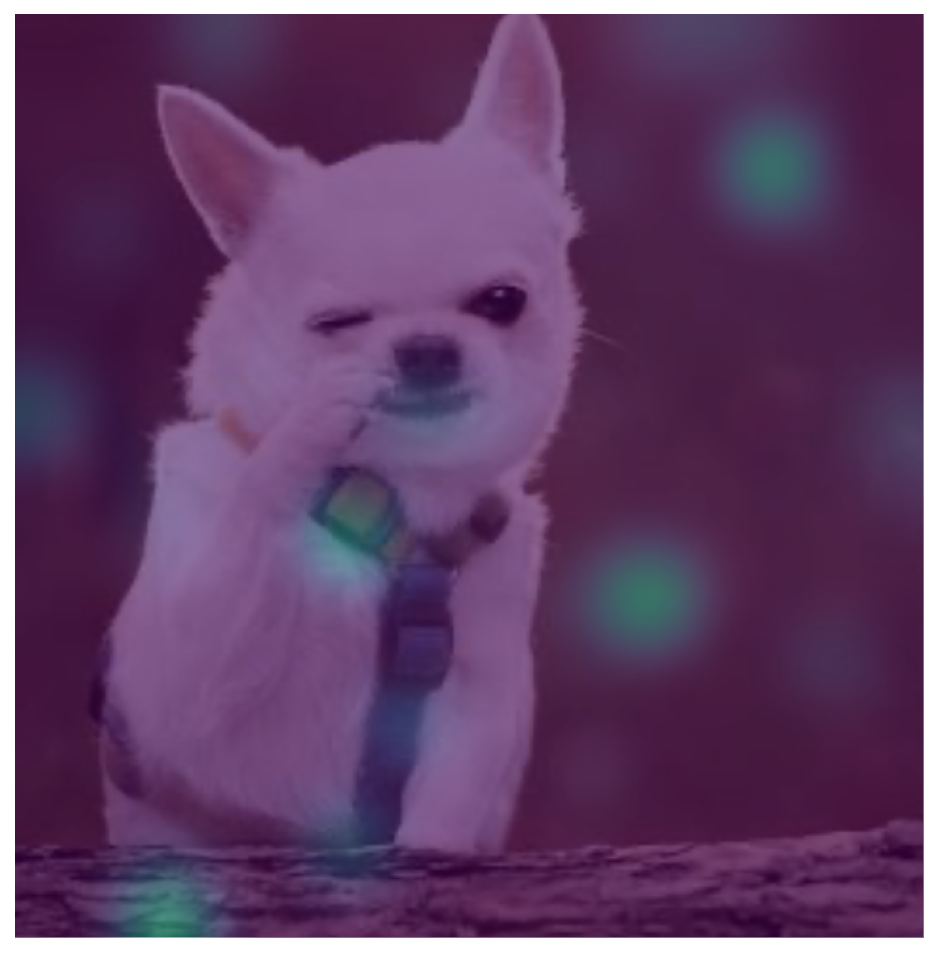}}\hfill
    \subfigure[Step 20]{\includegraphics[width=0.115\textwidth]{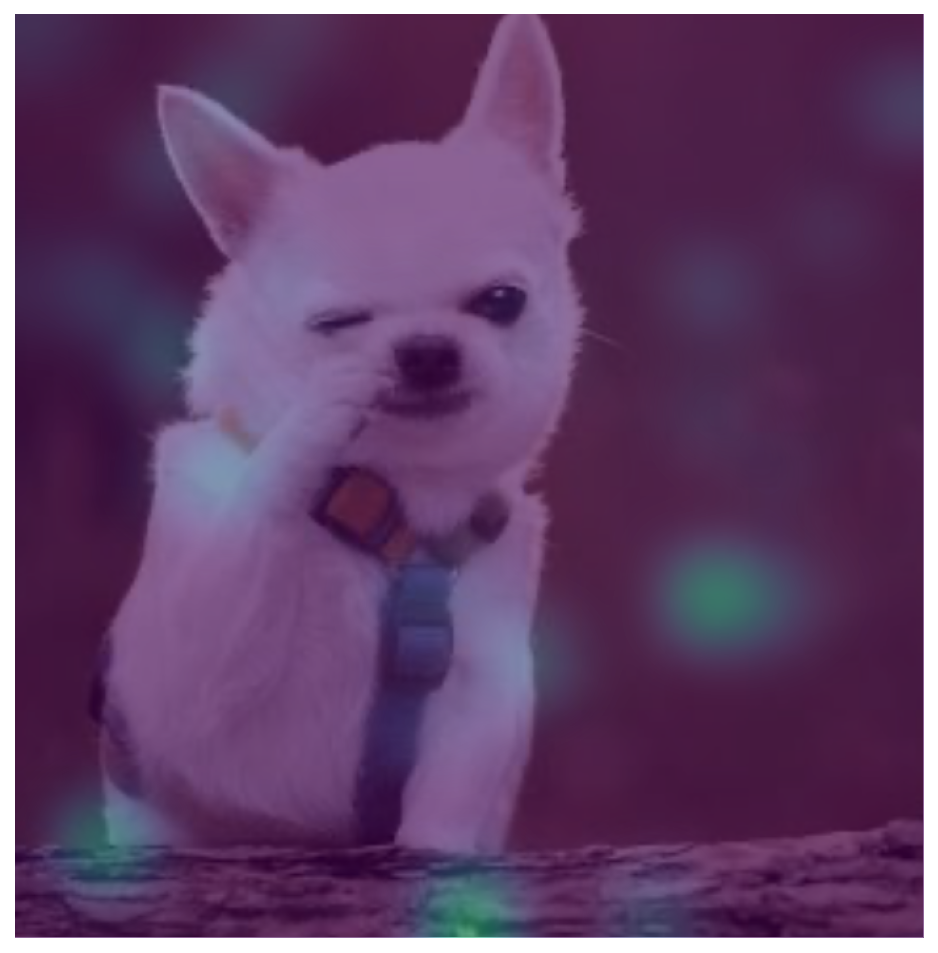}}\hfill
    \subfigure[Step 21]{\includegraphics[width=0.115\textwidth]{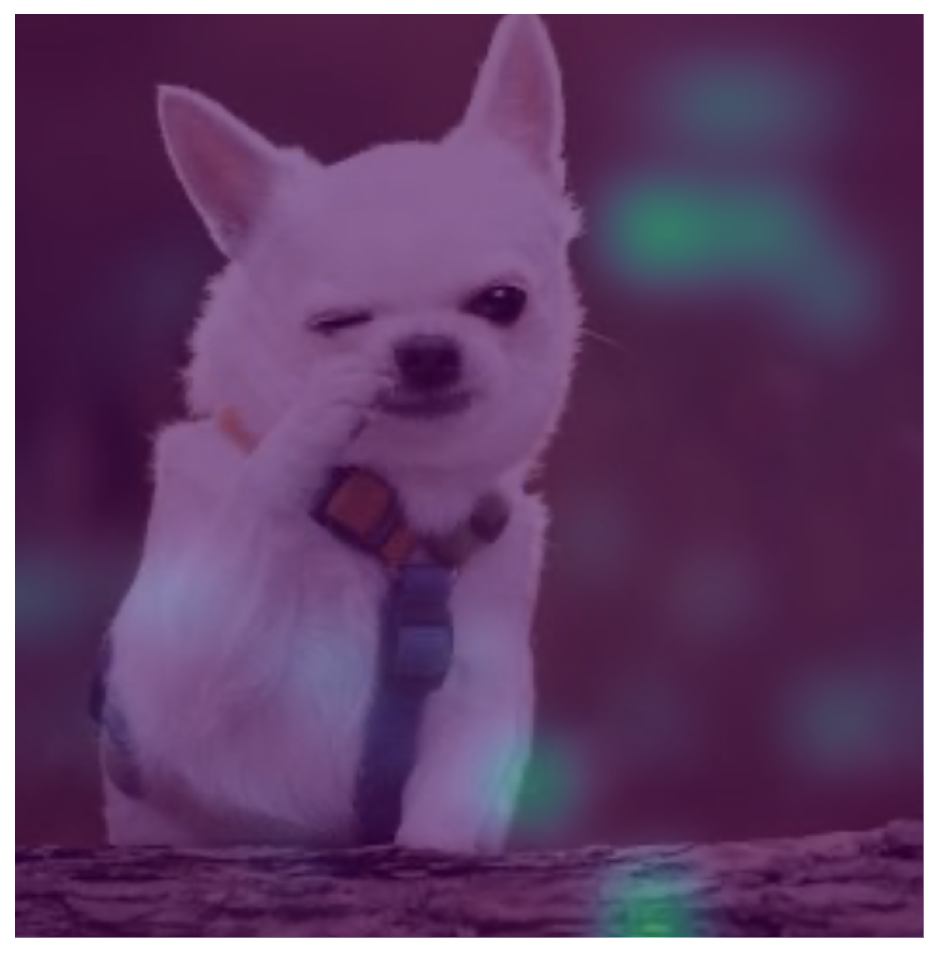}}\hfill
    \subfigure[Step 22]{\includegraphics[width=0.115\textwidth]{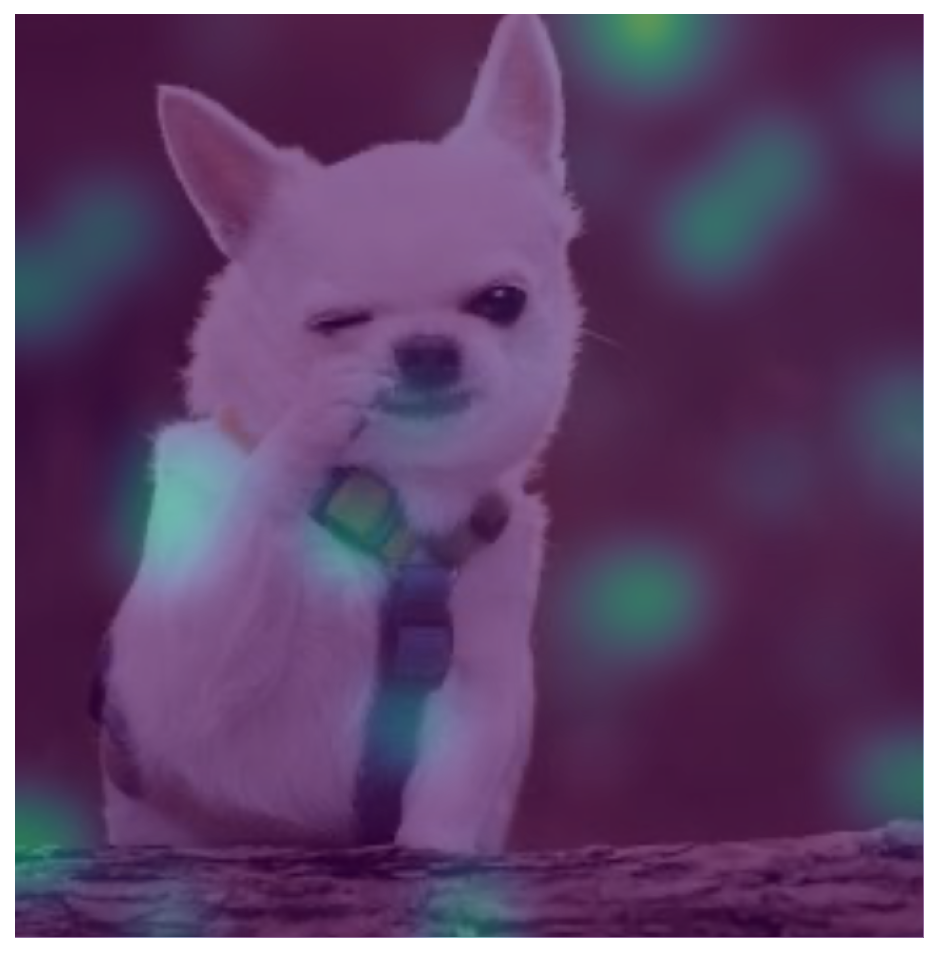}}\hfill
    \subfigure[Step 23]{\includegraphics[width=0.115\textwidth]{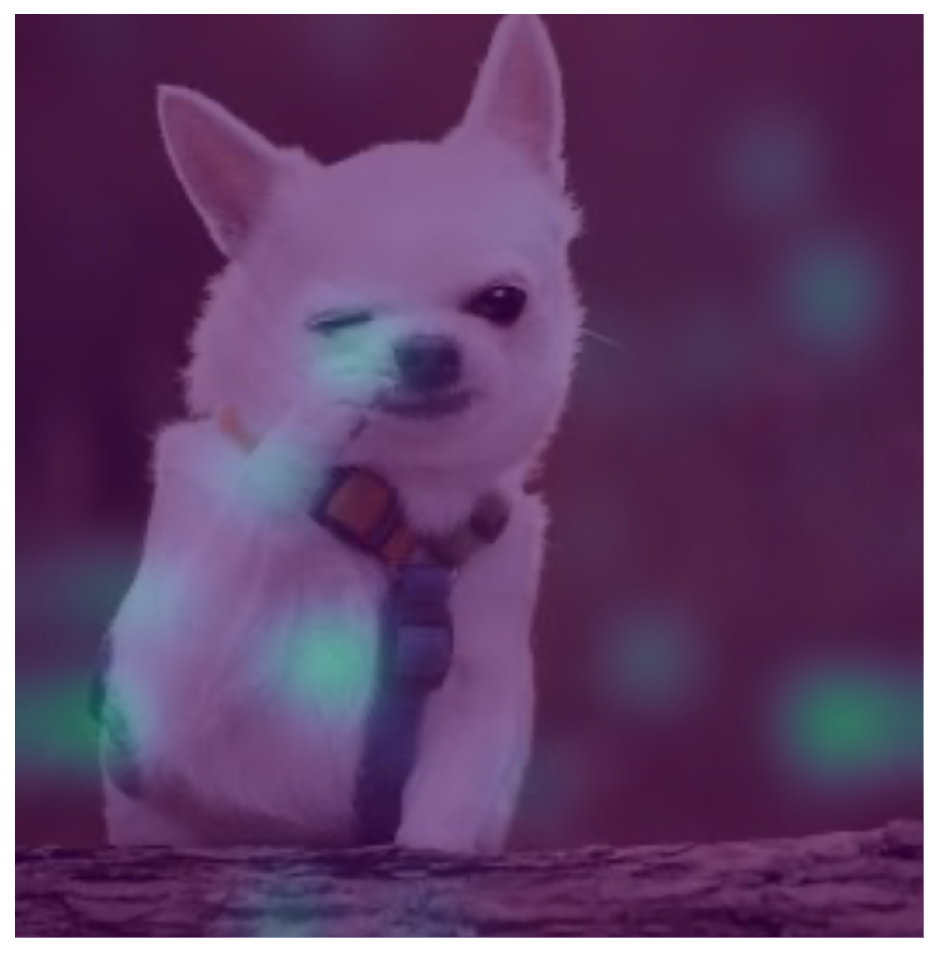}}\hfill
    \subfigure[Step 24]{\includegraphics[width=0.115\textwidth]{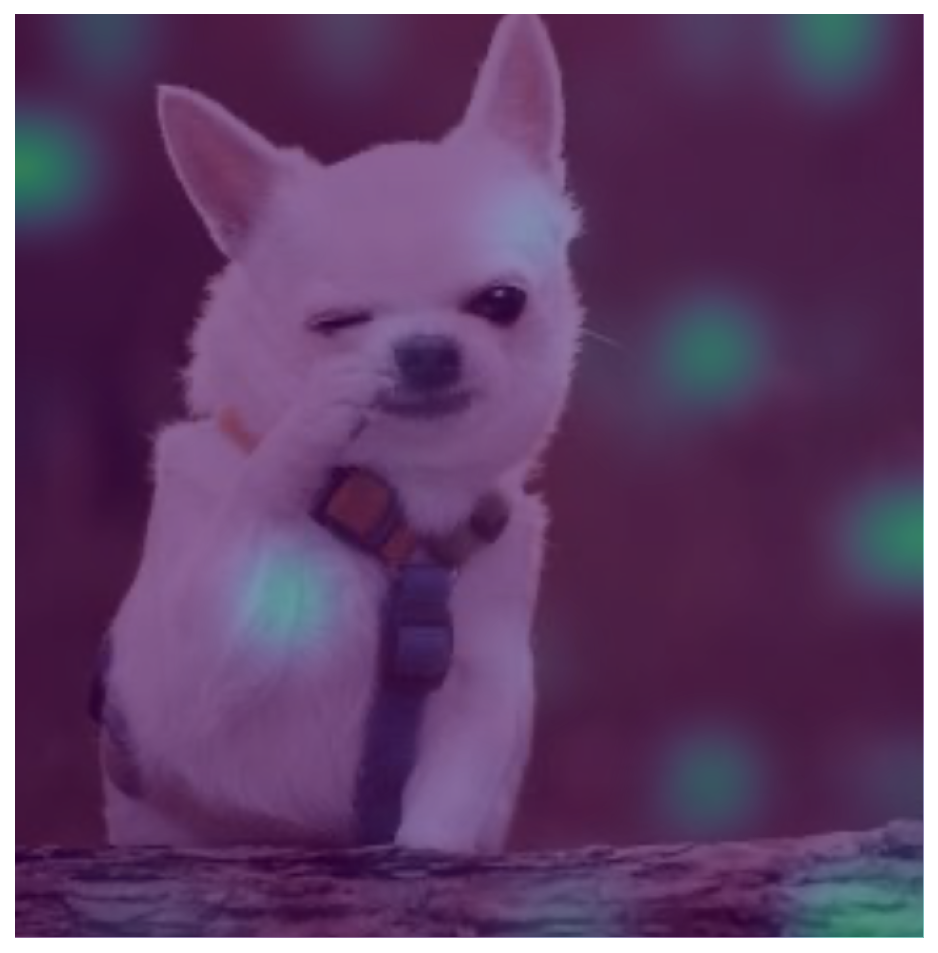}}
    \\[-1ex]
    \subfigure[Step 25]{\includegraphics[width=0.115\textwidth]{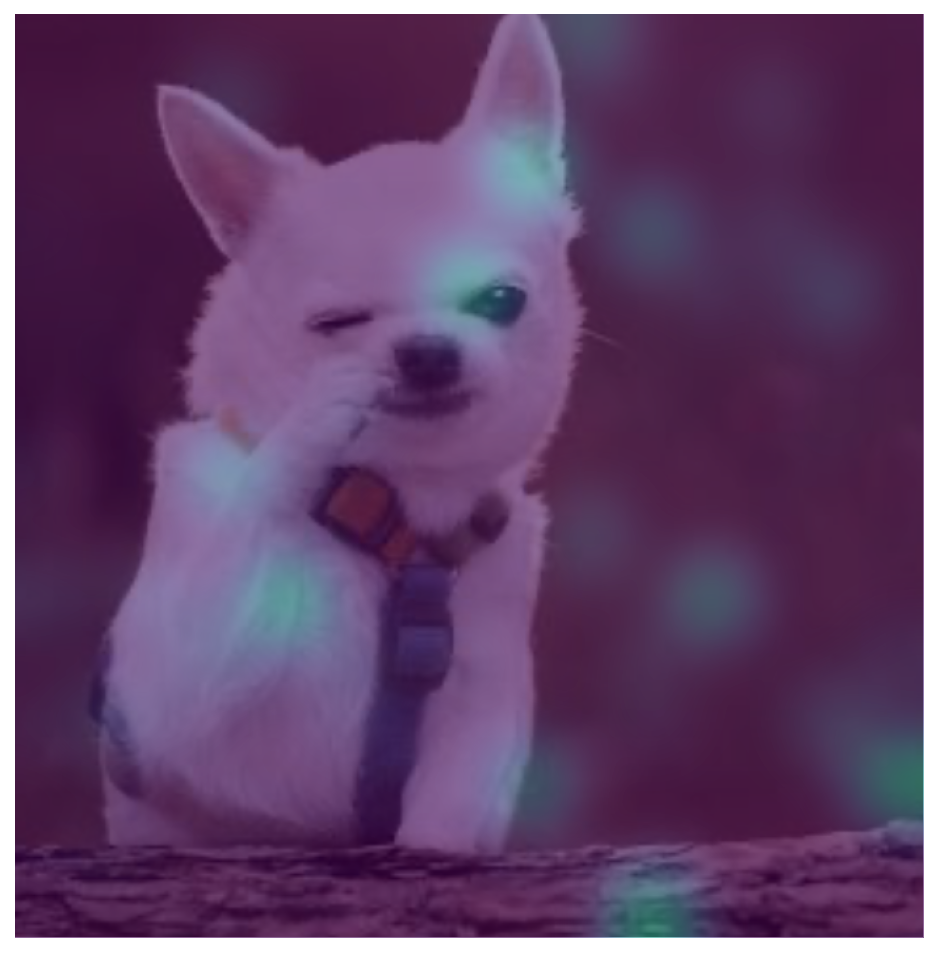}}\hfill
    \subfigure[Step 26]{\includegraphics[width=0.115\textwidth]{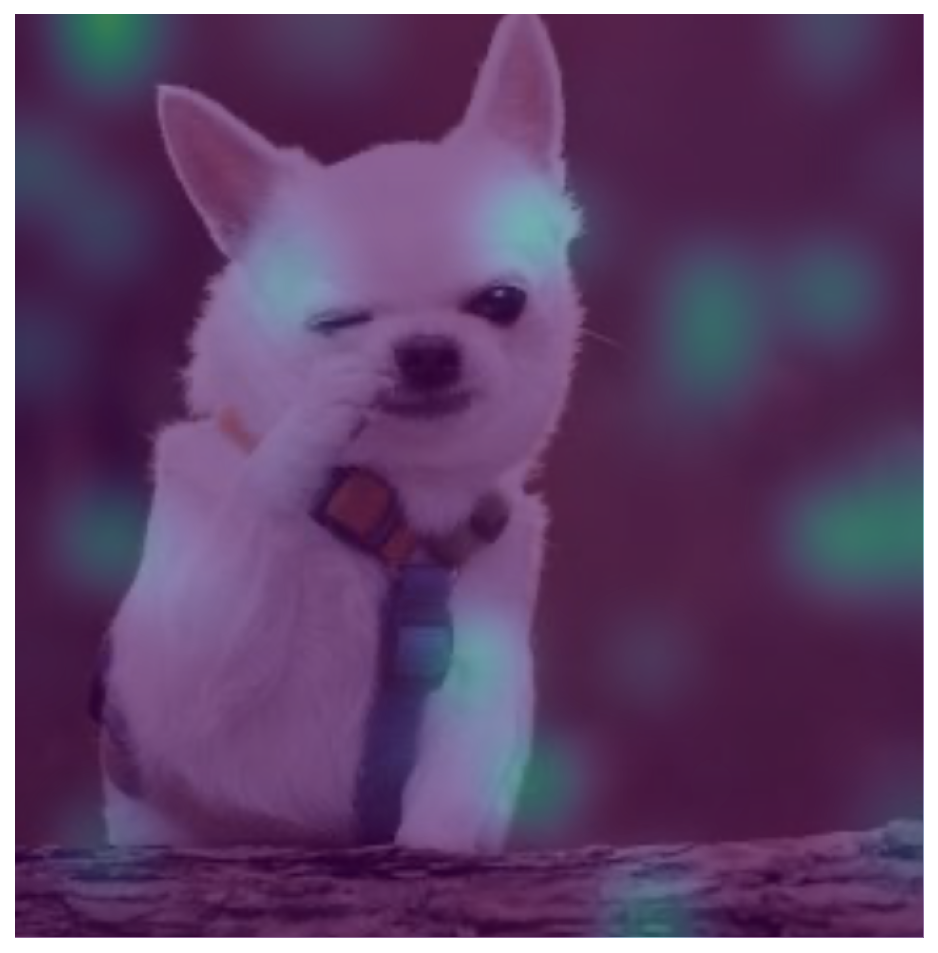}}\hfill
    \subfigure[Step 27]{\includegraphics[width=0.115\textwidth]{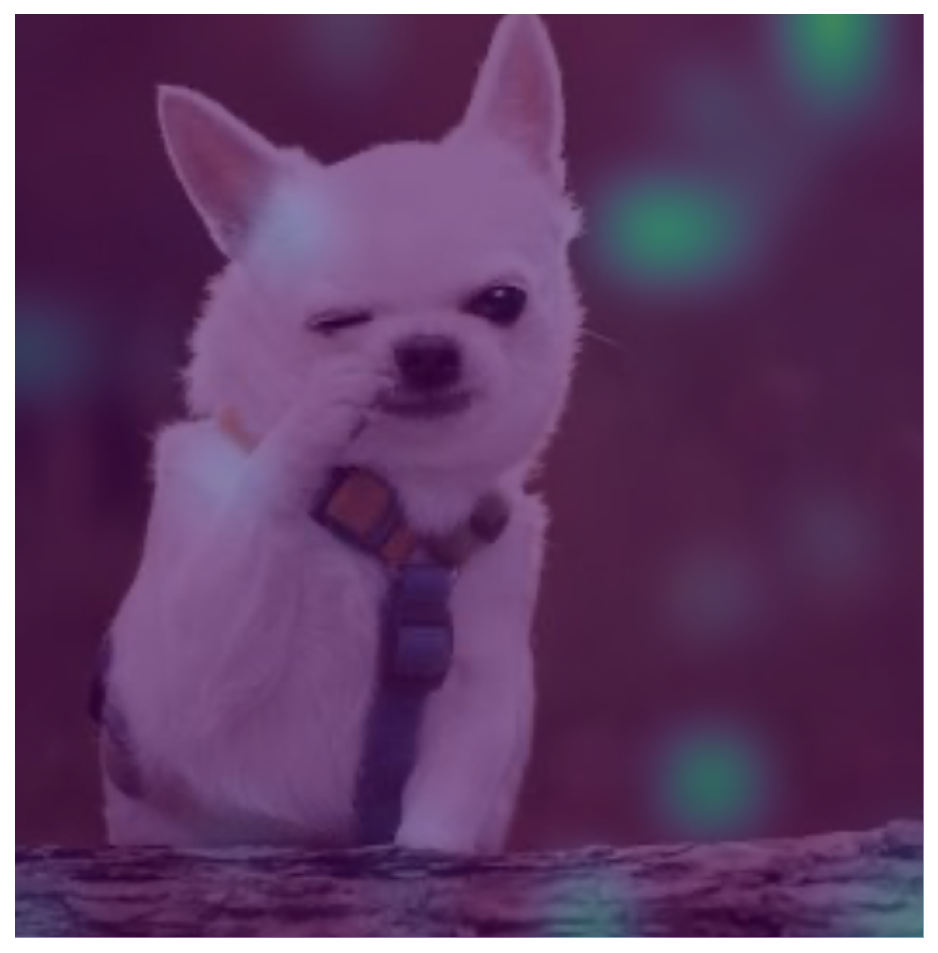}}\hfill
    \subfigure[Step 28]{\includegraphics[width=0.115\textwidth]{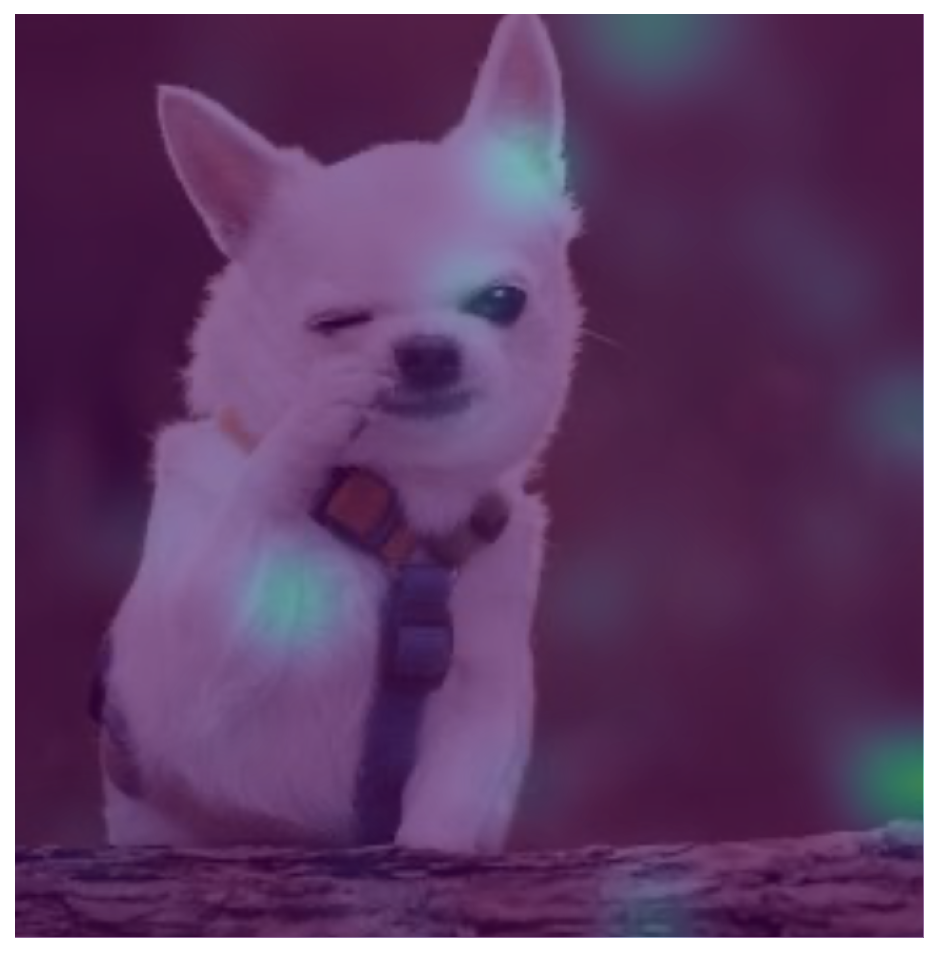}}\hfill
    \subfigure[Step 29]{\includegraphics[width=0.115\textwidth]{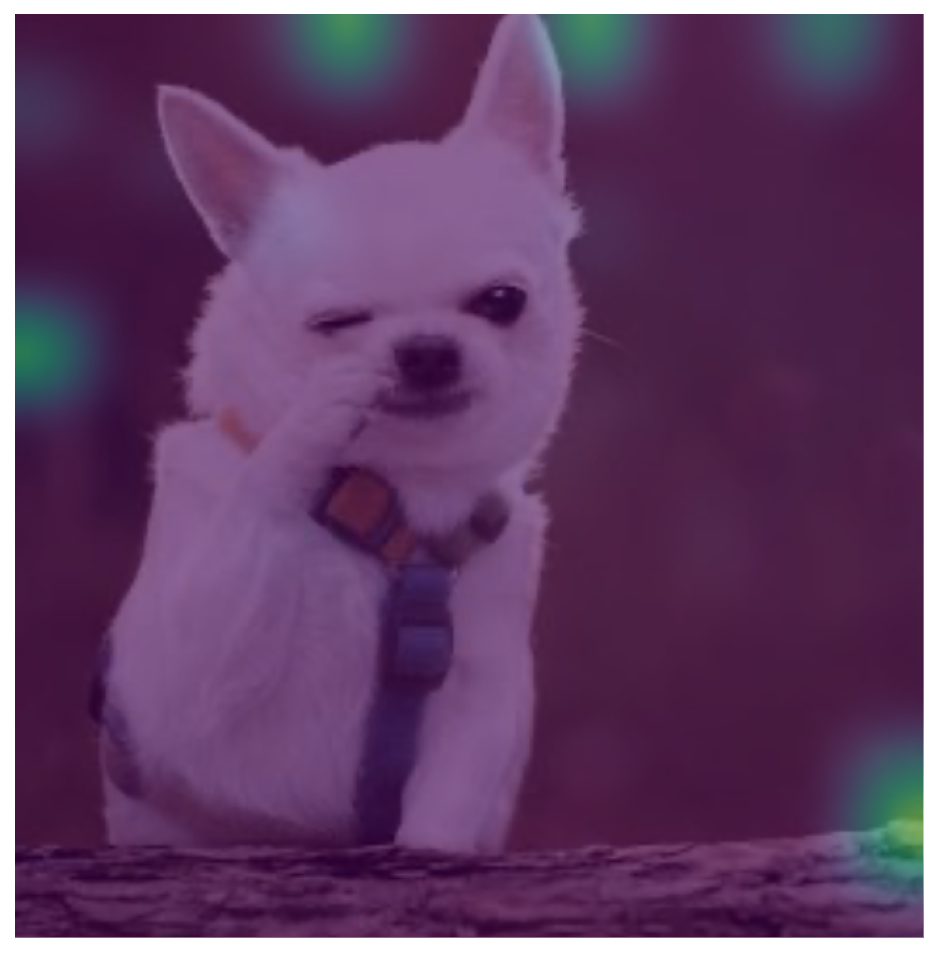}}\hfill
    \subfigure[Step 30]{\includegraphics[width=0.115\textwidth]{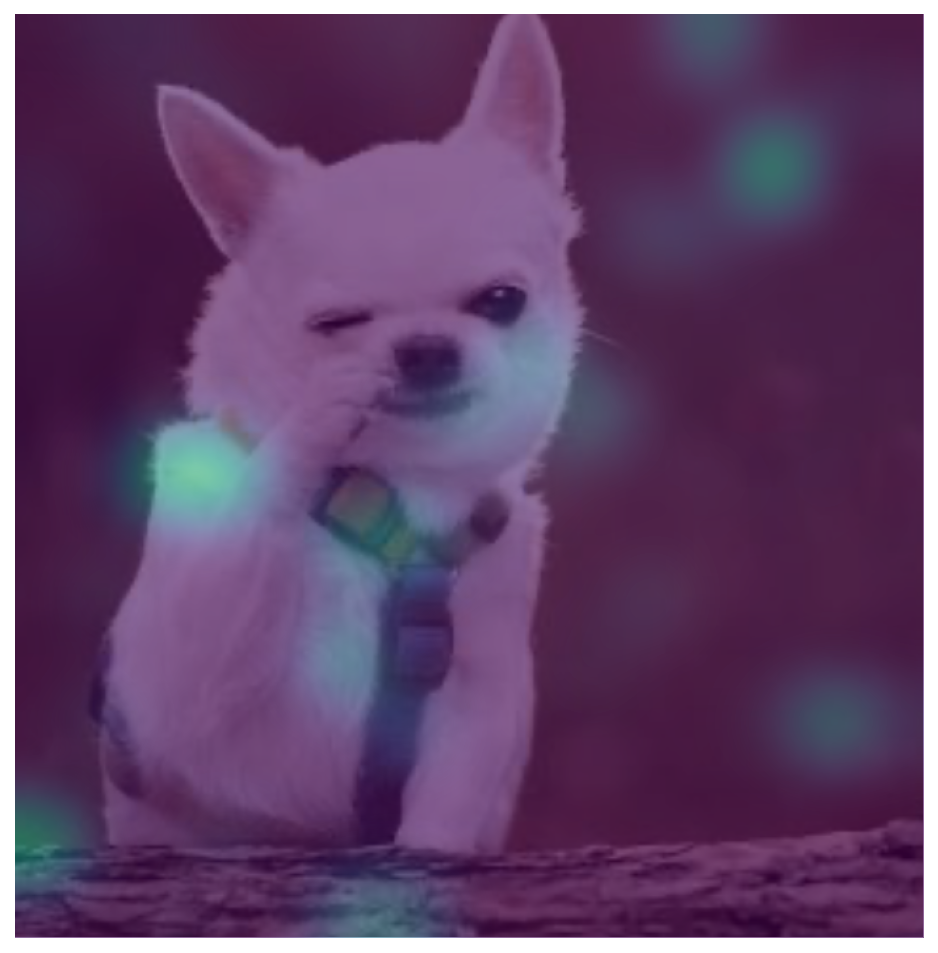}}\hfill
    \subfigure[Step 31]{\includegraphics[width=0.115\textwidth]{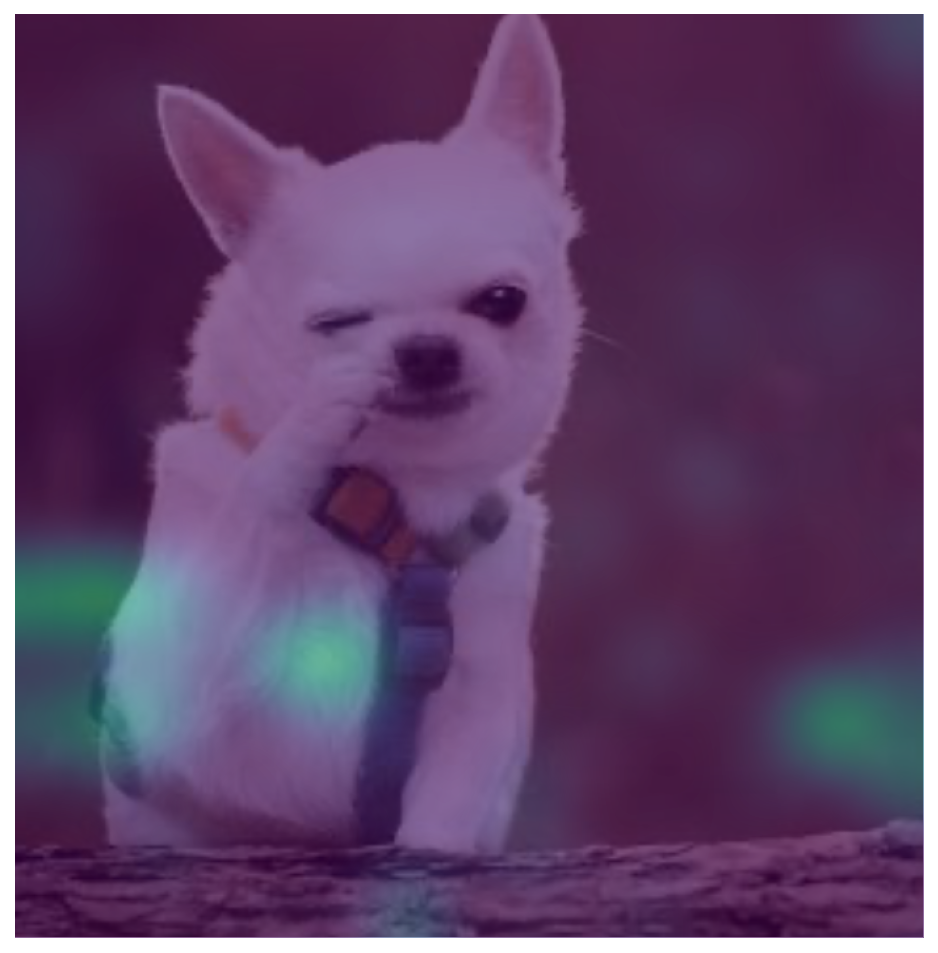}}\hfill
    \subfigure[Step 32]{\includegraphics[width=0.115\textwidth]{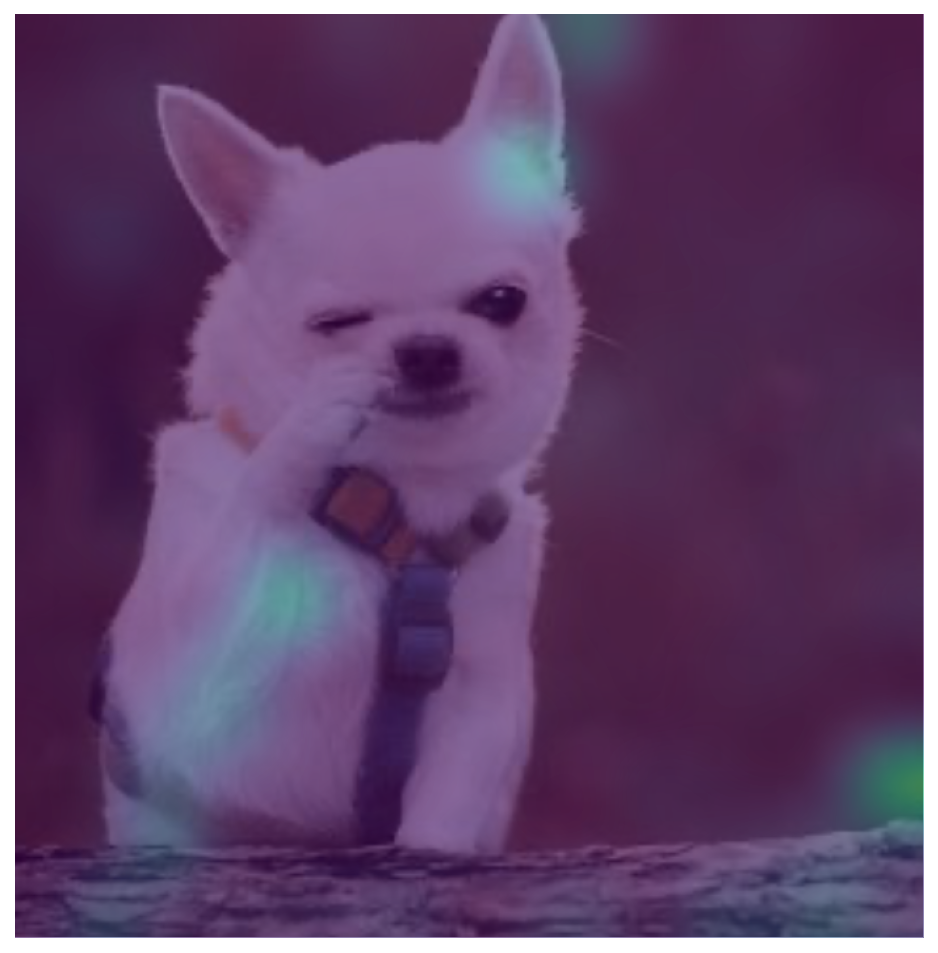}}
    \caption{Visualization of the most important visual token group at each denoising step (Part 2). The shifting focus demonstrates the dynamic nature of visual salience during the iterative generation process.}
    \label{fig:importance_evolution}
\end{figure*}

The entire dual-pass scheme introduces a negligible overhead. Since the merge occurs after the self-attention step of layer $l^{\ast}$, all subsequent layers operate on a physically shorter sequence. The total added cost is limited to one extra FlashAttention call and a linear-time merge operation, which is insignificant relative to the overall model FLOPs.

\section{Generalization Analysis}

\subsection{Visual Encoder Generalization}

Our D\textsuperscript{3}ToM method is agnostic to the choice of the visual encoder. This independence stems from its core design: the token merging module operates exclusively on the sequence of visual embeddings after they have been projected into the shared language embedding space. Consequently, no gradients or control signals are propagated back to the vision tower. This ensures that architectural differences in the encoder, such as patching strategies, positional encodings, or pre-training objectives (e.g., CLIP versus SigLIP), do not affect the merge logic.

This design allows various encoders to be used interchangeably. For example, the models evaluated in our work, LLaDA-V and LaViDa, use a SigLIP-400M encoder. However, other common encoders, such as the CLIP-ViT series frequently adopted by LLaVA-style models, are also directly compatible. The practical benefit is that no hyperparameter re-tuning or architectural search for our method is necessary when swapping visual front-ends; only the absolute number of visual tokens changes, not the downstream interface. Our empirical results confirm that after normalizing for token counts, the merge schedules remain consistent across different encoders. This suggests that future work could extend our method to other architectures, such as large-stride ConvNext or Swin models, by simply adjusting the projector's dimensions while keeping the core merge hyperparameters of D\textsuperscript{3}ToM unchanged.

\subsection{Diffusion-based MLLM Generalization}

D\textsuperscript{3}ToM also generalizes seamlessly across different Diffusion-based Multimodal Language Models (MLLMs), provided they are built upon the same language model backbone. We demonstrate this by applying our method to both LaViDa-LLaDA and LLaDA-V. Although these models differ in their visual front-ends and training curricula, they share the same LLaDA-8B diffusion language model. This results in an identical internal attention structure and token embedding space, which is the environment where our module operates.

The architectural consistency between these models is documented in their respective publications. For example, LLaDA-V explicitly "maps visual features into the LLaDA language embedding space", and LaViDa confirms that "we explored LLaDA-8B as our diffusion language model". Because our merge module operates directly within this shared Transformer stack, porting D\textsuperscript{3}ToM from one model to the other requires zero code modifications or changes to its hyperparameters, including the merge layer index $l^\ast$ and the merge schedule $\alpha_t$. As a result, the speed-quality trade-offs established on LaViDa-LLaDA are directly transferable to LLaDA-V. This robustness could be further validated in future work by applying D\textsuperscript{3}ToM with the same hyperparameters to other supported backbones, such as Dream-7B, to confirm its generalization across different model architecture.

\section{Extended Efficiency Analysis}
\label{app:extended_efficiency}

This section visualises the cost–accuracy trade–off of the five compression baselines—FastV, PDrop, VisionZip, D$^{3}$ToM, and D$^{3}$ToM-t.  LaViDa is omitted to avoid axis stretching and to focus the discussion on the practical operating regime ($10\%$–$50\%$ token budgets).

Figure~\ref{fig:tok_acc} confirms that accuracy decreases monotonically as fewer visual tokens are kept.  The slope, however, varies markedly among methods.  FastV drops by $11.6$ \% when the budget contracts from $50\%$ to $10\%$, whereas D$^{3}$ToM-t degrades by only $1.3$ \%, retaining $96.75\%$ of the full-model score at the most aggressive $10\%$ setting.

Figure~\ref{fig:flops_acc} casts the same results against relative compute.  D$^{3}$ToM-t achieves $96.75\%$ accuracy at $30.2\%$ FLOPs, while VisionZip reaches merely $88.68\%$ at a similar $19.6$–$32.6\%$ compute envelope and PDrop stays below $91.03\%$ even at $52.2\%$ FLOPs.  The convex hull of the plot therefore places D$^{3}$ToM-t on the Pareto frontier across the examined range.

Figure~\ref{fig:tok_flops} shows that, for a given retention level, all merge algorithms yield comparable theoretical FLOP reductions; differences arise from implementation overhead and are reflected in wall-clock time.  Figure~\ref{fig:tok_time} corroborates this: D$^{3}$ToM-t lowers latency from $786.4$ s at $50\%$ tokens to $493.9$ s at $10\%$, a $2.34\times$ speed-up that closely tracks the FLOP savings.

Taken together, the plots visually substantiate the numerical claim that timetable-aware merging (D$^{3}$ToM-t) provides the best cost–effectiveness: up to $69.8\%$ compute saved and $2.3\times$ faster inference while preserving at least $96.7\%$ of baseline accuracy on average.

\section{Broader Impacts}
\label{app:broader_impacts}

Our work introduces D\textsuperscript{3}ToM, the first method to accelerate diffusion-based MLLMs by exploiting visual-token redundancy directly within the iterative denoising loop. The core innovation is a dual-dynamic mechanism: the token merge pattern adapts at each step based on the model's evolving attention, while the merge ratio follows a coarse-to-fine schedule.

By reducing computational requirements by up to 70\% with minimal impact on accuracy, D\textsuperscript{3}ToM significantly lowers the barrier to entry for deploying state-of-the-art diffusion MLLMs. This enhances sustainability by reducing the energy cost per query and democratizes access for researchers and organizations with limited computational resources.

Our current validation is focused on image understanding benchmarks due to the limitation of modes. Besides, we use a single diffusion language model backbone LLaDA-8B architecture. Promising future research directions include extending the future diffusion-based video MLLMs and curiosity language model backbone such as Dream-7B. Critically, future work should also investigate methods to mitigate the risk of bias amplification. This could involve developing fairness-aware merge schedules or conducting detailed attribution analyses on demographically balanced datasets to ensure that information critical to equity is preserved during the merging process.

\section{Additional Visualizations}
\label{app:additional_vis}
\subsection{Attention dynamics during diffusion decoding}

Figure~\ref{fig:attn_dynamics_grid} visualizes the evolution of attention patterns, revealing a dual trend of specialization across both time and network depth. Temporally (left to right), attention refines from a diffuse state at early denoising steps to a highly sparse pattern by the final steps. Spatially (top to bottom), this sparsity is most pronounced in deeper layers (e.g., L24, L32), which concentrate computation on a small subset of visual tokens, while shallower layers retain a broader, more contextual focus.

These evolving patterns are a direct consequence of our decider-guided mechanism. The bright vertical bands in the `Output-to-Image` attention (lower-right quadrant) show newly generated "decider" tokens focusing on salient visual information. This attention pattern is used to score and prune the visual token set. The resulting sparsity in the `Image-to-Image` attention (upper-right quadrant) is the direct effect of this pruning, as merged tokens are removed from the computation. This provides clear visual evidence that our method successfully reallocates computational resources in the visual domain based on the dynamically generated text.
\subsection{Dynamic Decider-Guided Importance Peaks}
\label{app:importance_peaks}

Our decider-guided mechanism computes an importance score for each visual token group at every denoising step. Figure~\ref{fig:importance_evolution} visualizes the evolution of these scores by highlighting the visual patch that receives the maximum importance score at each of the 32 steps. The visualization clearly demonstrates that visual salience is dynamic; as the model refines its generated response from an initial coarse draft to a final polished output, the focus of its attention shifts across different regions of the input image. For full reproducibility, the complete set of importance scores for all groups at every step is provided in the supplementary materials.

\end{document}